\documentclass[10pt,twocolumn,letterpaper]{article}

\usepackage[accsupp]{axessibility}  % Improves PDF readability for those with disabilities.

\usepackage{wacv}
\usepackage{times}
\usepackage{epsfig}
\usepackage{graphicx}
\usepackage{amsmath}
\usepackage{amssymb}
\usepackage{booktabs}

\usepackage{multirow}

\wacvfinalcopy % *** Uncomment this line for the final submission

\ifwacvfinal
\usepackage[breaklinks=true,bookmarks=false]{hyperref}
\else
\usepackage[pagebackref=true,breaklinks=true,colorlinks,bookmarks=false,citecolor = blue]{hyperref}
\fi

\newcommand{\parsection}[1]{\vspace{1mm}\noindent\textbf{#1 }}

\begin{document}

%%%%%%%%% TITLE
\title{Dense Prediction with Attentive Feature Aggregation}

\author{
Yung-Hsu Yang$^{1}$ \\
% National Tsing Hua University \\
{\tt\small royyang@gapp.nthu.edu.tw}
% For a paper whose authors are all at the same institution,
% omit the following lines up until the closing ``}''.
% Additional authors and addresses can be added with ``\and'',
% just like the second author.
% To save space, use either the email address or home page, not both
\and
Thomas E. Huang$^{2}$ \\
% ETH Zürich \\
{\tt\small thomas.huang@vision.ee.ethz.ch}
\and
Min Sun$^{1}$ \\
% National Tsing Hua University \\
% Institution1 address \\
{\tt\small sunmin@ee.nthu.edu.tw}
\and
Samuel Rota Bulò$^{3}$ \\
% Facebook Reality Labs \\
{\tt\small rotabulo@fb.com}
\and
Peter Kontschieder$^{3}$ \\
% Facebook Reality Labs \\
{\tt\small pkontschieder@fb.com}
\and
Fisher Yu$^{2}$ \\
% ETH Zürich \\
{\tt\small i@yf.io}
\and
$^1$National Tsing Hua University \quad $^2$ETH Z\"urich \quad $^3$Meta Reality Labs Z\"urich
}

\maketitle
\thispagestyle{empty}

\begin{abstract}
Aggregating information from features across different layers is essential for dense prediction models. Despite its limited expressiveness, vanilla feature concatenation dominates the choice of aggregation operations.
In this paper, we introduce Attentive Feature Aggregation (AFA) to fuse different network layers with more expressive non-linear operations.
AFA exploits both spatial and channel attention to compute weighted averages of the layer activations.
Inspired by neural volume rendering, we further extend AFA with Scale-Space Rendering (SSR) to perform a late fusion of multi-scale predictions.
AFA is applicable to a wide range of existing network designs.
Our experiments show consistent and significant improvements on challenging semantic segmentation benchmarks, including Cityscapes and BDD100K at negligible computational and parameter overhead.
In particular, AFA improves the performance of the Deep Layer Aggregation (DLA) model by nearly 6\% mIoU on Cityscapes.
Our experimental analyses show that AFA learns to progressively refine segmentation maps and improve boundary details, leading to new state-of-the-art results on boundary detection benchmarks on NYUDv2 and BSDS500.
Code and video resources are available at http://vis.xyz/pub/dla-afa.
\end{abstract}

\vspace{-0.12in}
\section{Introduction} \label{sec:intro}
Dense prediction tasks such as semantic segmentation~\cite{cordts2016cityscapes,neuhold2017mapillary,yu2020bdd100k} and boundary detection~\cite{silberman2012indoor,arbelaez2010contour} are fundamental enablers for many computer vision applications. 
Semantic segmentation requires predictors to absorb intra-class variability while establishing inter-class decision boundaries.
Boundary detection also requires an understanding of fine-grained scene details and object-level boundaries.
A popular solution is to exploit the multi-scale representations to balance preserving spatial details from shallower features and maintaining relevant semantic context in deeper ones.

\begin{figure}
	\centering
	\includegraphics[width=0.47\textwidth]{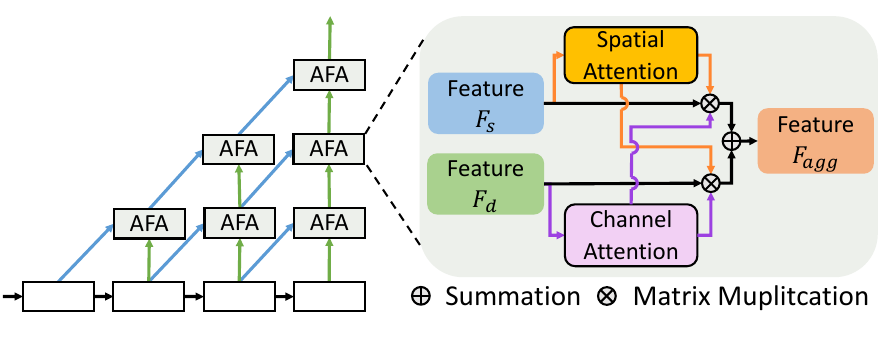}
	\vspace{-0.10in}
	\caption{ Attentive Feature Aggregation. $F_s$ is the shallower input feature and $F_d$ is the deeper one. We use attention to aggregate different scale or level information and obtain aggregated feature $F_\mathtt{agg}$ with rich representation.}
	\label{fig:teaser}
	\vspace{-0.24in}
\end{figure}

There are two major approaches to obtaining effective multi-scale representations.
Dilated convolutions~\cite{yu2015multi} can aggregate context information while preserving spatial information.
Most of the top performing segmentation methods adopt this approach~\cite{chen2018encoder,zhao2017pyramid,yu2017dilated,wang2020axial} to extract contextual pixel-wise information.
The drawback is the extensive usage of layer memory for storing high-resolution feature maps.
An alternative approach is to progressively downsample the layer resolution as in image classification networks and then upsample the resolution by aggregating information from different layer scales with layer concatenations~\cite{long2015fully,yu2018deep,ronneberger2015u}.
Methods using this approach achieve state-of-the-art results with reduced computational efforts and fewer parameters~\cite{wang2020deep}.
Even though many works design new network architectures to effectively aggregate multi-scale information, the predominant aggregation operations are still feature concatenation or summation~\cite{ronneberger2015u,long2015fully,wang2020deep,yu2018deep,zhao2018icnet}.
These linear operations do not consider feature interactions or selections between different levels or scales.

We propose \emph{Attentive Feature Aggregation} (AFA) as a nonlinear feature fusion operation to replace the prevailing tensor concatenation or summation strategies.
Our attention module uses both spatial and channel attention to learn and predict the importance of each input signal during fusion.
Aggregation is accomplished by computing a linear combination of the input features at each spatial location, weighted by their relevance.
Compared to linear fusion operations, our AFA module can attend to different feature levels depending on their importance.
AFA introduces negligible computation and parameter overhead, and can be easily used to replace fusion operations in existing methods.
Fig.~\ref{fig:teaser} illustrates the concept of AFA.

Another challenge of dense prediction tasks is that fine details are better handled in higher resolutions but coarse information are better in lower resolutions.
Multi-scale inference~\cite{chen2017deeplab,chen2016attention,tao2020hierarchical} has become a common approach to alleviate this trade-off, and using an attention mechanism is now a best practice.
Inspired by neural volume rendering~\cite{drebin1988volume,mildenhall2020nerf}, we extend AFA to \emph{Scale-Space Rendering} (SSR) as a novel attention mechanism to fuse multi-scale predictions.
We treat the prediction from each scale as sampled data in the scale-space and leverage the volume-rendering formulation to design a coarse-to-fine attention and render the final results.
Our SSR is
% proved to being
robust against the gradient vanishing problem and saves resources during training, thus achieving higher performance.

We demonstrate the effectiveness of AFA when applied to a wide range of existing networks on both semantic segmentation and boundary detection benchmarks.
We plug our AFA module into various popular segmentation models: FCN~\cite{long2015fully}, U-Net~\cite{ronneberger2015u}, HRNet~\cite{wang2020deep}, and Deep Layer Aggregation (DLA)~\cite{yu2018deep}.
Experiments on several challenging semantic segmentation datasets including Cityscapes~\cite{cordts2016cityscapes} and BDD100K~\cite{yu2020bdd100k} show that AFA can significantly improve the segmentation performance of each representative model.
Additionally, \emph{AFA-DLA} has competitive results compared to the state-of-the-art models despite having fewer parameters and using less computation.
Furthermore, AFA-DLA achieves the new state-of-the-art performances on boundary detection datasets NYUDv2~\cite{silberman2012indoor} and BSDS500~\cite{arbelaez2010contour}.
We conduct comprehensive ablation studies to validate the advantages of each component of our AFA module.
Our source code will be released.

\section{Related Work} \label{related_work}
% Recent semantic segmentation and boundary detection methods are based on fully convolutional networks~\cite{long2015fully} and focus on incorporating different components to the architecture to improve the performance.

\parsection{Multi-Scale Context.}
To better handle fine details, segmentation models with convolutional trunks use low output strides.
However, this limits the receptive field and the semantic information contained in the final feature.
Some works utilize dilated backbones~\cite{yu2015multi} and multi-scale context~\cite{yang2018denseaspp,lin2019zigzagnet,fu2019adaptive} to address this problem.
PSPNet~\cite{zhao2017pyramid} uses a Pyramid Pooling Module to generate multi-scale context and fuse them as the final feature.
The DeepLab models~\cite{chen2018encoder} use Atrous Spatial Pyramid Pooling to assemble context from multiple scales, yielding denser and wider features.
In contrast, our AFA-DLA architecture extensively uses attention to conduct multi-scale feature fusion to increase the receptive field without using expensive dilated convolutions.
Thus, our model can achieve comparable or even better performance with much less computation and parameters.

\parsection{Feature Aggregation.}
Aggregation is widely used in the form of skip connections or feature fusion nodes in most deep learning models~\cite{long2015fully,ronneberger2015u,wang2020deep}.
The Deep Layer Aggregation~\cite{yu2018deep} network shows that higher connectivity inside the model can enable better performance with fewer parameters, but its aggregation is still limited to linear operators.
Recently, some works have explored better aggregation of multi-scale features.
\cite{takikawa2019gated, li2020semantic, li2020gated, zhang2020feature, huang2021fapn, huang2021alignseg} improves the original FPN architecture with feature alignment and selection during fusion.
CBAM~\cite{woo2018cbam} and SCA-CNN~\cite{chen2016sca} uses channel and spacial self-attention to perform adaptive feature refinement and improves convolutional networks in image classification, object detection and image captioning.
DANet~\cite{fu2019dual} appends two separate branches with Transformer~\cite{vaswani2017attention} self-attention as the proposed spatial and channel module on top of dilated FCN.
In contrast, our AFA module leverages extracted spatial and channel information during aggregation to efficiently select the essential features with respect to the property of input features.
With the efficient design, AFA can be directly adopted in popular architectures and extensively used at negligible computational and parameter overhead.

% \parsection{Attention Module.}
% Attention mechanisms are commonly used to improve the expressiveness of features for downstream tasks.
% Spatial attention captures the inter-spatial relationship of features, while channel attention associates different semantic information across input features.
% CBAM~\cite{woo2018cbam} and SCA-CNN~\cite{chen2016sca} uses channel and spacial self-attention to perform adaptive feature refinement and improves convolutional networks in image classification, object detection and image captioning.
% DANet~\cite{fu2019dual} appends two separate branches with the proposed spatial and channel attention module on top of dilated FCN.
% They refine the features in spatial and channel dimension respectively and fuse them through the summation operator as the final representation and boost the performance in the semantic segmentation task.
% Inspired by the aforementioned works, we propose attentive feature aggregation, AFA, to leverage attention mechanism to perform feature aggregation from different levels.
% Our AFA module generates attention with respect to the property of input features and can be directly adapted to in the popular architectures.
% Furthermore, without using heavy self-attention like DANet, AFA can be extensively used at negligible computational and parameter overhead.

\begin{figure*}[t]
	\centering
	\small
	\includegraphics[width=0.7\textwidth]{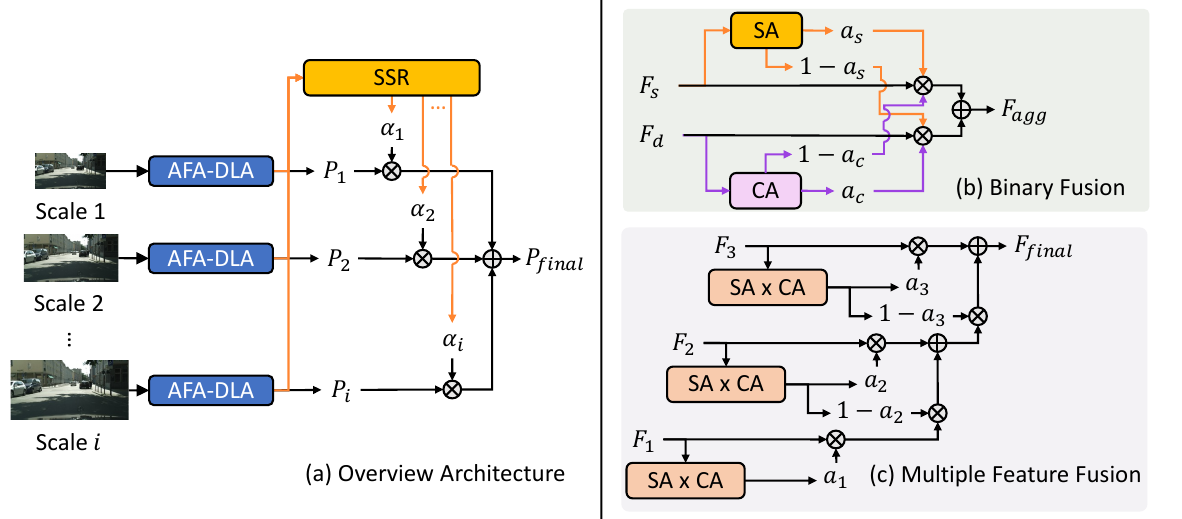}
	\caption{ 
        (a) Overview of AFA-DLA with SSR. We use two scales [0.5, 1.0] during training and more scales during inference to pursue higher performance. (b) Binary fusion module for input feature $F_s$ and $F_d$. SA denotes our spatial attention module and generates spatial attention $a_s$. CA stands for our channel attention module and responsible for channel attention $a_c$. (c) Multiple feature fusion module for three input features $F_1$, $F_2$, and $F_3$. \emph{SA $\times$ CA} represents computing spatial attention $a_s$ and channel attention $a_c$ first and then using element-wise multiplication to get the attention $a_i$ for $F_i$.
	}
	\label{fig:overview}
	\vspace{-0.18in}
\end{figure*}

\parsection{Multi-Scale Inference.}
Many computer vision tasks leverage multi-scale inference to get higher performance.
The most common way to fuse the multi-scale results is to use average pooling~\cite{chen2017deeplab,yu2018deep,li2020improving}, but it applies the same weighting to each scale.
Some approaches use an explicit attention model~\cite{chen2016attention,yang2018attention} to learn a suitable weighting for each scale.
However, the main drawback is the increased computational requirement for evaluating multiple scales.
To overcome this problem, HMA~\cite{tao2020hierarchical} proposes a hierarchical attention mechanism that only needs two scales during training but can utilize more scales during inference.
In this work, we propose scale-space rendering (SSR), a more robust multi-scale attention mechanism that generalizes the aforementioned hierarchical approach and exploits feature relationships in scale-space to improve the performance further.

\section{Method} \label{sec:method}
\setlength{\abovedisplayskip}{3pt}
\setlength{\belowdisplayskip}{4pt}
In this section, we introduce our attentive feature aggregation (AFA) module and then extend AFA to scale-space rendering (SSR) attention for multi-scale inference. The overview of the complete architecture is shown in Fig.~\ref{fig:overview}.

\subsection{Attentive Feature Aggregation}
Our attentive feature aggregation (AFA) module computes both spatial and channel attention based on the relation of the input feature maps.
The attention values are then used to modulate the input activation scales and produce one merged feature map.
The operation is nonlinear in contrast to standard feature concatenation or summation.
We use two different basic self-attention mechanisms to generate spatial and channel attention maps and reassemble them concerning the relation between the input features.

For the input feature $F_s \in \mathbb{R}^{C \times H \times W}$, the spatial attention uses a convolutional block $\omega_s$ consisting of two $3 \times 3$ convolutions to encode $F_s$. It is defined as
\begin{equation}
    a_s\triangleq\sigma(\omega_s(F_s))\,,
\end{equation}
where $a_s \in \mathbb{R}^{1 \times H \times W}$ and $\sigma$ is the sigmoid activation.

For computing the channel attention of input feature $F_d \in \mathbb{R}^{C \times H \times W}$, we first apply average pooling to get $F^{\mathtt{avg}}_d$ and max pooling to get $F^{\mathtt{max}}_d$.
Then, we further transform the features to $F^{\mathtt{avg}}_c$ and $F^{\mathtt{max}}_c$ using another convolutional block $\omega_c$, which consists of two $1 \times 1$ convolutions with a bottleneck input-output channel design.
We sum them up with equal weighting and use sigmoid $\sigma$ as the activation function to generate channel attention $a_c \in \mathbb{R}^{C \times 1 \times 1}$ as
\begin{equation}
    a_c\triangleq\sigma(\omega_c(\mathtt{AvgPool}(F_d)) + \omega_c(\mathtt{MaxPool}(F_d)))\,.
\end{equation}

With the basis of the above attention mechanisms, we design two types of AFA for different aggregation scenarios and enable the network to model complex feature interactions and attend to different features.

\begin{figure*}[t]
    \small
	\centering
	\includegraphics[width=0.85\textwidth]{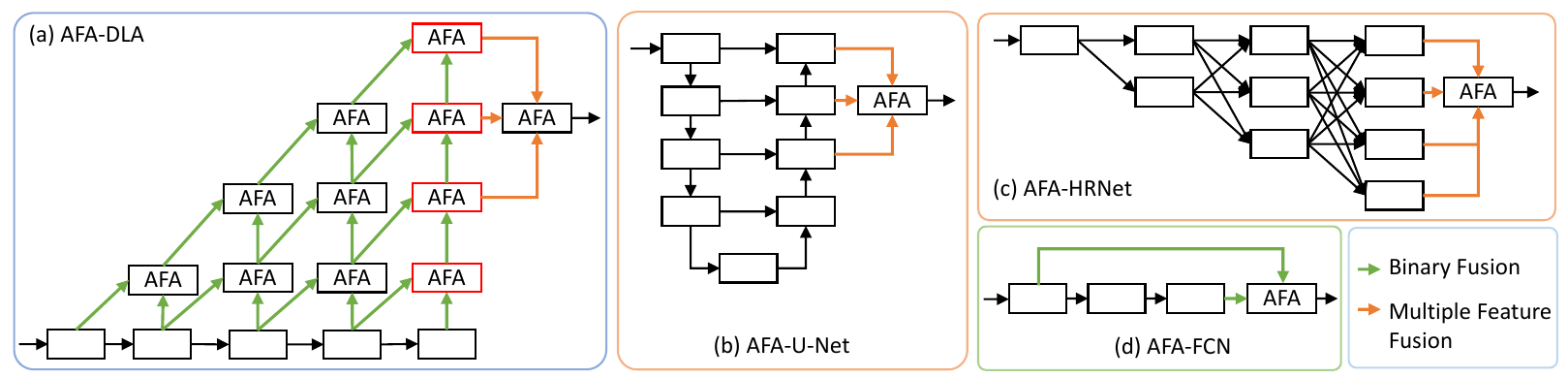}
	\caption{Segmentation models with our AFA module. We show parts of the original models related to feature aggregation and our modifications. Red blocks represent auxiliary segmentation heads added during training.}
	\label{fig:AFA-Seg}
	\vspace{-0.20in}
\end{figure*}

\parsection{Binary Fusion.}
We employ a simple attention-based aggregation mechanism using our spatial and channel attentions to replace standard binary fusion nodes.
When merging two input feature maps, we apply channel and spatial attention separately to capture the relation of input features.
As shown in Fig.~\ref{fig:overview}~(b), when two features are aggregated, we denote the shallower feature map as $F_s$ and the other as $F_d$.
$F_s$ is used to compute $a_s$ and $F_d$ is responsible for $a_c$, as the shallower layers will contain richer spatial information and the deeper ones will have more complex channel features.
% Then, we define our attentive binary fusion to aggregate the inputs and obtain the aggregated feature $F_{\mathtt{agg}}$ as
Then, we obtain the aggregated feature $F_{\mathtt{agg}}$ as
\begin{equation}
    F_\mathtt{agg}\triangleq a_s\odot{(1-a_c)\odot{F_s}}+(1-a_s)\odot{a_c\odot{F_d}}\,,
\end{equation}
where $\odot$ denotes element-wise multiplication (with broadcasted unit dimensions).
By leveraging the input features properties, our binary fusion is simple yet effective.

\parsection{Multiple Feature Fusion.}
We extend the binary fusion node to further fuse together multiple multi-scale features.
Recent works~\cite{ronneberger2015u,wang2020deep,yu2018deep} iteratively aggregate features across the model, but only exploit the final feature for downstream tasks, neglecting intermediate features computed during the aggregation process.
By applying AFA on these intermediate features, we give the model more flexibility to select the most relevant features.

Given $k$ multi-scale features $F_i$ for $i \in \{1,\ldots,k\}$, we first order them based on the amount of aggregated information they contain, \textit{i.e.}, a feature with higher priority will have gone through a higher number of aggregations.
Then, we compute both spatial and channel attention for each feature and take the product as the new attention.
The combined attention $a_i$ is defined as
\begin{equation}
    a_i\triangleq\mathtt{SA}(F_i)\odot{\mathtt{CA}(F_i)}\,,
\end{equation}
where $\mathtt{SA}$ denotes our spatial attention function and $\mathtt{CA}$ our channel attention function.
For fusing the multi-scale features, we perform hierarchical attentive fusion by progressively aggregating features starting from $F_1$ to $F_k$ to obtain the final representation $F_\mathtt{final}$ as
\begin{equation}
    F_\mathtt{final}\triangleq\sum_{i=1}^k \left[a_i\odot{F_i\odot{\prod_{j=i+1}^{k} (1-a_j)}}\right]\,.
\end{equation}
In Fig.~\ref{fig:overview} (c), we show an example of this process with $k=3$.
The new final representation $F_\mathtt{final}$ is an aggregation of features at multiple scales, combining information from shallow to deep levels.

AFA is flexible and can be applied to widely used segmentation models, as shown in Fig.~\ref{fig:AFA-Seg}.
In U-Net~\cite{ronneberger2015u} and HRNet~\cite{wang2020deep}, we add our multiple feature fusion module to fully utilize the previously unused aggregated multi-scale features.
In FCN~\cite{long2015fully}, we replace the original linear aggregation node in the decoder with our attentive binary fusion.
For DLA~\cite{yu2018deep}, we not only substitute the original aggregation nodes but also add our multiple feature fusion module.
Due to higher connectivity of its nodes, the DLA network can benefit more from our improved feature aggregation scheme, and thus we use AFA-DLA as our final model.

% \begin{figure}[t]
%     \centering
%     \includegraphics[width=0.47\textwidth]{images/Attention_Comparison.pdf}
%     \caption{(a) CBAM~\cite{woo2018cbam} module. CA and SA are their channel attention and spatial attention blocks. (b) DANet~\cite{fu2019dual} architecture. PAM stands for their proposed Position Attention Module while CAM denotes channel attention module.}
%     \label{fig:Attention_comparision}
% \end{figure}

\parsection{Comparison with other Attention Modules.}
Unlike previous attention methods~\cite{chen2016sca,woo2018cbam,fu2019dual}, AFA focuses on aggregating feature maps of different network layers to obtain more expressive representations with a lightweight module.
Compared to GFF~\cite{li2020gated}, AFA consumes $1/4$ FLOPs and model parameters for binary fusion and $1/2$ FLOPS and $1/5$ model parameters for multiple feature fusion.
Without using heavy self-attention mechanism as DANet~\cite{fu2019dual}, AFA consumes only $1/2$ FLOPs and model parameters with $1/4$ GPU memory under the same input features.
With a simple yet effective design, AFA can be extensively used in existing architectures without much additional overhead.

\begin{figure*}[t]
    \small
    \centering
    \small
    \setlength{\tabcolsep}{2.5pt}
    \begin{tabular}{cccccc}
        \toprule
        \multicolumn{2}{c}{scale = 0.5} &
        \multicolumn{2}{c}{scale = 1.0} &
        \multicolumn{2}{c}{scale = 1.5} \\
        \midrule
        \includegraphics[width=0.15\linewidth]{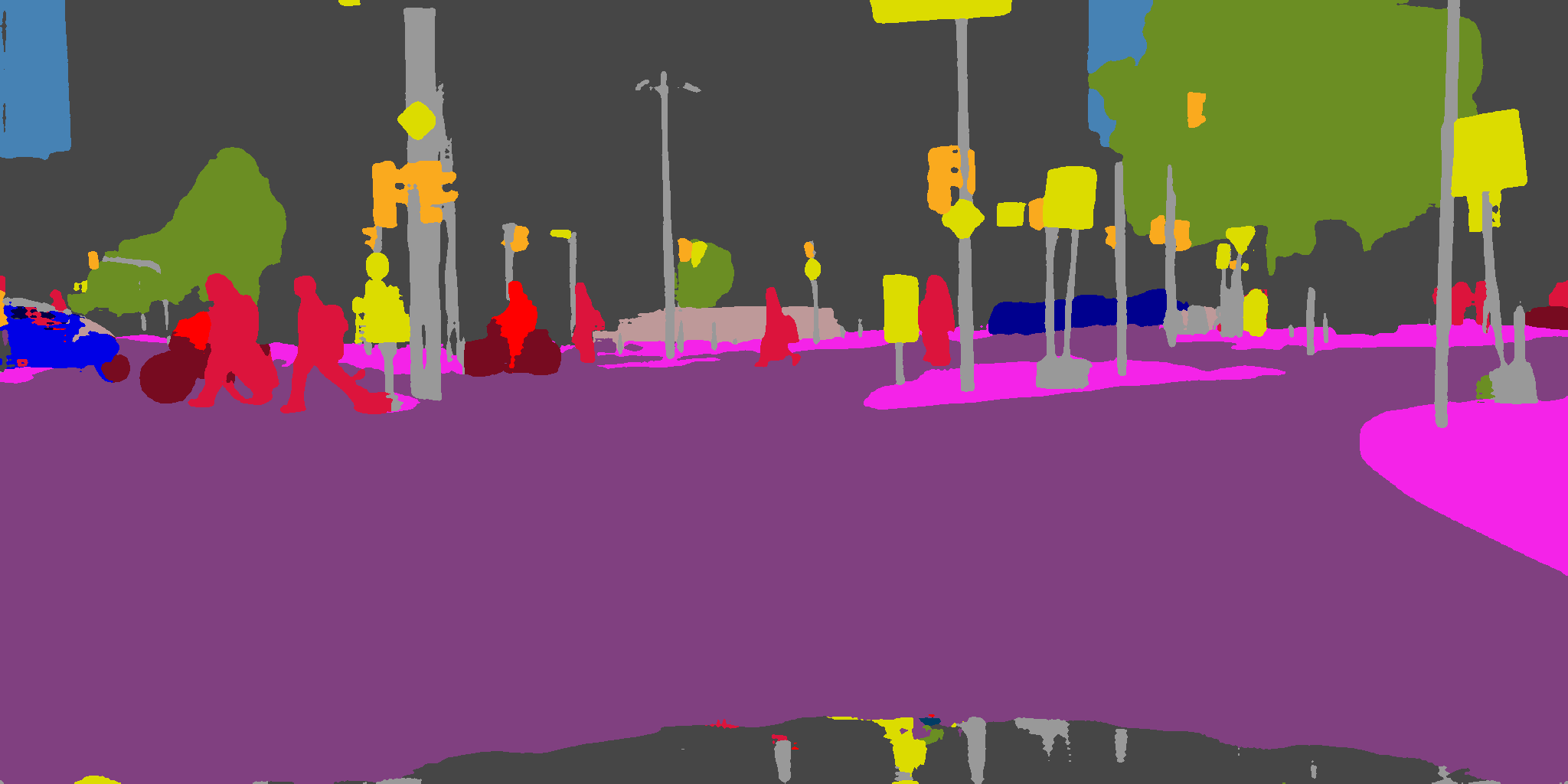} & \includegraphics[width=0.15\linewidth]{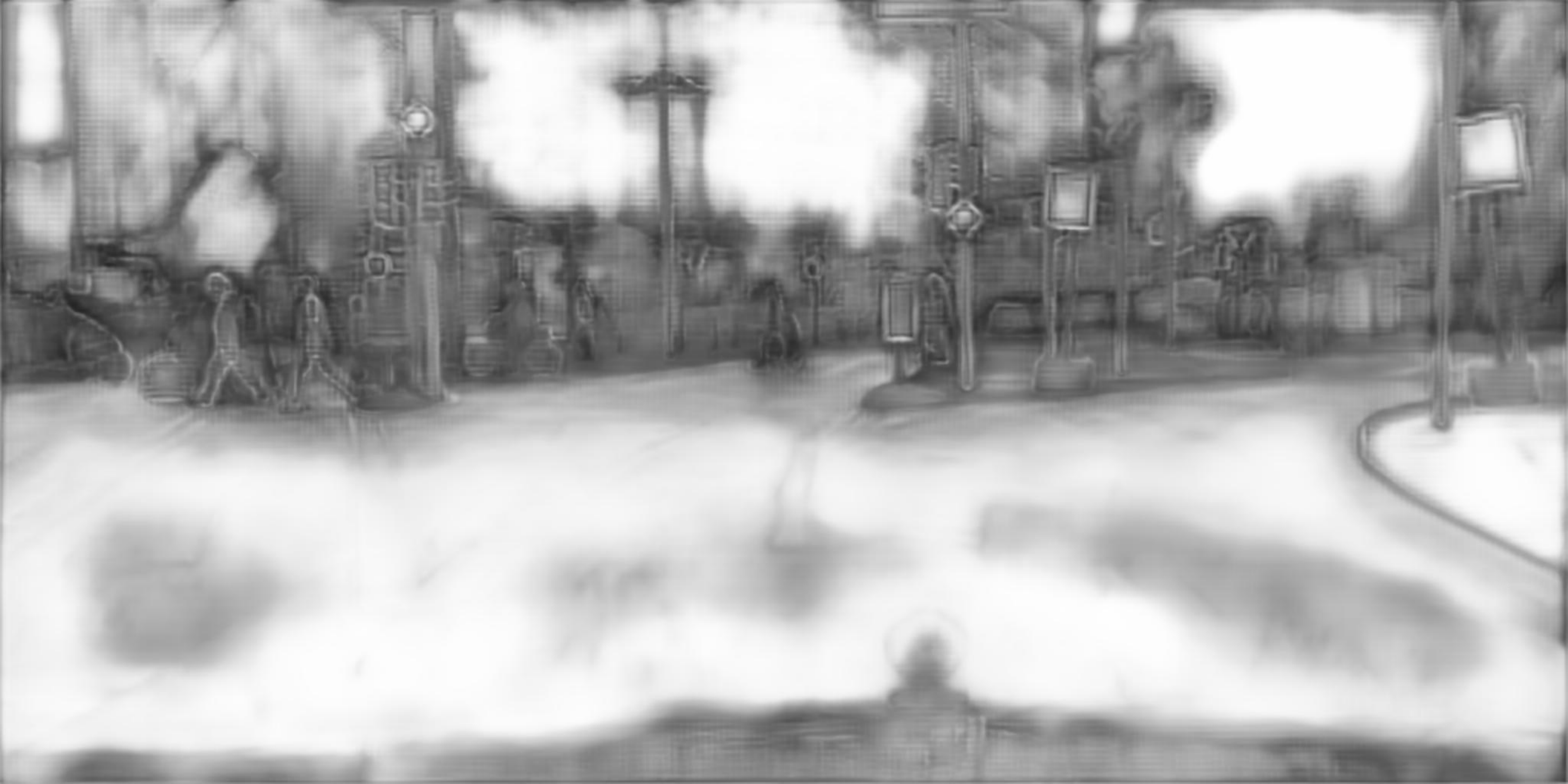} &
        \includegraphics[width=0.15\linewidth]{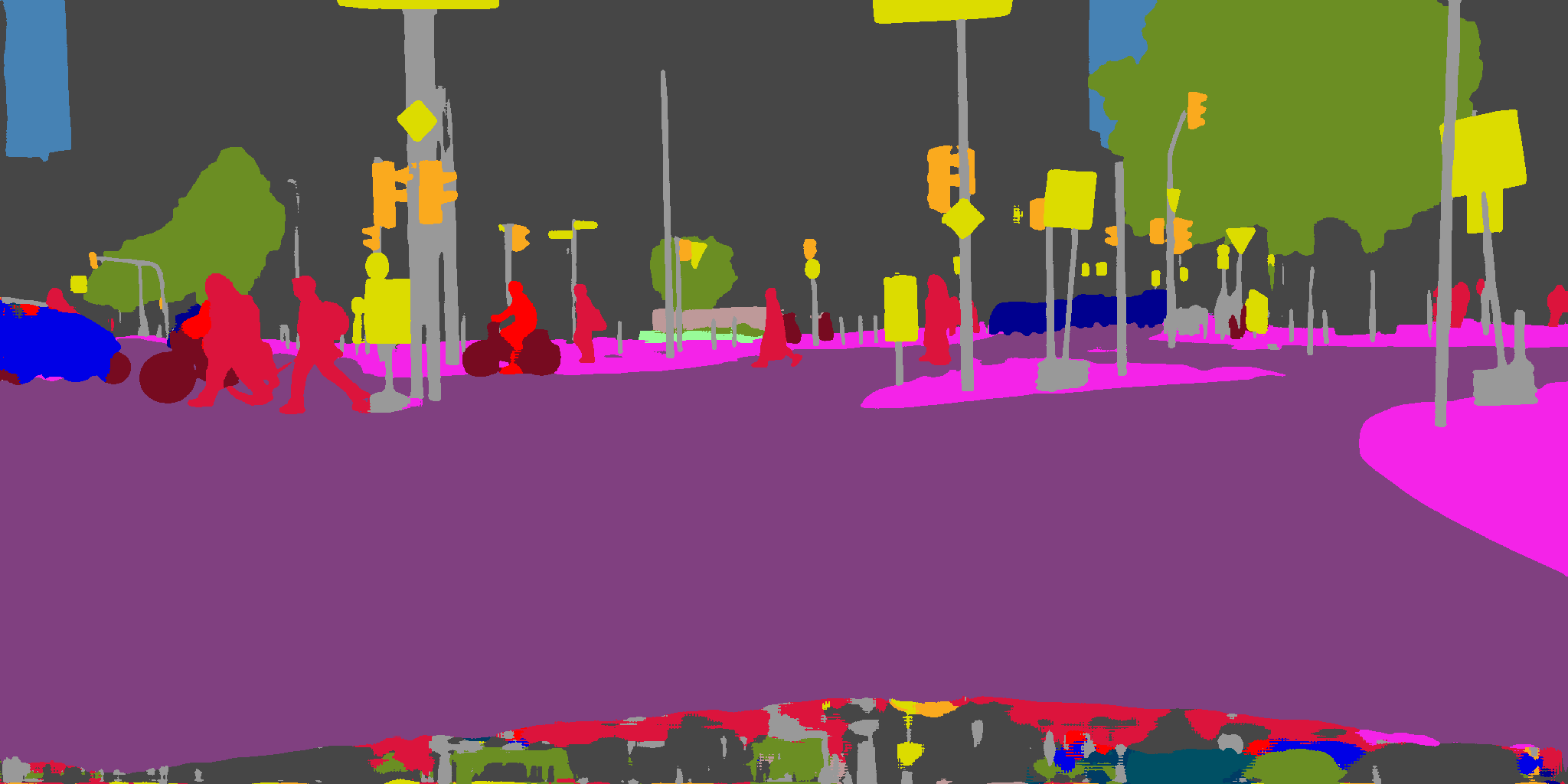} & \includegraphics[width=0.15\linewidth]{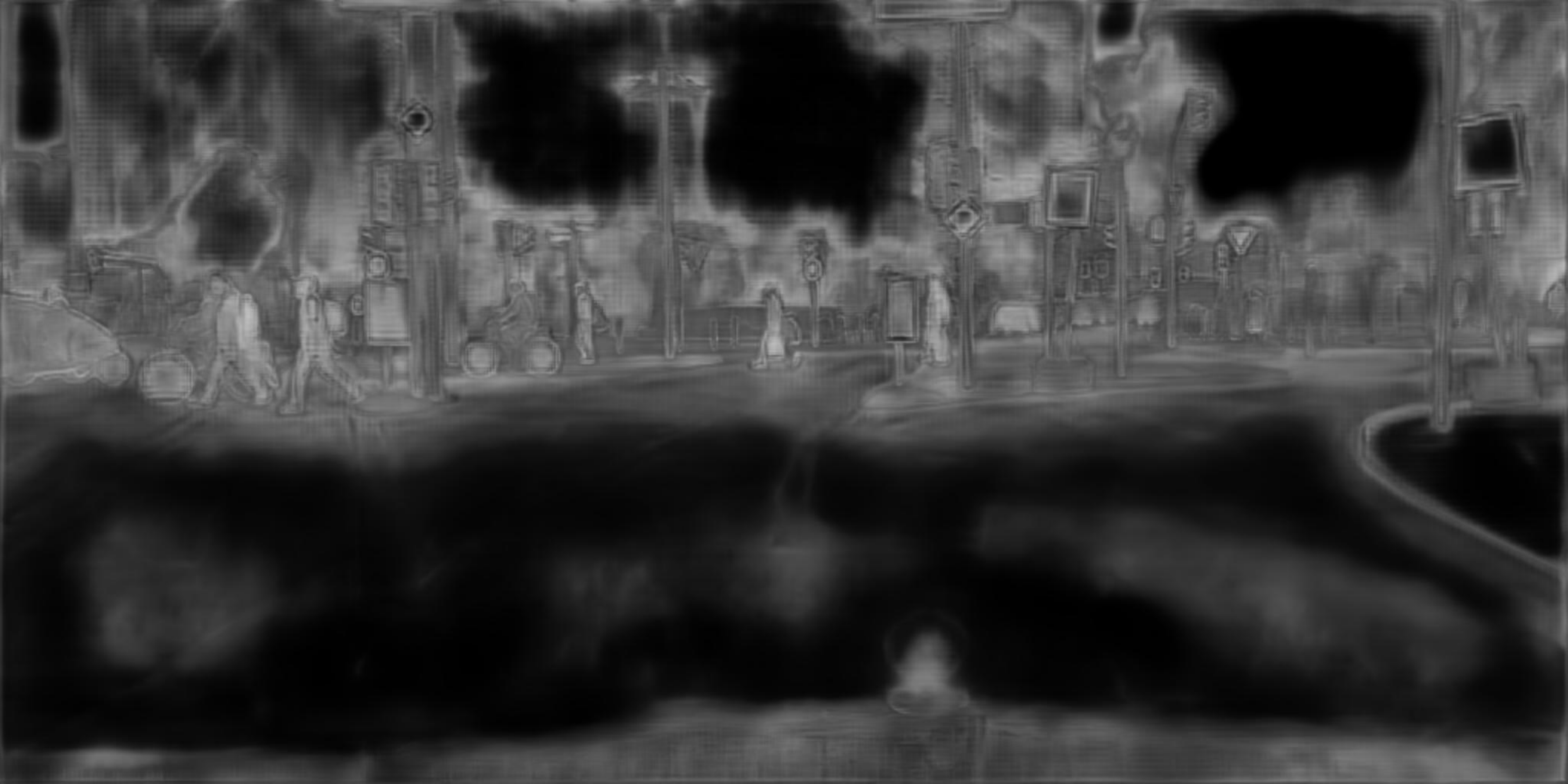} &
        \includegraphics[width=0.15\linewidth]{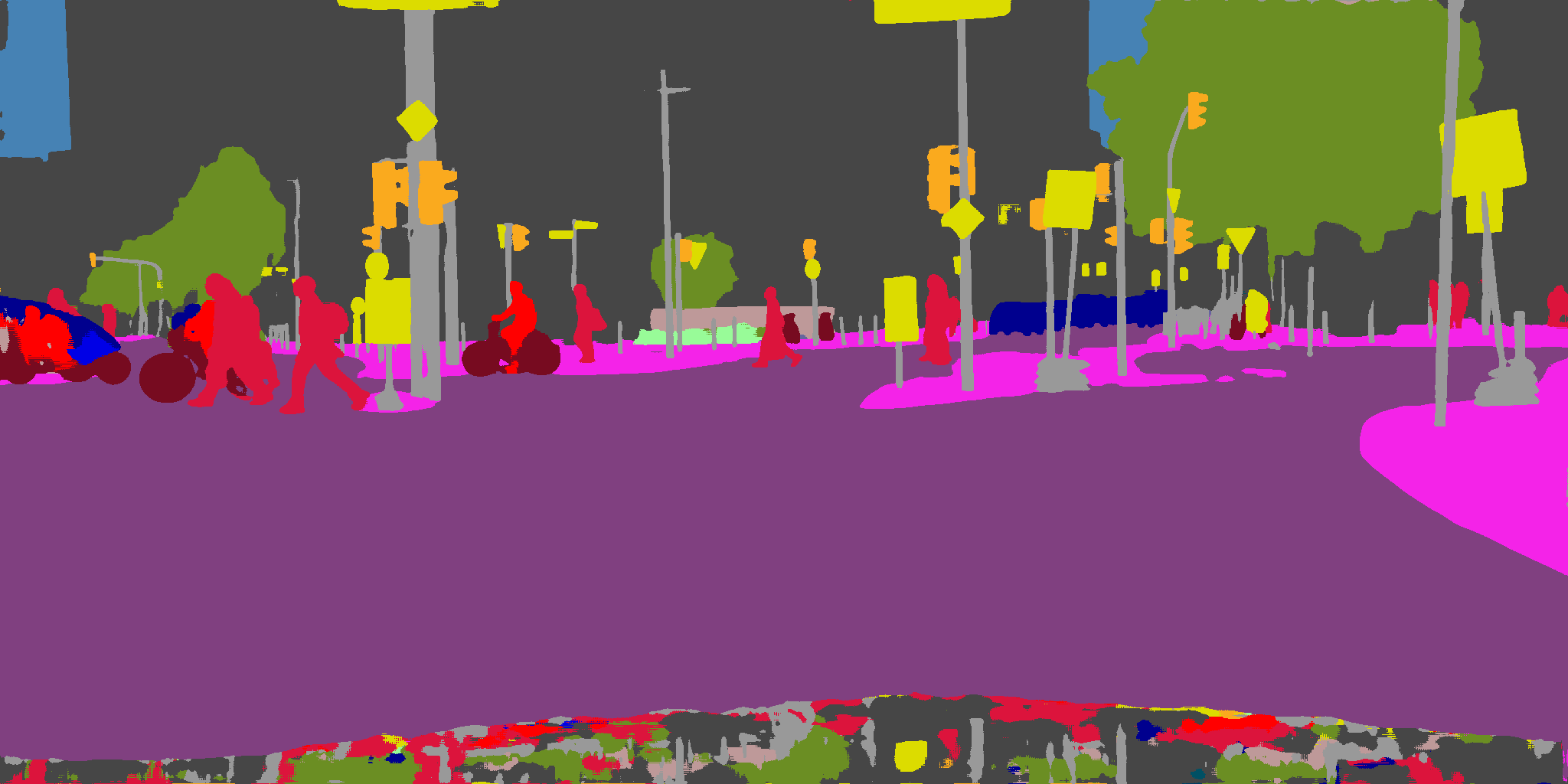} & \includegraphics[width=0.15\linewidth]{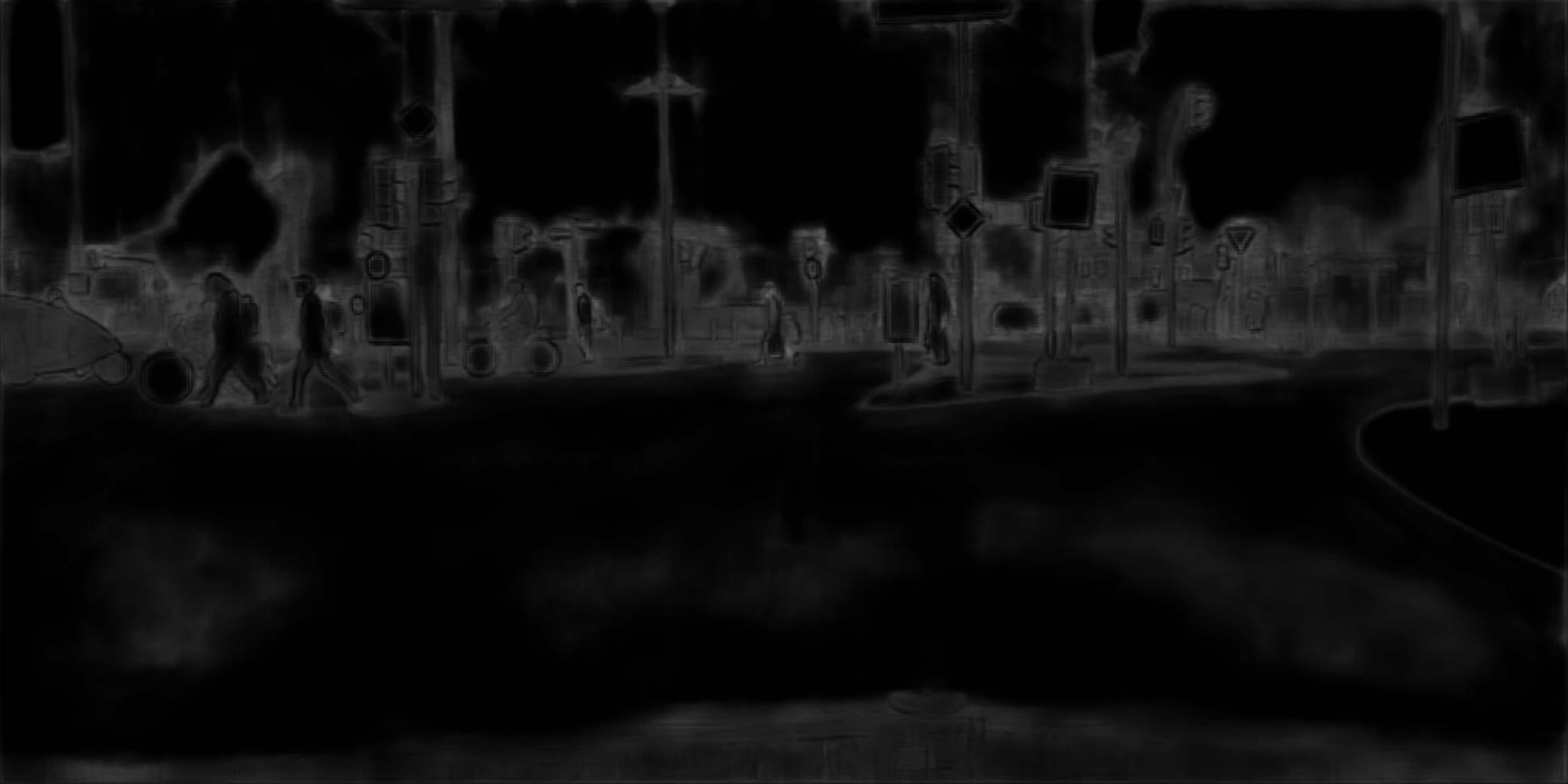} \\
        \includegraphics[width=0.15\linewidth]{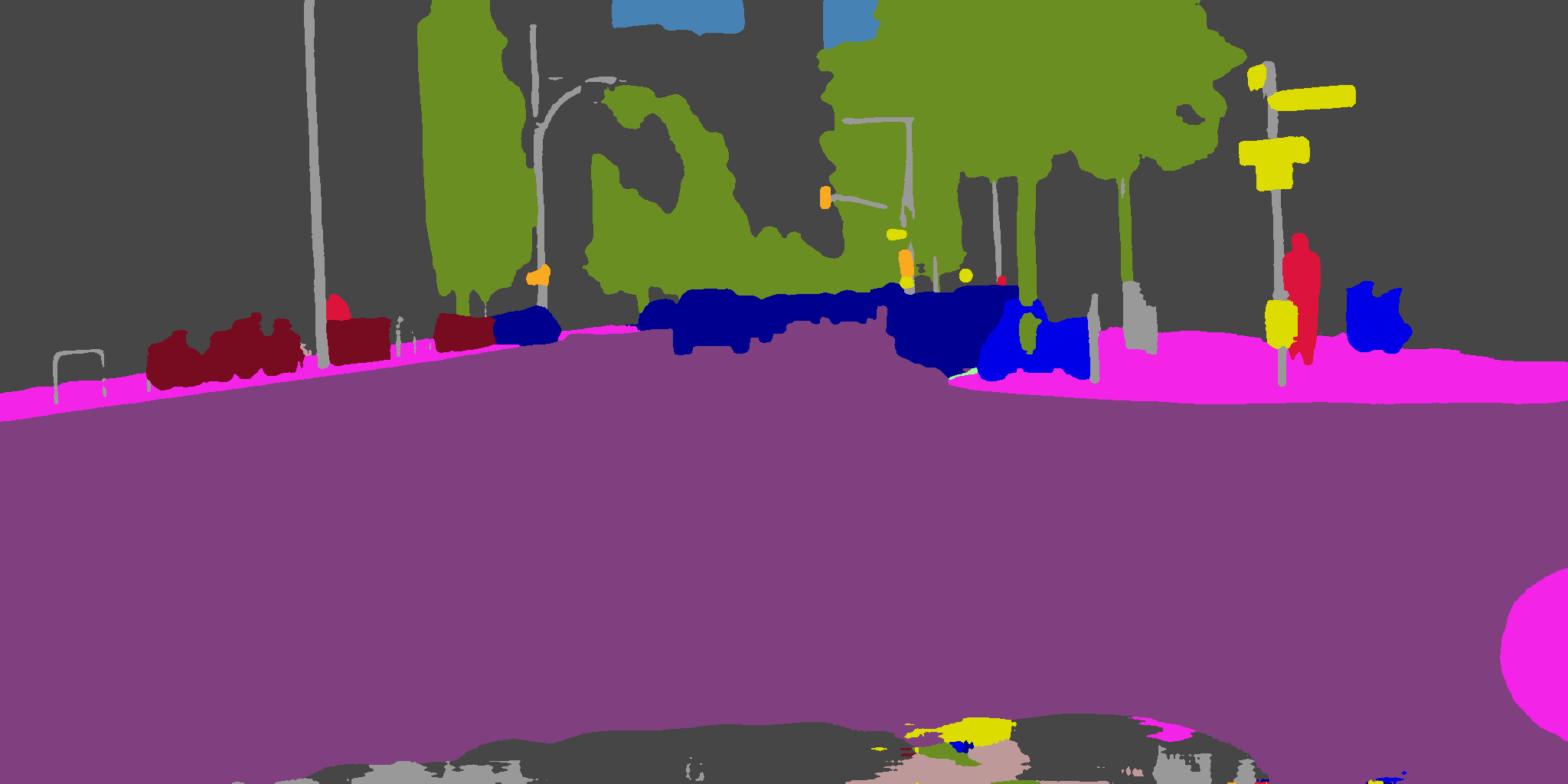} & \includegraphics[width=0.15\linewidth]{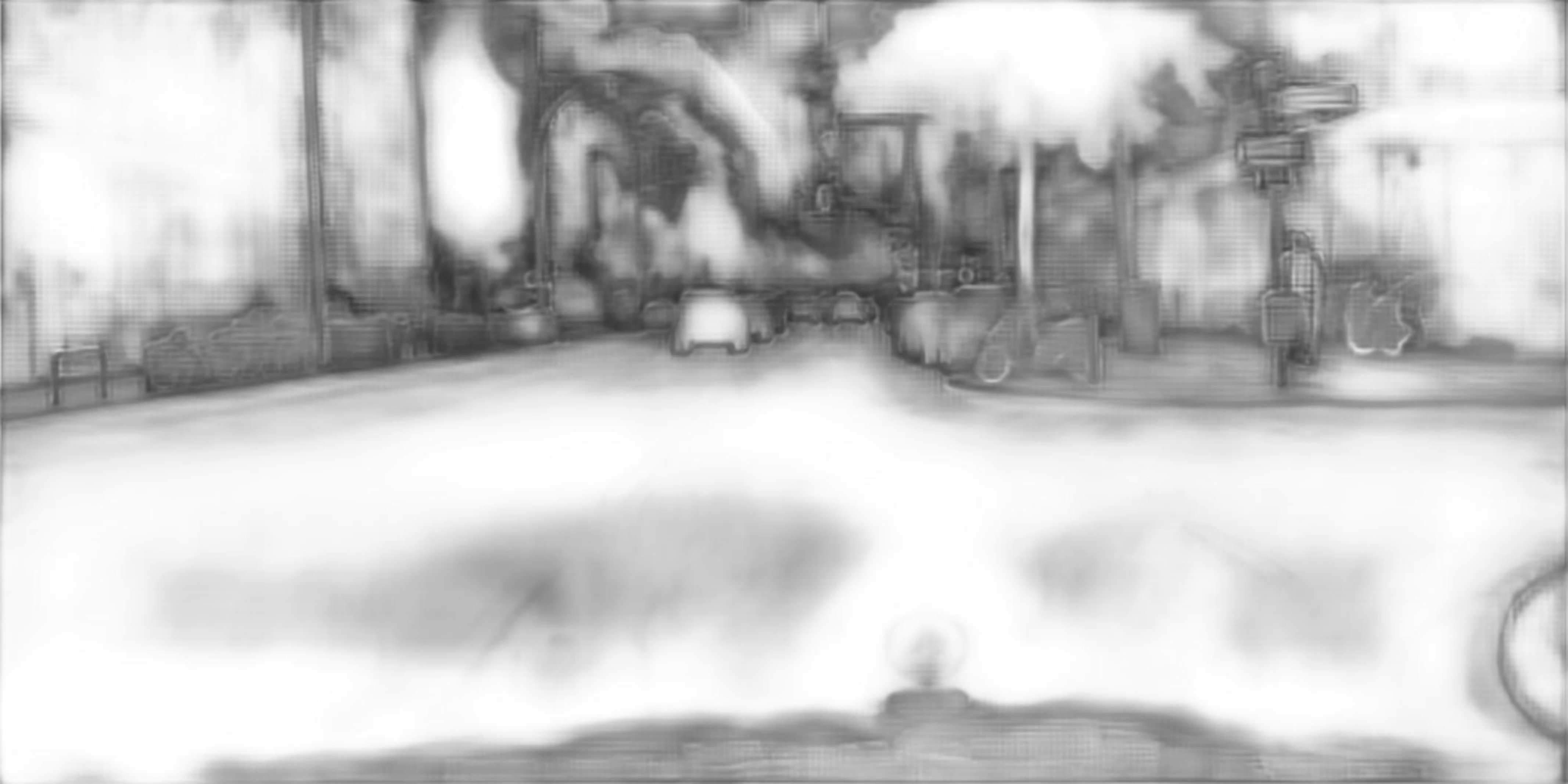} &
        \includegraphics[width=0.15\linewidth]{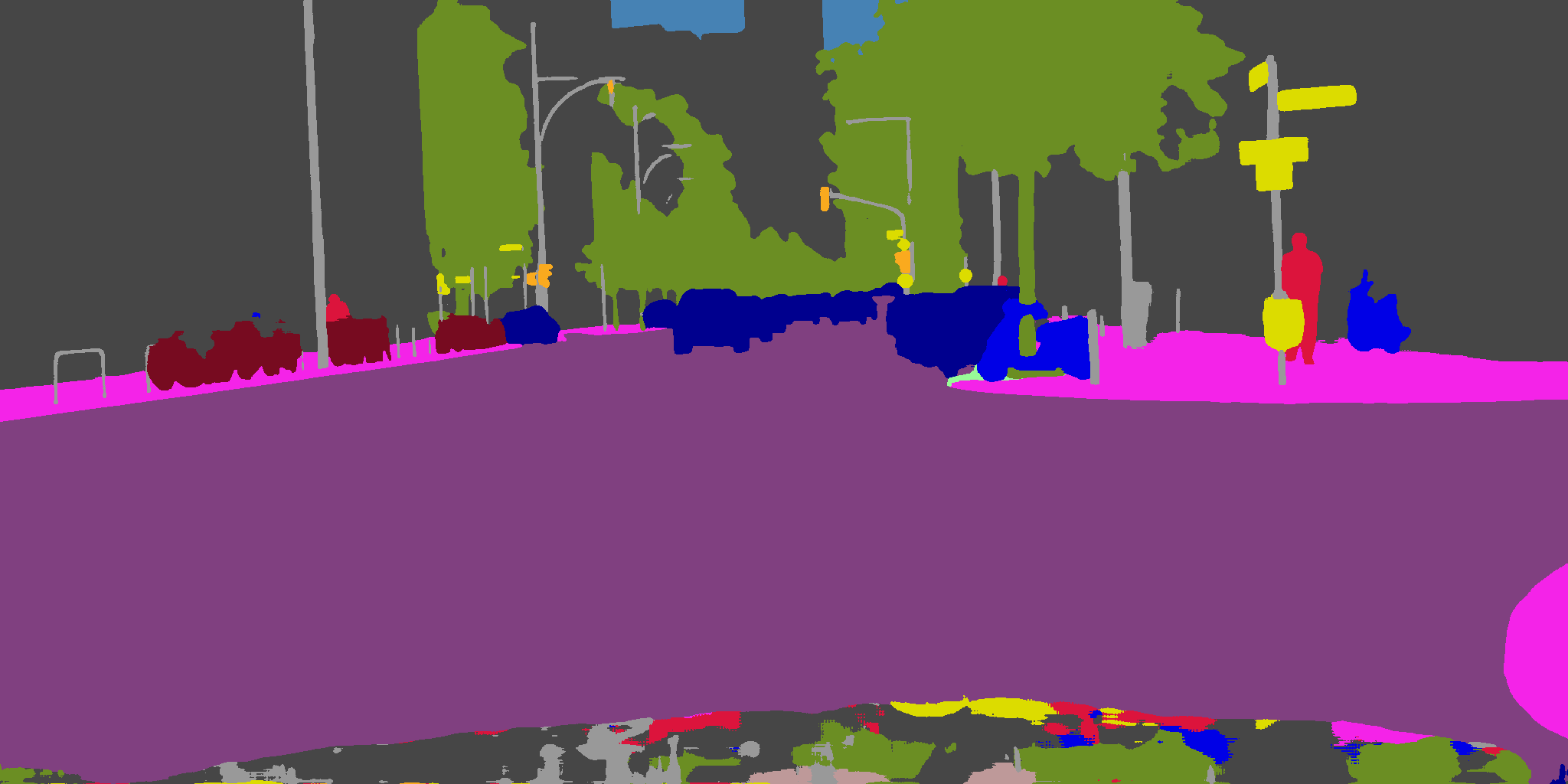} & \includegraphics[width=0.15\linewidth]{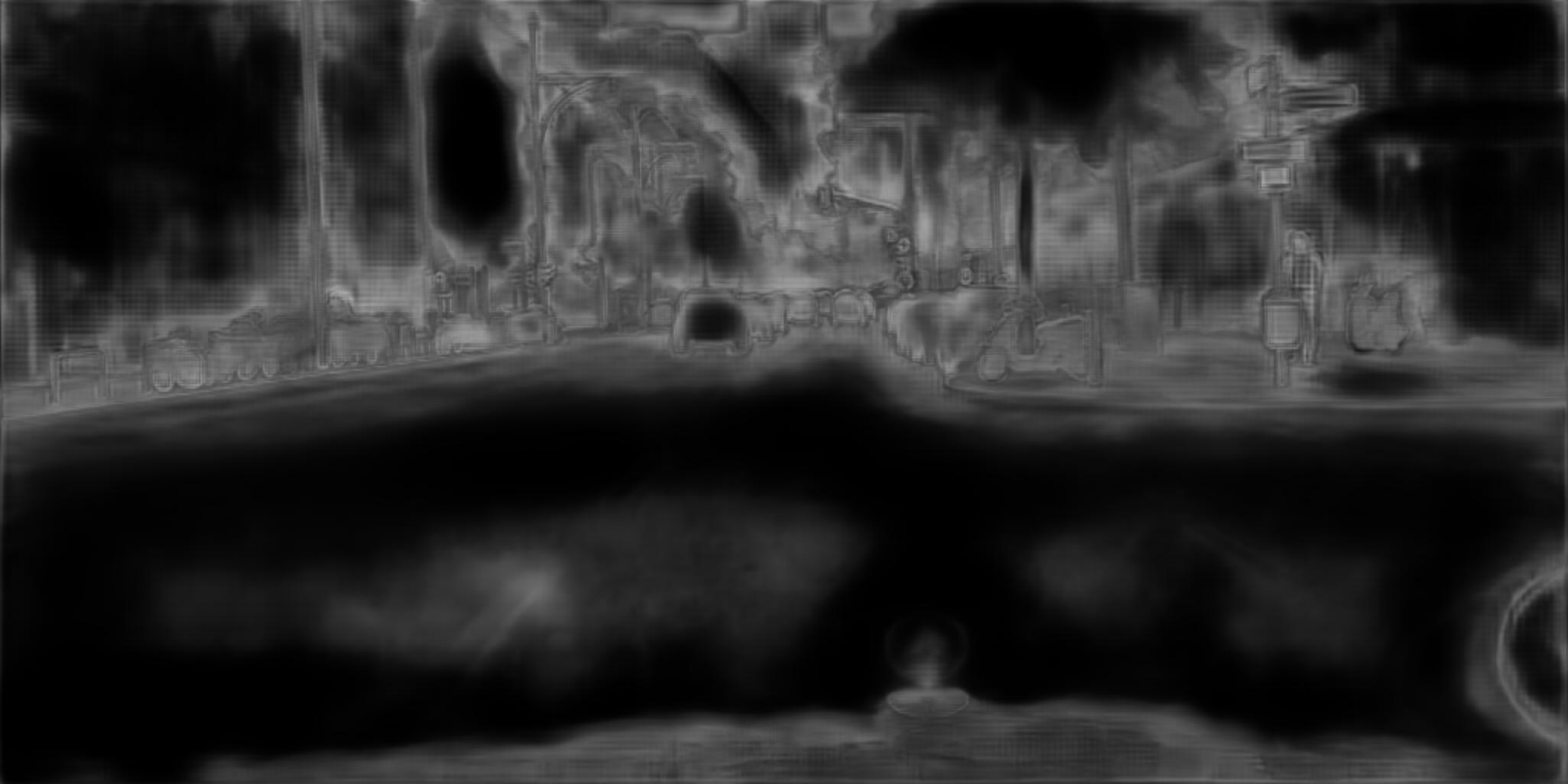} &
        \includegraphics[width=0.15\linewidth]{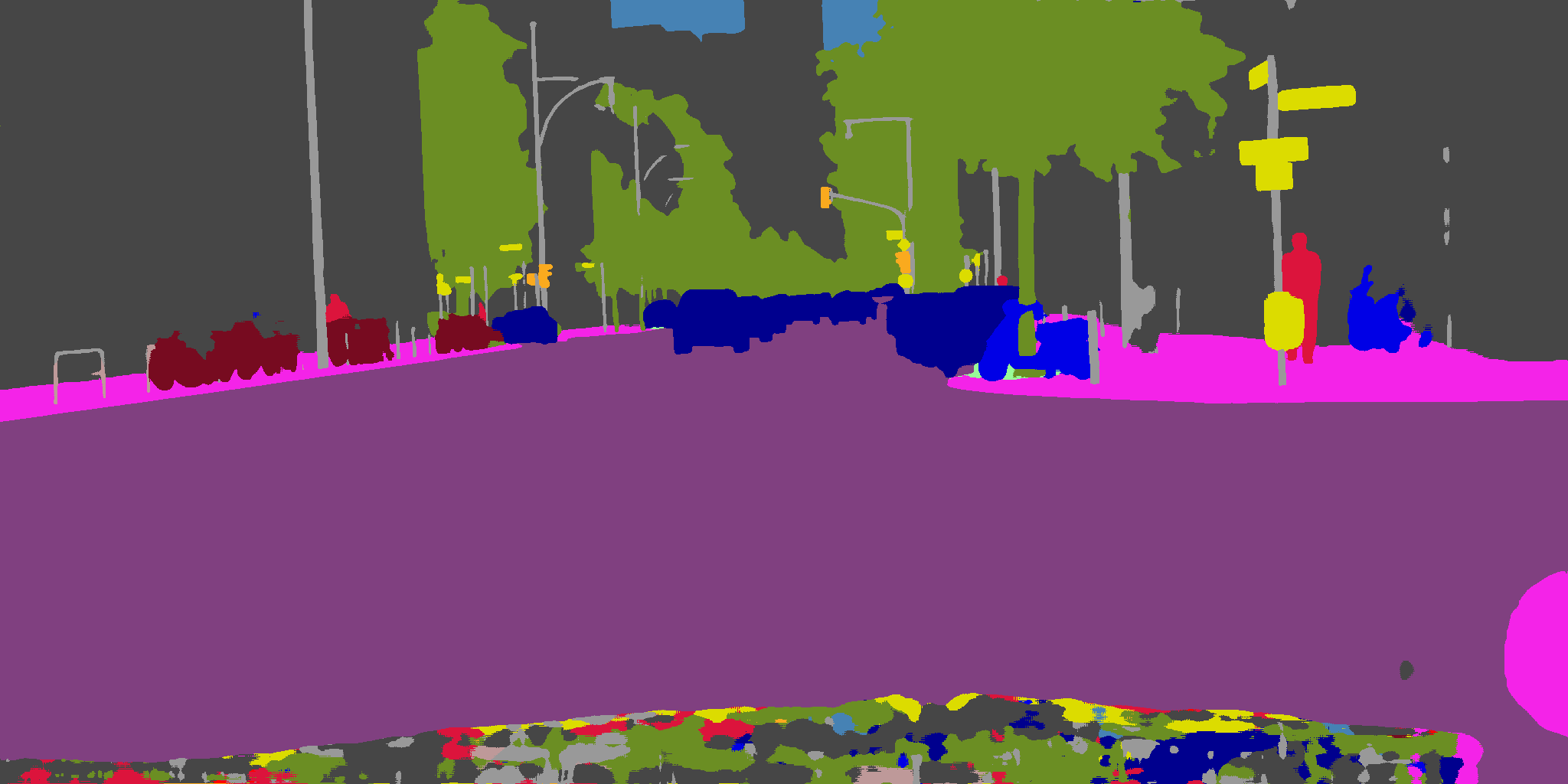} & \includegraphics[width=0.15\linewidth]{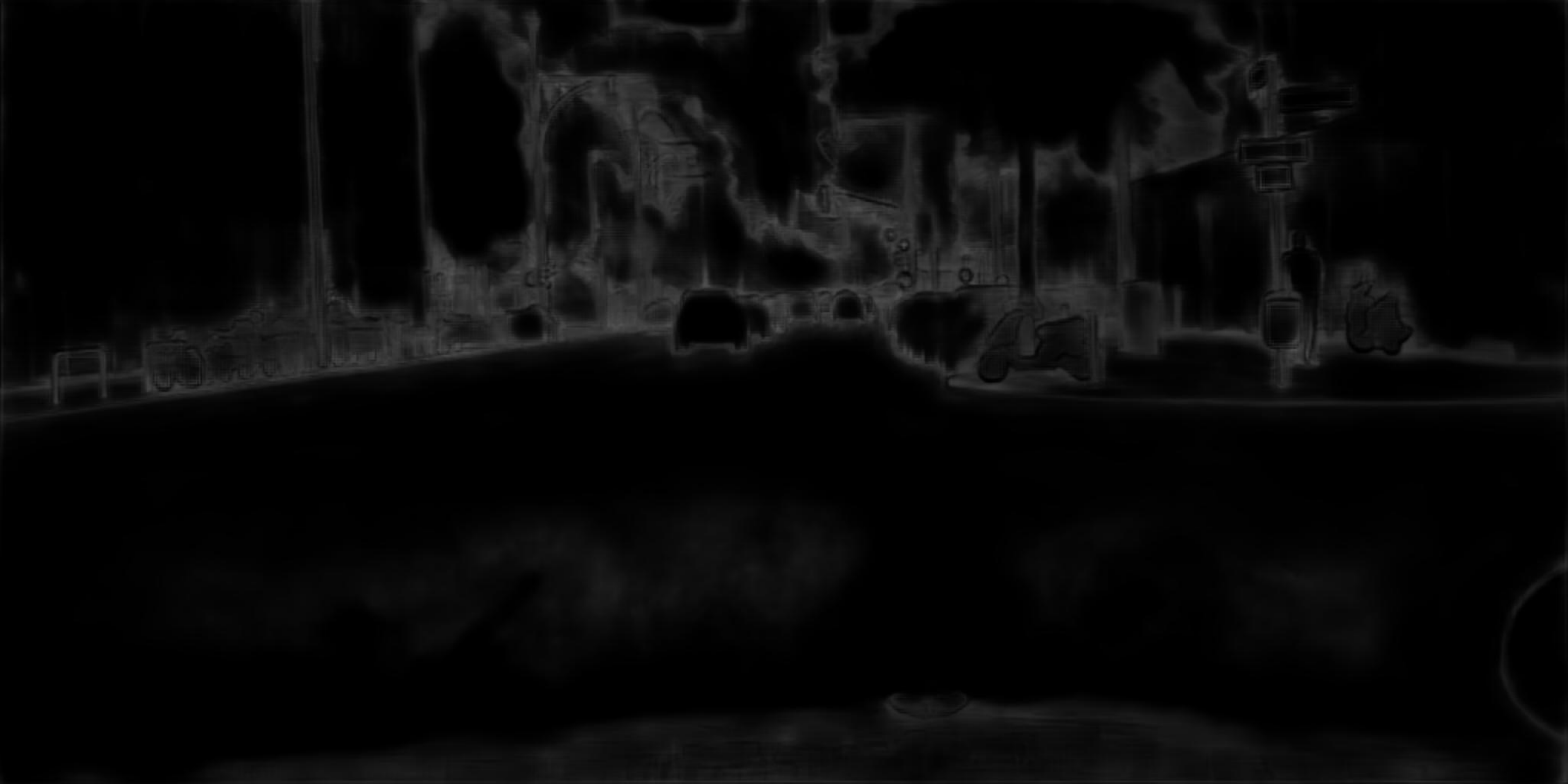} \\ 
        \bottomrule
    \end{tabular}
    \vspace{0.05in}
    \caption{Visualization of attention maps generated by scale-space rendering (SSR) with the predictions. Whiter regions denote higher attention. SSR learns to focus on detailed regions in larger scale images and on lower frequency information in smaller scale ones.}
	\label{fig:ms_results}
    \vspace{-0.20in}
\end{figure*}

\subsection{Scale-Space Rendering}
Multi-scale attention~\cite{chen2016attention,tao2020hierarchical} is typically used to fuse multi-scale predictions and can alleviate the trade-off in performance on fine and coarse details in dense prediction tasks.
However, repeated use of attention layers may lead to numerical instability or vanishing gradients, which hinders its performance.
To resolve this issue, we extend the attention mechanism mentioned above using a volume rendering scheme applied to the scale space.
By treating the multi-scale predictions as samples in a scale-space representation, this scheme provides a hierarchical, coarse-to-fine way of combining predictions using a scale-specific attention mechanism.
We will also show that our approach generalizes the hierarchical multi-scale attention method~\cite{tao2020hierarchical}.

Without loss of generality, we focus on a single pixel and assume that our model provides a dense prediction for the target pixel at $k$ different scales. The prediction for the $i$th scale is denoted by $P_i\in\mathbb R^d$. Accordingly, $P\triangleq(P_1,\ldots,P_k)$ denotes the feature representation of the target pixel in our scale-space. Furthermore, we assume that $i<j$ implies that scale $i$ is coarser than scale $j$.

Our target pixel can be imagined as a ray moving through scale-space, starting from scale $1$ towards scale $k$. We redesign the original hierarchical attention in the proposed multiple feature fusion mechanism to mimic the volume-rendering equation, where the volume is implicitly given by the scale-space. To this end, besides the feature representation $P_i$ at scale $i$, we assume our model to predict for the target pixel also a scalar $y_i\in\mathbb R$ so that $e^{-\phi(y_i)}$ represents the probability that the particle will cross scale $i$, given some non-negative scalar function $\phi:\mathbb R\to\mathbb R_+$. We can then express the scale attention $\alpha_i$ as the probability of the particle to reach scale $i$ and stop there, \textit{i.e.},
\begin{equation}
    \alpha_i(y)\triangleq\left[1-e^{-\phi(y_i)}\right]\prod_{j=1}^{i-1}e^{-\phi(y_j)}\,
\end{equation}
where $y\triangleq(y_1,\ldots,y_k)$.
Finally, the fused multi-scale prediction for the target pixel can be regarded as the ``rendered'' pixel, where the pixel features at the different scales $P_i$ are averaged by the attention coefficients $\alpha_i$ following the volume rendering equations. Accordingly, $P_\mathtt{final}\triangleq \sum_{i=1}^k P_i\alpha_i(y)$ represents the feature for the target pixel that we obtain after fusing $P$ across all scales with attention driven by $y$.

The proposed scale-space rendering (SSR) mechanism can be regarded as a generalization of the hierarchical multi-scale attention proposed in~\cite{tao2020hierarchical}, for the latter can be obtained from our formulation by simply setting $\phi(y_i)\triangleq\log(1+e^{y_i})$, \textit{i.e.}, $\phi$ is the soft-plus function, and by fixing $\phi(y_k)\triangleq\infty$.

\parsection{Choice of $\phi$.}
In our experiments, we use the absolute value function as our $\phi$, \textit{i.e.}, $\phi(y_i)\triangleq|y_i|$. This is motivated by a better preservation of the gradient flow through the attention mechanism, as we found existing attention mechanisms to suffer from vanishing gradient issues. Consider the Jacobian of the attention coefficients, which takes the form:
\begin{equation}
    J_{i\ell} \triangleq \frac{\partial \alpha_i(y)}{\partial y_\ell}=
    \begin{cases}
    \phi'(y_i)\prod_{j=1}^i e^{-\phi(y_j)} & \text{if $\ell=i$}\\
    0& \text{if $\ell>i$}\\
    -\phi'(y_\ell)\alpha_i(y)& \text{if $\ell<i$}\,.
    \end{cases}
\end{equation}
In the presence of two scales, this becomes:
\begin{equation}
    J=\begin{bmatrix}
    \phi'(y_1)a_1&0\\
    -\phi'(y_1)a_1(1-a_2)&\phi'(y_2)a_1a_2\,
    \end{bmatrix},
\end{equation}
where $a_i\triangleq e^{-\phi(y_i)}$.
As $a_1\to 0$, the gradient vanishes, for $J$ tends to a null matrix.
Otherwise, irrespective of the value of $a_2$, the gradient will vanish only depending on the choice of $\phi$.
In particular, by taking the absolute value as $\phi$ we have that the Jacobian will not vanish for $a_1>0$ and $(y_1,y_2)\neq(0,0)$, thus motivating our choice of using the absolute value as $\phi$.
If we consider instead the setting in HMA~\cite{tao2020hierarchical}, we have that $a_2=0$ and $\phi'(y_i)=1-a_i$.
It follows that the Jacobian vanishes also as $a_1\to 1$.
The conclusion is that the choice of $\phi$ plays a role in determining the amount of gradient that flows through the predicted attention and that the approach in HMA~\cite{tao2020hierarchical} is more subject to vanishing gradient issues than our proposed solution.
We compare HMA and SSR quantitatively in Section~\ref{sec:exp}.

To understand which parts of the image SSR attends to at each scale, we visualize generated attention maps in Fig.~\ref{fig:ms_results}.
Detailed regions are processed more effectively in larger scale images due to the higher resolution, while the prediction of lower frequency region is often better in smaller scales. SSR learns to focus on the right region for different scales and boosts the final performance.

We combine AFA-DLA with SSR to produce the final predictions.
As shown in Fig.~\ref{fig:overview}, AFA-DLA propagates information from different scales to the SSR module, which then generates attention masks $\alpha_i$ used to fuse the predictions $P_i$ to get our final prediction $P_{\mathtt{final}}$.

\parsection{Training Details.}
For fair comparison with other methods~\cite{yuan2020object,tao2020hierarchical}, we reduce the number of filters from 256 to 128 in the OCR~\cite{yuan2020object} module and add it after AFA-DLA to refine our final predictions.
Our final model can be trained at $k$ different scales. Due to the limitation of computational resources, we use $k=2$ for training and RMI~\cite{zhao2019region} to be the primary loss function $L_{\mathtt{primary}}$ for our final prediction $P_{\mathtt{final}}$.
We add three different types of auxiliary cross-entropy losses to stabilize the training.
First, we use the generated SSR attention to fuse the auxiliary per-scale predictions from OCR, yielding $P^{\mathtt{aux}}_{\mathtt{ocr}}$ and the loss $L_{\mathtt{ocr}}$.
Second, we compute and sum up cross-entropy losses for each scale prediction $P_i$ yielding $L_{\mathtt{scale}}$.
Lastly, we add auxiliary segmentation heads inside AFA-DLA as in Fig.~\ref{fig:AFA-Seg}~(a) and have predictions for each scale. We fuse them with SSR across scales and get $P^{\mathtt{aux}}_j$, where $1\leq j\leq 4$. We compute the auxiliary loss for each and sum them up as $L_{\mathtt{aux}}$.
Accordingly, the total loss function is the weighted sum as
\begin{equation}
    L_{\mathtt{all}}\triangleq L_{\mathtt{primary}}+\beta_o L_{\mathtt{ocr}}+\beta_s L_{\mathtt{scale}}+\beta_a L_{\mathtt{aux}},
\end{equation}
where we set $\beta_o \triangleq 0.4$, $\beta_s \triangleq 0.05$ and $\beta_a \triangleq 0.05$.
We provide more details in the appendix.

\section{Experiments} \label{sec:exp}
We conduct experiments on several public datasets on both semantic segmentation and boundary detection tasks, and conduct a thorough analysis with a series of ablation studies.
Due to the space limit, we leave additional implementation details to our appendix.

\subsection{Results on Cityscapes}
The Cityscapes dataset~\cite{cordts2016cityscapes} provides high resolution (2048 x 1024) urban street scene images and their corresponding segmentation maps.
It contains 5K well annotated images for 19 classes and 20K coarsely labeled image as extra training data.
Its finely annotated images are split into 2975, 500, and 1525 for training, validation and testing.
We use DLA-X-102 as the backbone for AFA-DLA with a batch size of 8 and full crop size.
Following~\cite{tao2020hierarchical}, we train our model with auto-labeled coarse training data with 0.5 probability and otherwise use the fine labeled training set.
During inference, we use multi-scale inference with [0.5, 1.0, 1.5, 1.75, 2.0] scales, image flipping, and Seg-Fix~\cite{yuan2020segfix} post-processing.
We detail the effect of each post-processing technique in the appendix.

The results on the validation and test set are shown in Table~\ref{cityscapes-results}.
With only using ImageNet~\cite{deng2009imagenet} pre-training and without using external segmentation datasets, AFA-DLA obtains 85.14 mean IoU on the Cityscapes validation set, achieving the best performance compared to other methods in the same setting.
AFA-DLA outperforms the previous multi-scale attention methods and the recent methods using the Vision Transformer~\cite{dosovitskiy2020image} architecture.
On the Cityscapes test set, AFA-DLA also obtains competitive performance with a top performing method, DecoupleSegNet~\cite{li2020improving}, while using around 75\% fewer operations and parameters as shown in Table~\ref{cityscapes-flops}.

We additionally evaluate the application of AFA to other widely used segmentation models, including FCN, U-Net, and HRNet.
We build the baselines on our own and use the same shorter learning schedule and smaller training crop size for all models for fair comparison in Table~\ref{AFA-other-models}.
Since we only modify the aggregation operations of each model, we can still use the original ImageNet~\cite{deng2009imagenet} pre-training weights.

Combined with AFA, the segmentation models can each obtain at least 2.5\% improvement in mIoU, with only a small computational and parameter overhead.
In particular, we even lighten HRNet by replacing its concatenation in the last layer with our multiple feature fusion and still achieve 2.5\% improvement.
This demonstrates AFA as a lightweight module that can be readily applied to existing models for segmentation.

\begin{table}[t]
    \caption{Segmentation results on Cityscapes validation and testing sets. We only compare to published methods without using extra segmentation datasets. AFA-DLA achieves the best performance on the validation set and competitive performance with the top performing method on the test set.}
    \label{cityscapes-results}
    \vspace{0.05in}
    \centering
    \small
    \begin{tabular}{lcc}
        \toprule
        Method & mIoU (val) & mIoU (test) \\
        \midrule
        DLA~\cite{yu2018deep} & 75.10 & 75.90 \\
        SFNet~\cite{li2020semantic} & \textit{N/A} & 81.80 \\
        DeepLabV3+~\cite{chen2018encoder} & 79.55 & 82.10 \\
        DANet~\cite{fu2019dual} & 81.50 & 81.50 \\
        FPT~\cite{zhang2020feature} & 81.70 & 82.20 \\
        Gated-SCNN~\cite{takikawa2019gated} & 81.80 & 82.80 \\
        GFF~\cite{li2020gated} & 81.80 & 82.20 \\
        SETR~\cite{strudel2021segmenter} & 82.20 & 81.60 \\
        SegFormer~\cite{xie2021segformer} & 82.40 & 82.20 \\
        AlignSeg~\cite{huang2021alignseg} & 82.40 & 82.60 \\
        OCR~\cite{yuan2020object} & 82.40 & 83.00 \\
        DecoupleSegNets~\cite{li2020improving} & 83.50 & \textbf{83.70} \\
        Mask2Former~\cite{cheng2021masked} & 84.30 & \textit{N/A} \\
        \midrule
        AFA-DLA (Ours) & \textbf{85.14} & \emph{83.58} \\
        \bottomrule
    \end{tabular}
    \vspace{-0.12in}
\end{table}

\begin{table}[t]
    \caption{Resource usage of different models. AFA-DLA uses much fewer operations and parameters when compared to top performing methods.}
    \vspace{0.05in}
    \centering
    \small
    \begin{tabular}{lcc}
        \toprule
        Method & FLOPs (G) & Param. (M) \\
        \midrule
        DLA-X-102~\cite{yu2018deep} & 533 & 34.7 \\
        DeepLabV3+~\cite{chen2018encoder} & 2514 & 54.4 \\
        DecoupleSegNet~\cite{li2020improving} & 6197 & 138.4 \\
        \midrule
        AFA-DLA-X-102 (Ours) & 1333 & 36.3 \\
        \bottomrule
        \end{tabular}
    \label{cityscapes-flops}
    \vspace{-0.18in}
\end{table}

\begin{table}[t]
    \small
    \caption{Combining AFA with other widely used segmentation models on the Cityscapes validation set. With AFA, each model can obtain at least 2.5\% improvement in mIoU, with only a small computational and parameter overhead. The baselines are implemented on our own and all experiments are under fair comparison.}
    \vspace{0.05in}
    \begin{tabular}{l|cc|c|c}
        \toprule
        Method & FLOP (G) & Param. (M) & mIoU & $\Delta$ ($\%$) \\
        \midrule
        FCN & 1581.8 & 49.5 & 75.52 & - \\
        AFA-FCN & 1659.2 & 51.9 & \textbf{77.88} & 3.1 \\
        \midrule
        U-Net-S5-D16 & 1622.8 & 29.1 & 62.73 & - \\
        AFA-U-Net & 2146.7 & 29.4 & \textbf{64.42} & 2.7 \\
        \midrule
        HRNet-W48 & 748.7 & 65.9 & 78.48 & - \\
        AFA-HRNet & 701.4 & 65.4 & \textbf{80.41} & 2.5 \\
        \bottomrule
    \end{tabular}
    \label{AFA-other-models}
    \vspace{-0.12in}
\end{table}

\subsection{Results on BDD100K}
\begin{table}[t]
    \caption{Segmentation results on BDD100K validation and testing set. $\dagger$~denotes using Cityscapes data for pre-training. AFA-DLA achieves the new state-of-the-art performance on both sets.}
    \label{bdd100k-results}
    \vspace{0.05in}
    \centering
    \small
    \begin{tabular}{lcc}
        \toprule
        Method & mIoU (val) & mIoU (test) \\
        \midrule
        DLA~\cite{yu2018deep} & 57.84 & \textit{N/A} \\
        CCNet~\cite{huang2019ccnet} & 64.03 & 55.93 \\
        DNL~\cite{yin2020disentangled} & \textit{N/A} & 56.31 \\
        PSPNet~\cite{zhao2017pyramid} & \textit{N/A} & 56.32 \\
        DeepLabv3+~\cite{chen2018encoder} & 64.49 & 57.00 \\
        $\text{DecoupleSegNet}\dagger$~\cite{li2020improving} & 66.90 & \textit{N/A} \\
        \midrule
        AFA-DLA (Ours) & \textbf{67.46} & \textbf{58.47} \\
        \bottomrule
    \end{tabular}
    \vspace{-0.18in}
\end{table}

BDD100K~\cite{yu2020bdd100k} is a diverse driving video dataset for multitask learning.
For the semantic segmentation task, it provides 10K images with same categories as Cityscapes at 1280 x 720 resolution.
The dataset consists 7K, 1K, and 2K images for training, validation, and testing.
Considering the amount of training data is twice as Cityscapes, we use DLA-169 as the backbone with full image crop and 16 training batch size for 200 epochs.
During inference, we use multi-scale inference with [0.5, 1.0, 1.5, 2.0] scales and image flipping.

The results on validation and test sets are shown in Table~\ref{bdd100k-results}.
AFA-DLA achieves new state-of-the-art performances on both sets despite using fewer operations and parameters when compared to the top performing methods as shown in Table~\ref{cityscapes-flops}.
Our method achieves 67.46 mIoU on the validation set and is even higher than DecoupleSegNet~\cite{li2020improving}, which uses Cityscapes pre-trained weights.
Moreover, AFA-DLA obtains 58.47 mIoU on the test set, which outperforms all the strong official baselines.

\subsection{Boundary Detection}
We additionally conduct experiments on boundary detection, which involves predicting a binary segmentation mask indicating the existence of boundaries.
We evaluate on two standard boundary detection datasets, NYU Depth Dataset V2 (NYUDv2)~\cite{silberman2012indoor} and Berkeley Segmentation Data Set and Benchmarks 500 (BSDS500)~\cite{arbelaez2010contour}.
For each dataset, we follow the standard data preprocessing and evaluation protocol in literature \cite{xie2015holistically,liu2017richer}.
Specifically, we augment each dataset by randomly flipping, scaling, and rotating each image.
We evaluate using commonly used metrics, which are the F-measure at the Optimal Dataset Scale (ODS) and at the Optimal Image Scale (OIS).
Following~\cite{yu2018deep}, we also scale the boundary labels by 10 to account for the label imbalance.
For simplicity, we do not consider using multi-scale images during inference, so SSR is not used.

\begin{table}[t]
    \caption{Boundary detection results on NYUDv2 test set. AFA-DLA achieves the new state-of-the-art results.}
    \label{tab:nyud-test-results}
    \vspace{0.05in}
    \centering
    \small
    \begin{tabular}{lcc}
    \toprule
        Method & ODS & OIS \\
        \midrule
        PiDiNet~\cite{su2021pixel} & 0.756 & 0.773 \\
        BDCN~\cite{he2019bi} & 0.765 & 0.781 \\
        AMH-Net~\cite{liu2017richer} & 0.771 & 0.786 \\
        AFA-DLA (Ours) & \textbf{0.780} & \textbf{0.792} \\
        \bottomrule
    \end{tabular}
    \vspace{-0.10in}
\end{table}

\begin{table}[t]
    \caption{Boundary detection results on BSDS500 test set. AFA-DLA outperforms all other methods in ODS.}
    \label{bsds500-test-results}
    \vspace{0.05in}
    \centering
    \small
    \begin{tabular}{lcc}
        \toprule
        Method & ODS & OIS \\
        \midrule
        DLA~\cite{yu2018deep} & 0.803 & 0.813 \\
        LPCB~\cite{deng2018learning} & 0.800 & 0.816 \\
        BDCN~\cite{he2019bi} & 0.806 & \textbf{0.826} \\
        AFA-DLA (Ours) & \textbf{0.812} & \textbf{0.826} \\
        \bottomrule
    \end{tabular}
    \vspace{-0.20in}
\end{table}

\parsection{Results on NYUDv2.}
The NYUDv2 dataset contains both RGB and depth images.
There are 381 training images, 414 validation images, and 654 testing images.
We follow the same procedure as~\cite{xie2015holistically,liu2017richer,he2019bi} and train a separate model on RGB and HHA~\cite{gupta2014learning} images.
We evaluate using both RGB and HHA images as input by averaging each model's output during inference.
The results are shown in Table~\ref{tab:nyud-test-results}.
AFA-DLA outperforms all other methods by a large margin and achieves state-of-the-art performances.
In particular, when using both RGB and HHA as input, AFA-DLA can achieve a high score of 0.780 in ODS and 0.792 in OIS.

\parsection{Results on BSDS500.}
The BSDS500 dataset contains 200 training images, 100 validation images, and 200 testing images.
We follow standard practice~\cite{xie2015holistically} and only use boundaries annotated by three or more annotators for supervision.
We do not consider augmenting the training set with additional data, so we only utilize the available data in the BSDS500 dataset.
As in Table~\ref{bsds500-test-results}, AFA-DLA achieves superior performance when compared to methods only trained on the BSDS500 dataset and obtains 0.812 in ODS.
% In particular, AFA-DLA improves the performance of DLA by 0.09 in ODS and 0.013 in OIS, despite using a smaller backbone.

\begin{table}[t]
    \caption{Ablation study on Cityscapes validation set with DLA-34 as backbone. Aux. Head denotes using auxiliary heads, MFF denotes multiple feature fusion, SA and CA denote using spatial and channel attention for feature fusion, Swap denotes switching the input of spatial and channel attention modules, and Both stands for using both input features to generate attention for fusion.}
    \vspace{0.05in}
    \label{ablation-results}
        \centering
        \small
        \setlength{\tabcolsep}{5pt}
        \begin{tabular}{lcc|c|c}
        \toprule
        Binary Fusion & Aux. Head & MFF & MS Inference & mIoU \\
        \midrule
        Original & - & - & Single Scale & 74.43 \\
        Swap & - & - & Single Scale & 75.37 \\
        CA & - & - & Single Scale & 75.54 \\
        CBAM~\cite{woo2018cbam} & - & - & Single Scale & 75.70 \\
        Both & - & - & Single Scale & 75.77 \\
        SA + CA & - & - & Single Scale & \textbf{76.14} \\
        \midrule
        SA + CA & \checkmark & - & Single Scale & 76.45 \\
        SA + CA & \checkmark & \checkmark & Single Scale & \textbf{77.08} \\
        \midrule
        SA + CA & \checkmark & \checkmark & Avg. Pooling & 78.56 \\
        SA + CA & \checkmark & \checkmark & HMA~\cite{tao2020hierarchical} & 80.18 \\
        SA + CA & \checkmark & \checkmark & SSR & \textbf{80.74} \\
        \bottomrule
    \end{tabular}
    \vspace{-0.12in}
\end{table}

\subsection{Ablation Experiments}
In this section, we conduct several ablation studies on the Cityscapes validation set to validate each component of AFA-DLA.
The main baseline model we compare to is DLA~\cite{yu2018deep} with DLA-34 as backbone.
All the results are listed in Table~\ref{ablation-results}. We also provide visualizations in order to qualitatively evaluate our model.

\parsection{Binary Fusion.}
We first evaluate our attentive binary fusion module, which can learn the importance of each input signal during fusion.
Compared to using standard linear fusion operators, introducing nonlinearity and using channel attention (denoted CA) during binary fusion achieves around 1.1 mean IoU improvement.
This demonstrates that more expressive aggregation can drastically improve the results.
When we additionally use spatial attention (denoted SA + CA), we observe 0.6 points further improvement.

\parsection{Attention Mechanism.}
We validate the design of AFA by evaluating various other strategies for computing attention.
Switching the input of spatial and channel attention modules (denoted Swap) can lead to a minor improvement, but it is even worse than only using channel attention.
We also apply the CBAM~\cite{woo2018cbam} module on top of the original DLA linear aggregation nodes to refine the aggregated features as another baseline.
Finally, we concatenate both input features and use it to generate each attention (denoted Both), which requires much more computation.
On the contrary, our attentive binary fusion design can achieve the best performance.
This shows that the design of the aggregation node should consider the properties of the input features and AFA is the most effective.
% We validate our AFA design with taking opposite strategy and using both input features to generate each attention. We also apply the CBAM~\cite{woo2018cbam} module on top of the original DLA linear aggregation node to refine the aggregated feature as another baseline. As shown in Table~\ref{ablation-results}, switching the attention module inputs will have even worse performance than only using channel attention. If we generate the the spatial and channel attention by using both two features concatenated together, the performance also decrease to 75.77 mIoU and use more computation. Furthermore, while CBAM improves the original DLA by 1.27 mIoU, our attentive binary fusion further improves by 0.44 mIoU. The results show that the design of the aggregation node should consider the properties of the input features and AFA is the most effective.

\begin{figure*}[t]
    \centering
    \small
    \begin{tabular}{cccccc}
        \toprule
        \multirow{2}{*}{Input Image} & \multicolumn{2}{c}{Binary Fusion} & \multicolumn{3}{c}{Multiple Feature Fusion} \\
        \cmidrule{2-3} \cmidrule{4-6} & 
        $a_s$ for $F_s$
        &
        $a_s$ for $F_d$ 
        &
        $a_s$ for $F_1$
        &
        $a_s$ for $F_2$
        &
        $a_s$ for $F_3$ \\
        \includegraphics[width=0.14\linewidth]{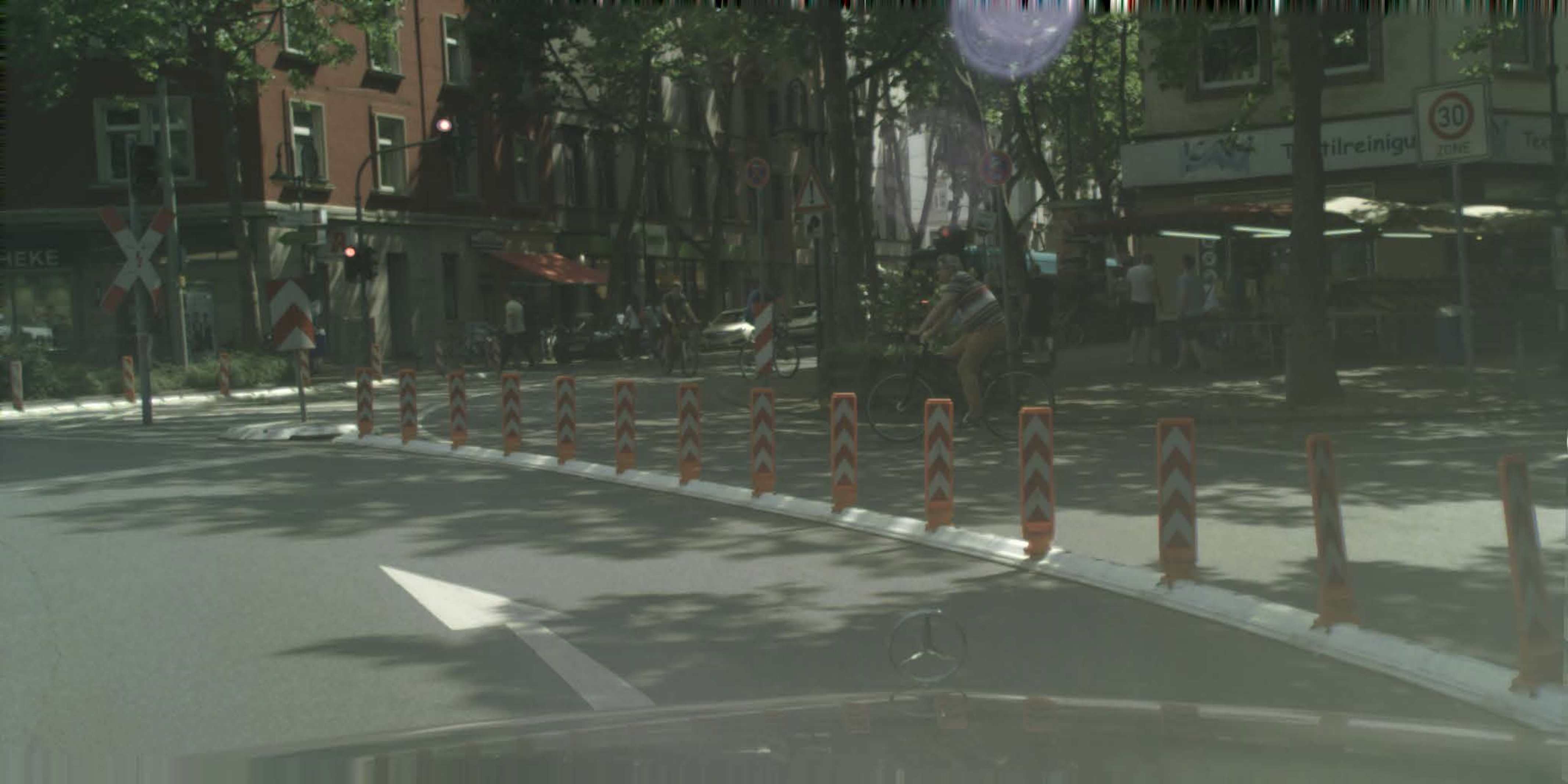}
        &
        \includegraphics[width=0.14\linewidth]{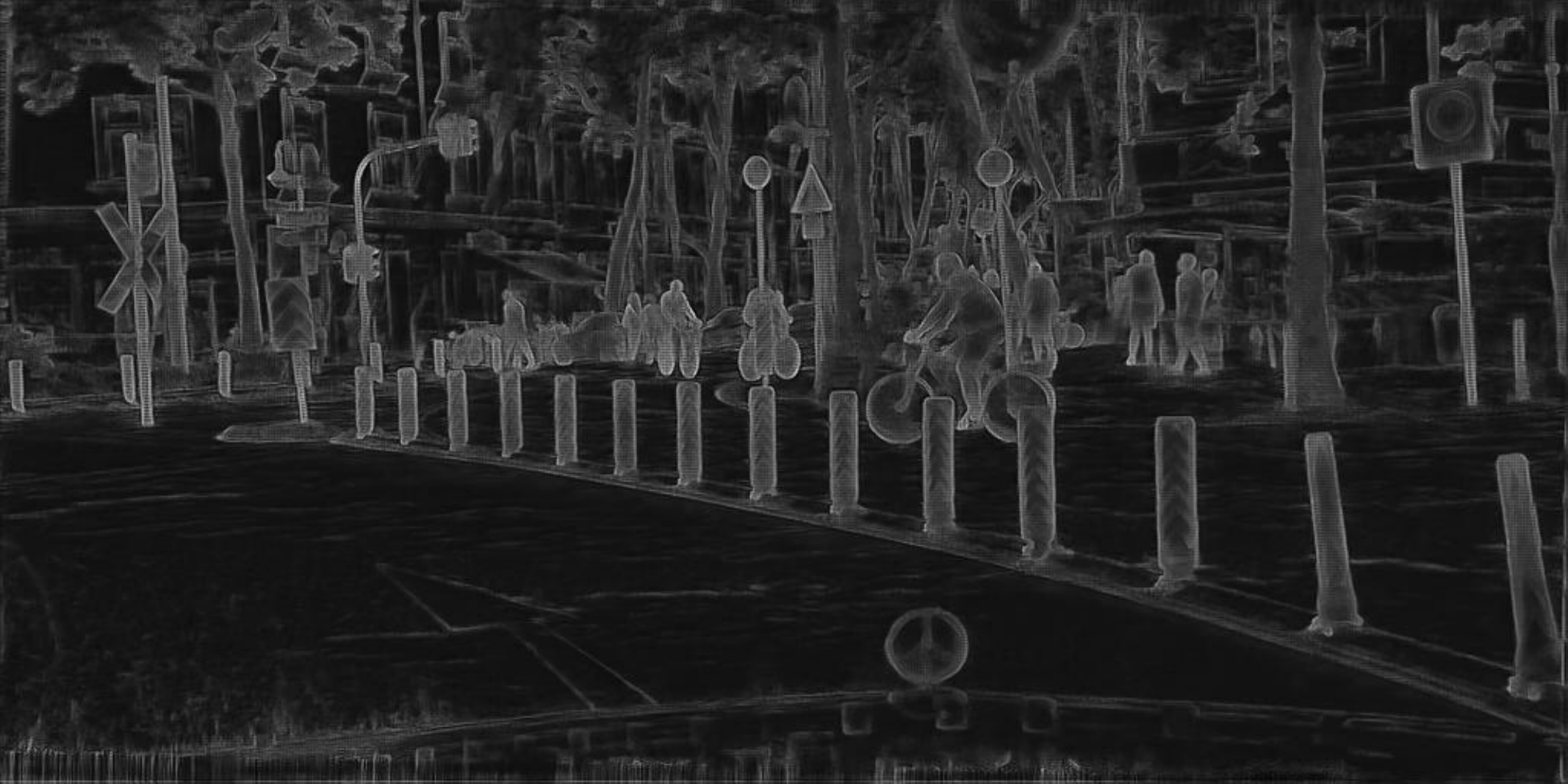}
        &
        \includegraphics[width=0.14\linewidth]{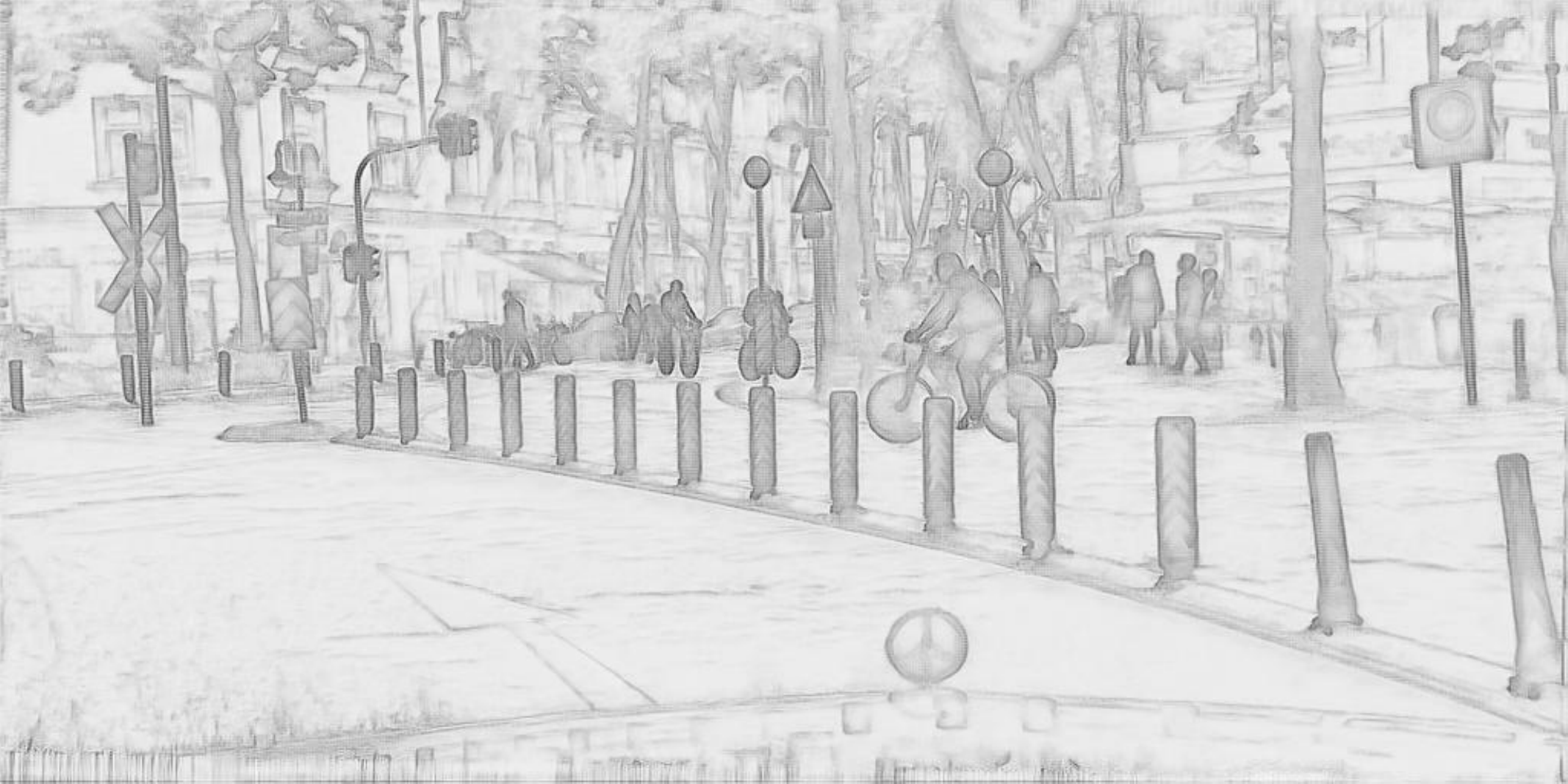}
        &
        \includegraphics[width=0.14\linewidth]{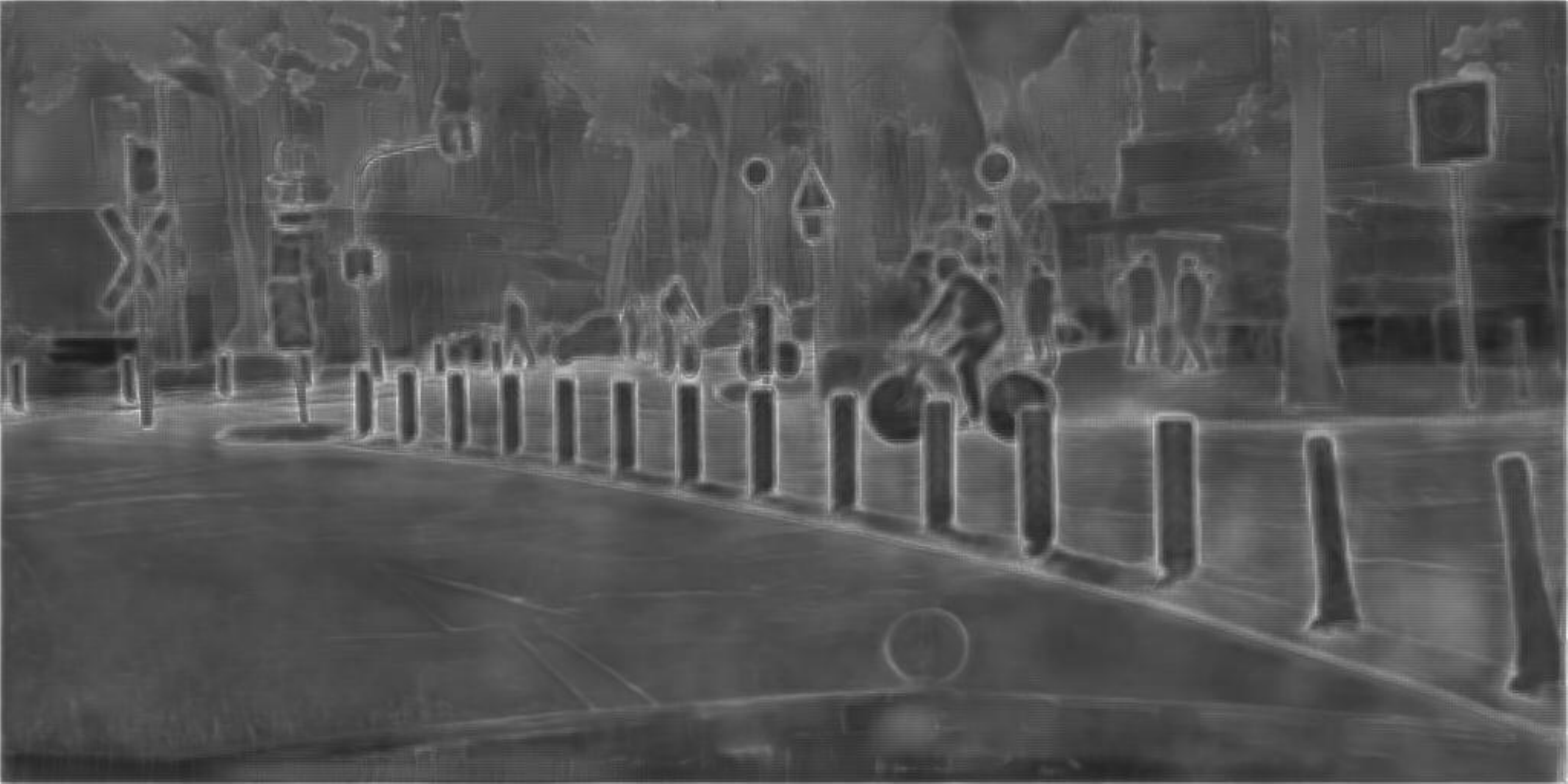}
        &
        \includegraphics[width=0.14\linewidth]{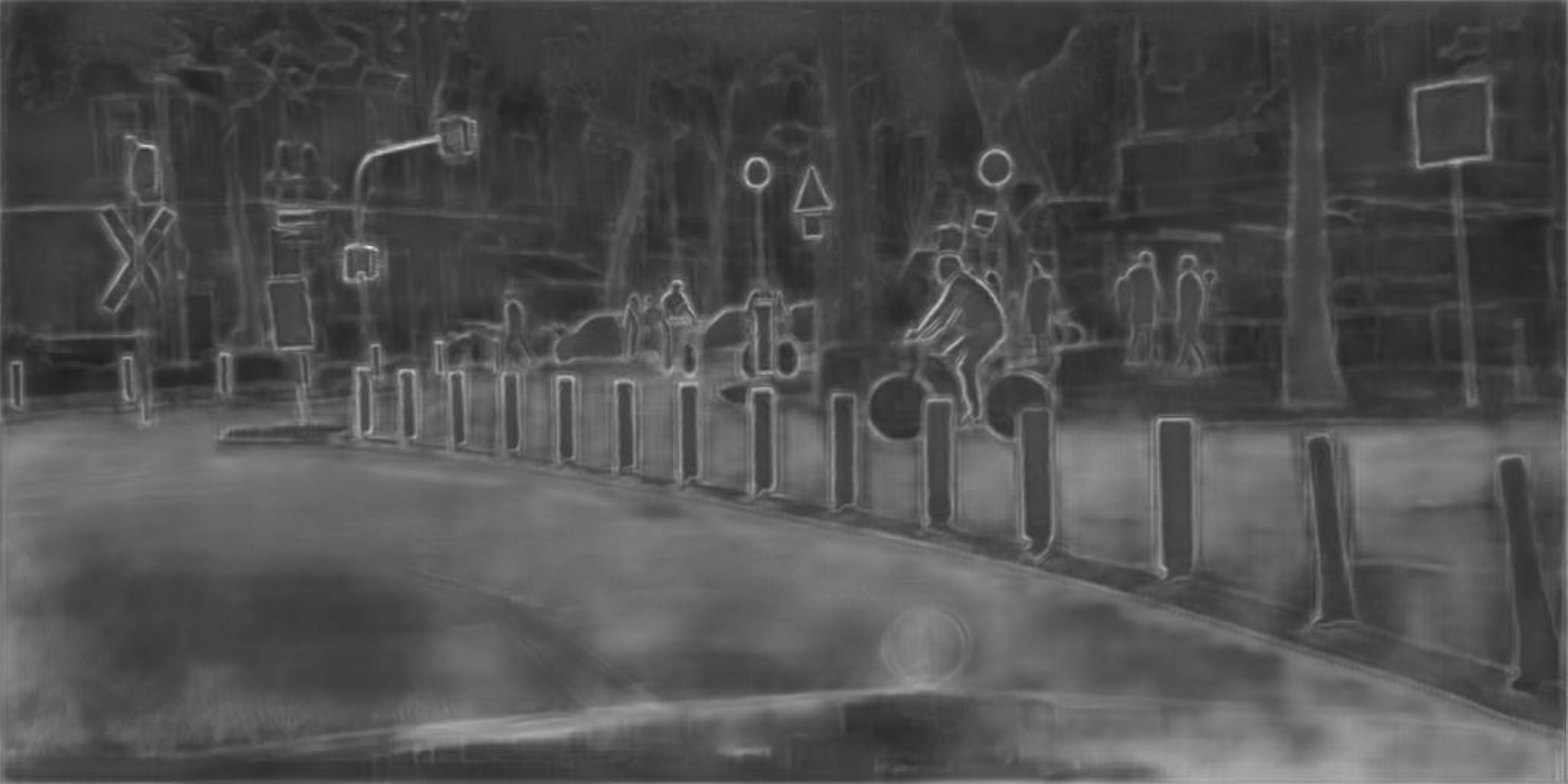}
        &
        \includegraphics[width=0.14\linewidth]{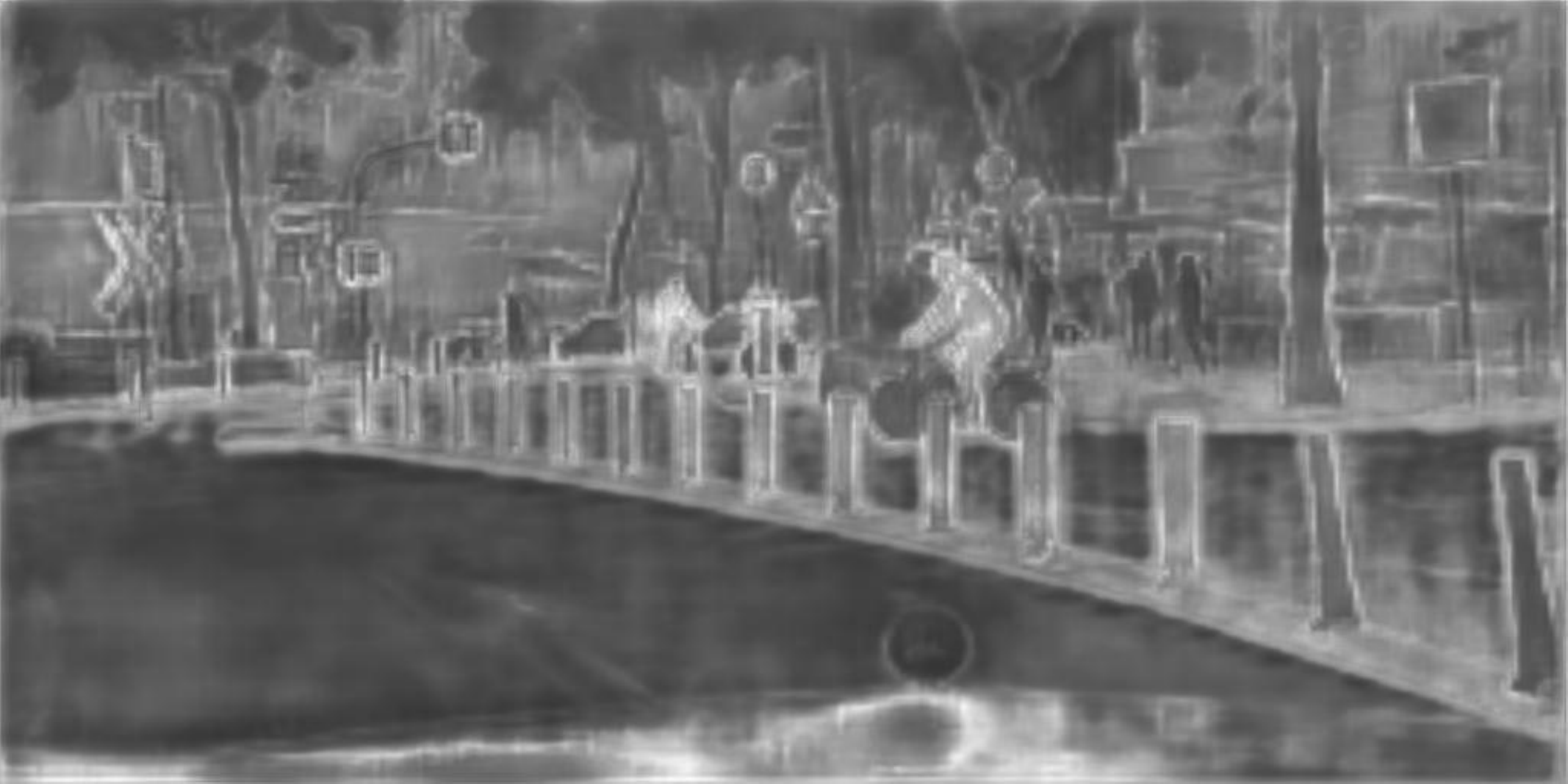} \\
        \includegraphics[width=0.14\linewidth]{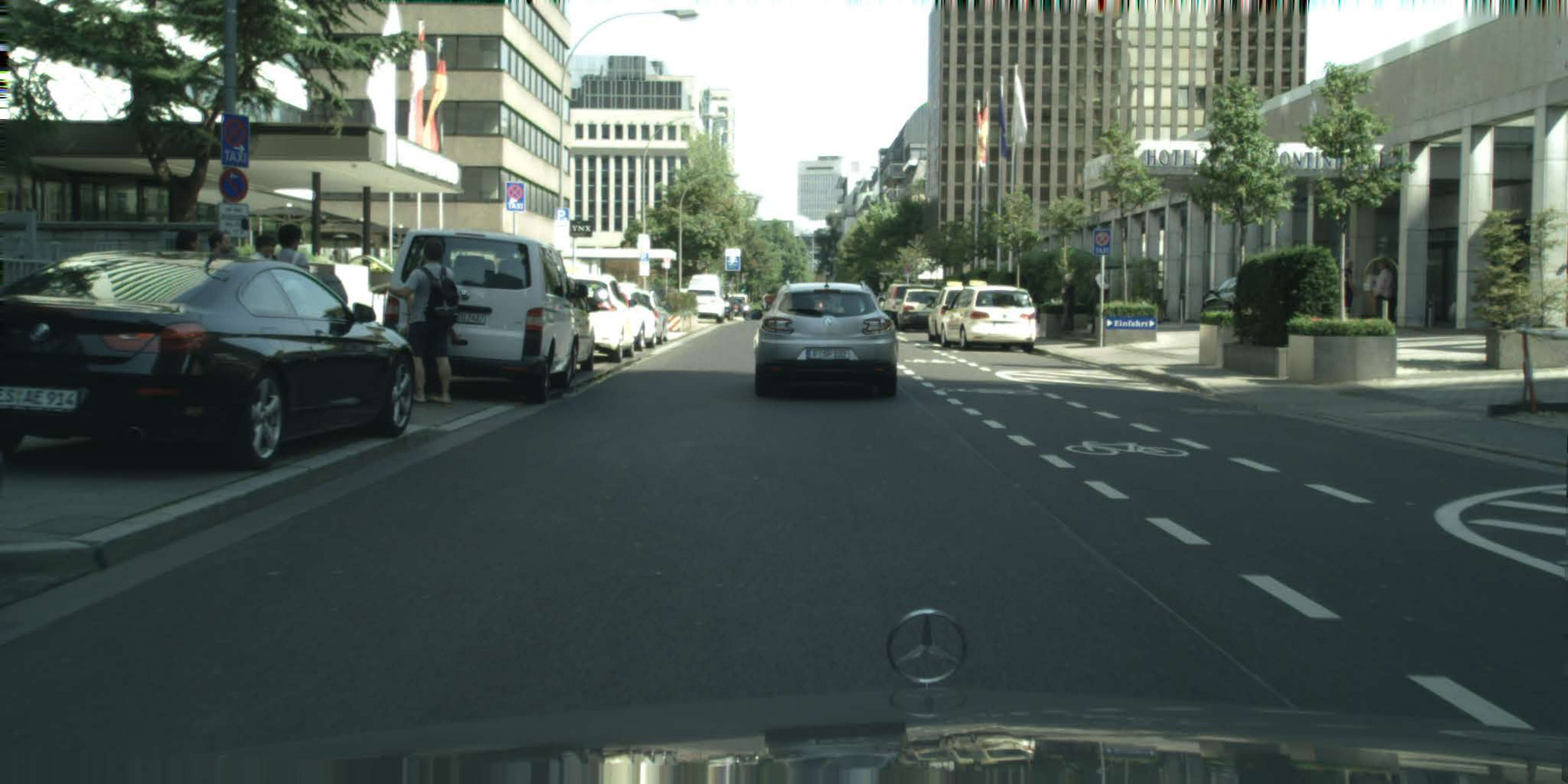}
        &
        \includegraphics[width=0.14\linewidth]{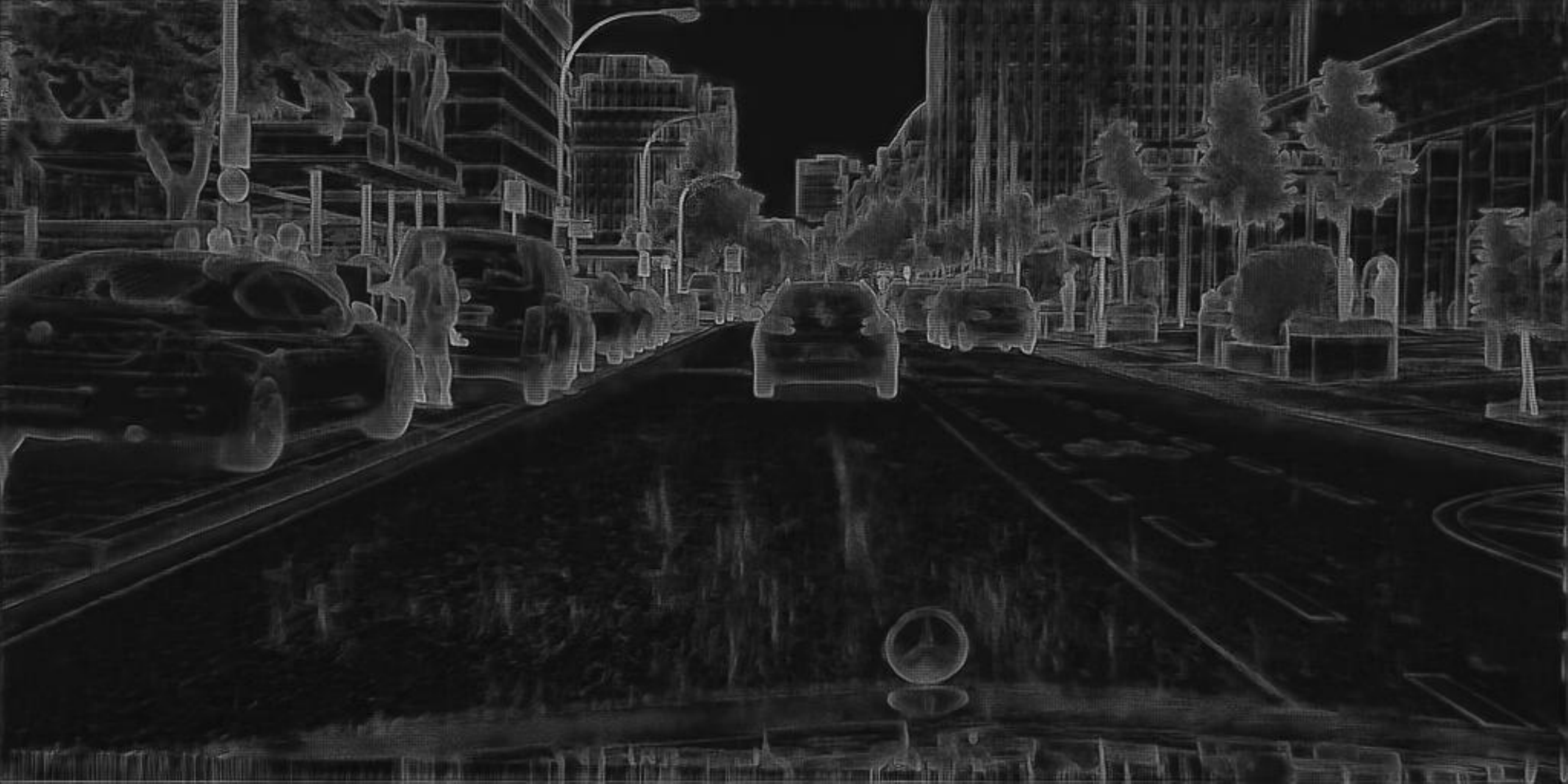}
        &
        \includegraphics[width=0.14\linewidth]{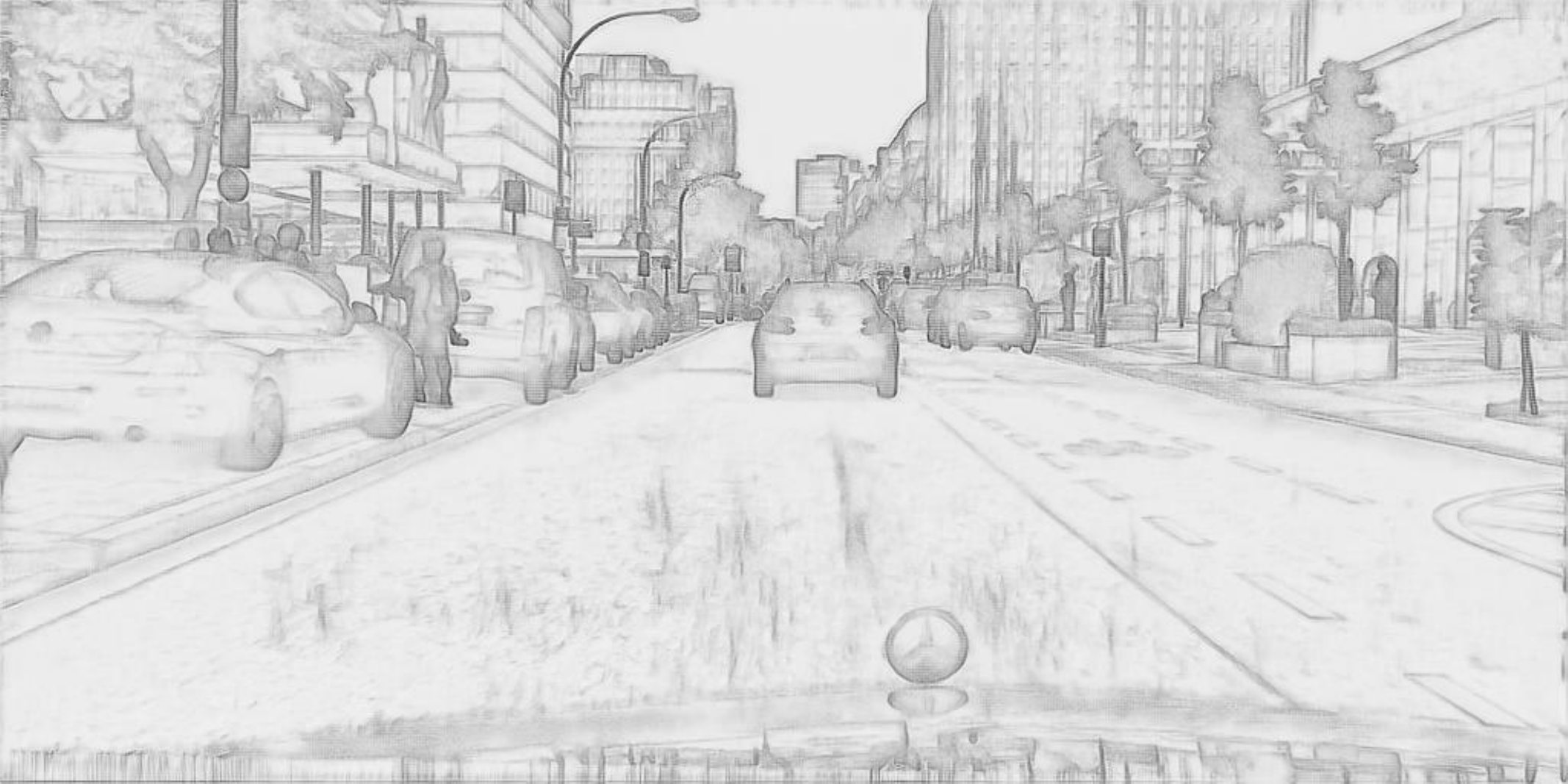}
        &
        \includegraphics[width=0.14\linewidth]{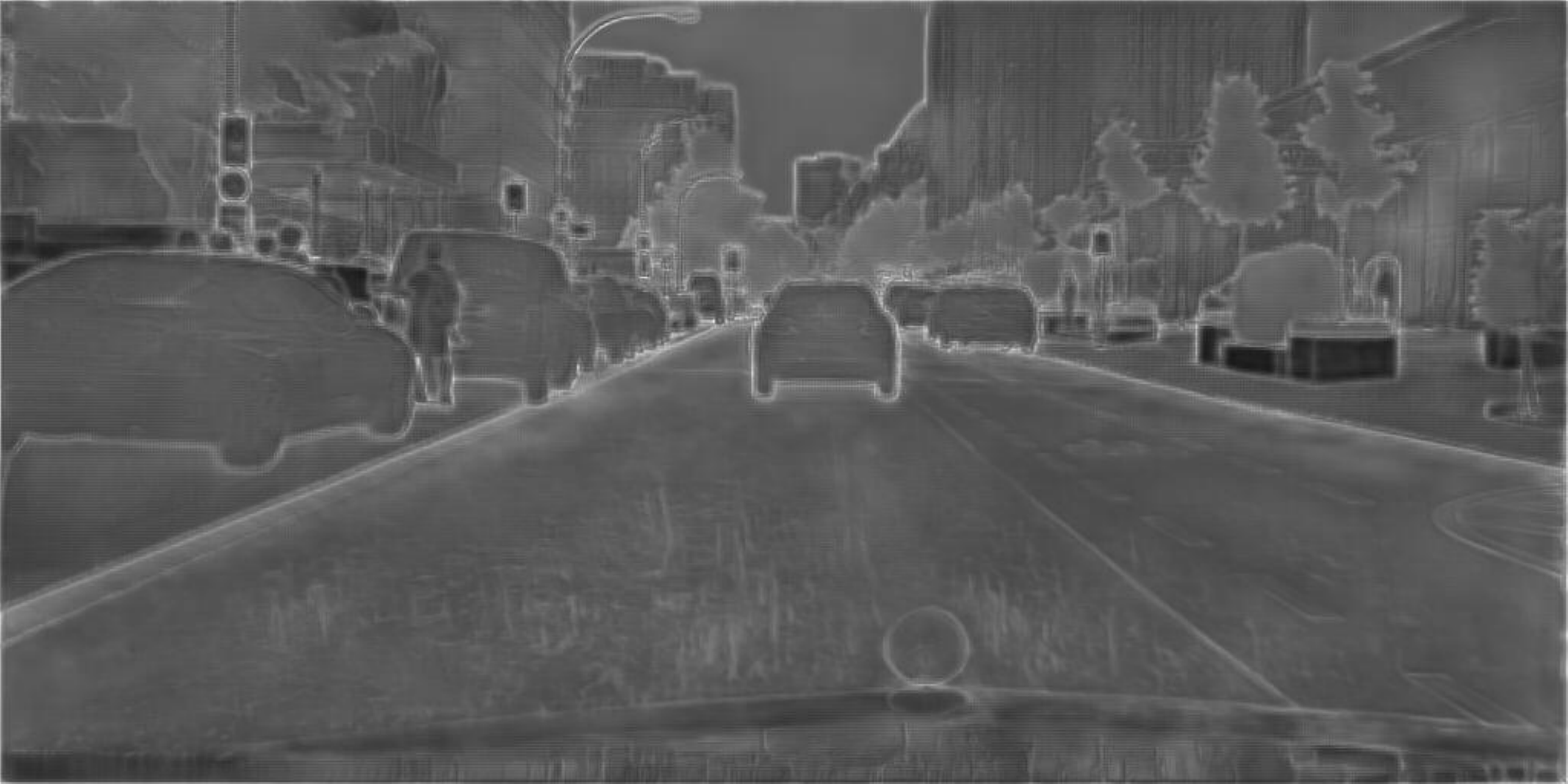}
        &
        \includegraphics[width=0.14\linewidth]{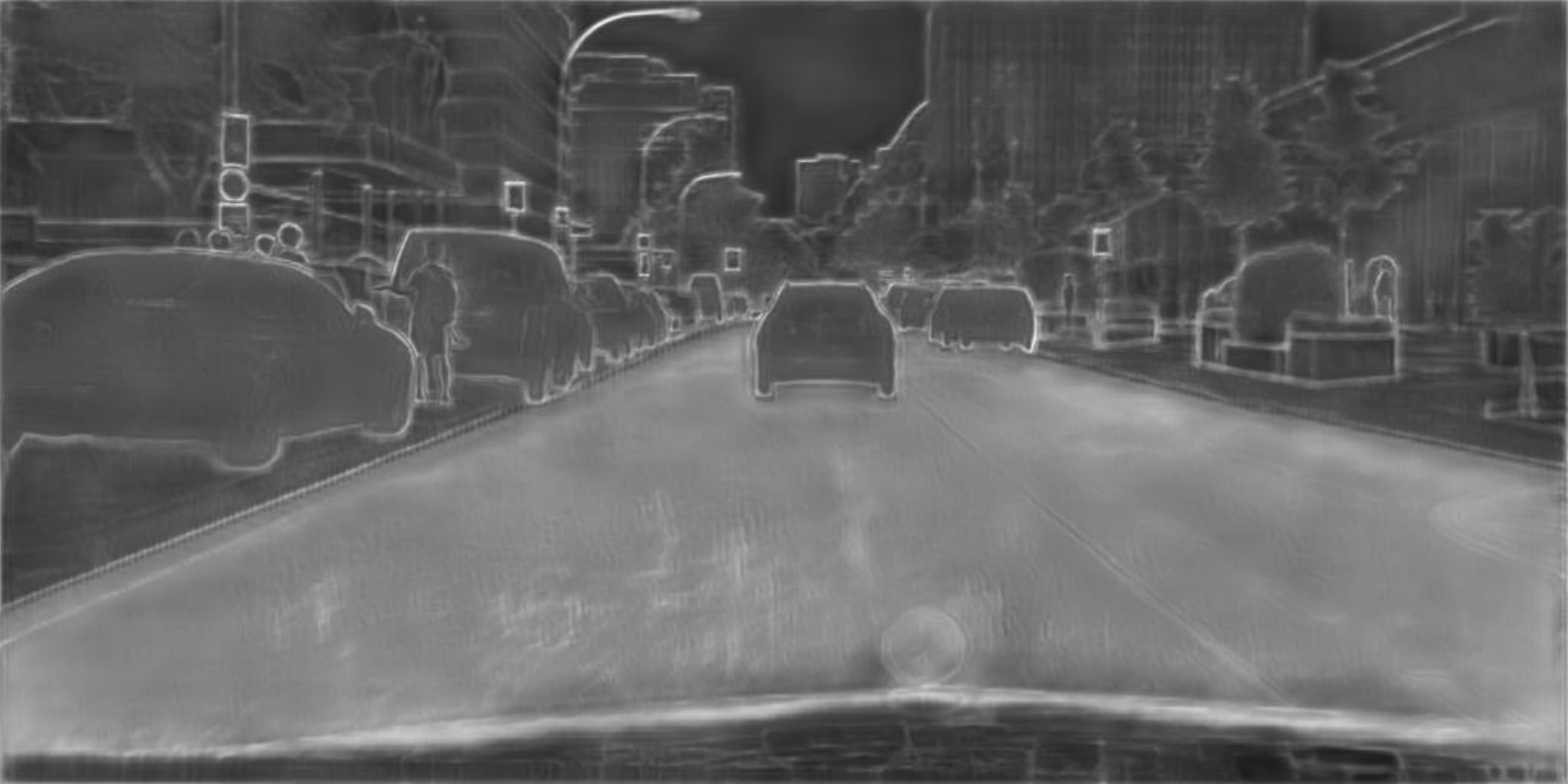}
        &
        \includegraphics[width=0.14\linewidth]{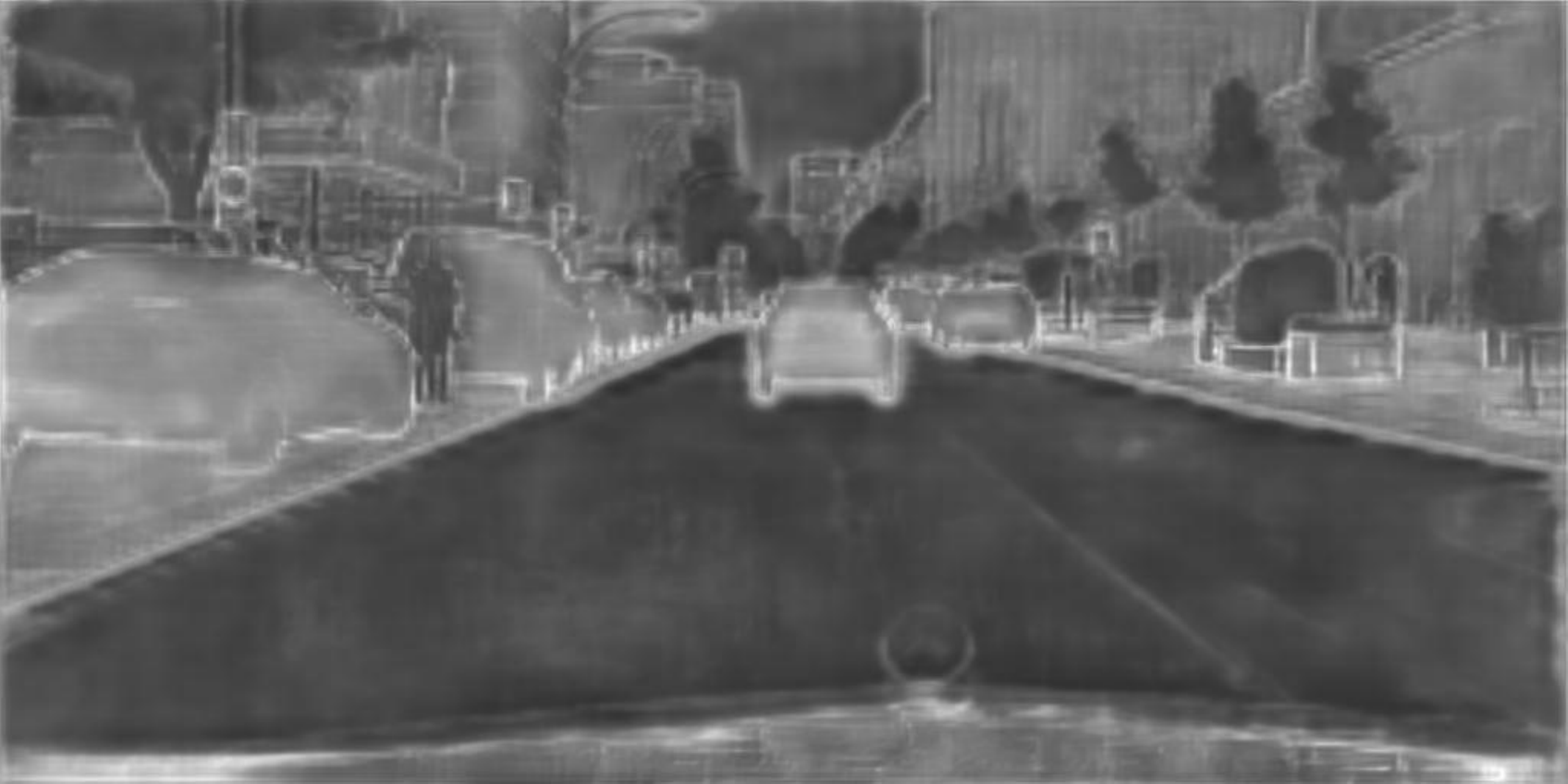} \\
        \bottomrule
    \end{tabular}
    \vspace{0.05in}
    \caption{Visualization of spatial attention maps $a_s$ generated by our attentive feature aggregation modules. Whiter regions denote higher attention. Compared to linear fusion operations, our AFA modules provide a more expressive way of combining features.}
    \label{fig:AFA_Attention}
    \vspace{-0.20in}
\end{figure*}

\begin{table}[t]
    \caption{Validation performance (mIoU) on Cityscapes between SSR and HMA across early training epochs. SSR achieves better performance over HMA across all epochs.}
    \vspace{0.05in}
    \small
    \centering
    \begin{tabular}{l|cccc}
        \toprule
        Method & epoch 1 & epoch 50 & epoch 100 & epoch 150 \\
        \midrule
        HMA~\cite{tao2020hierarchical} & 3.57 & 64.78 & 71.61 & 73.03 \\
        SSR & \textbf{5.49} & \textbf{68.16} & \textbf{72.76} & \textbf{74.48} \\
        \bottomrule
    \end{tabular}
    \label{tab:ssr_hma_comparison}
    \vspace{-0.20in}
\end{table}

\parsection{Auxiliary Segmentation Head.}
We add several auxiliary heads into AFA-DLA to stabilize the training, which is common practice among other popular baseline models.
The whole backbone can be supervised by the auxiliary losses efficiently.
We see about 0.3 mIoU improvement.

\parsection{Multiple Feature Fusion.}
We apply our multiple feature fusion to enable AFA-DLA to fully leverage intermediate features in the network.
This gives the network more flexibility in selecting relevant features for computing the final feature.
By adding the multiple feature fusion module, we gain another 0.6 mIoU.

\parsection{Scale-Space Rendering.}
We employ our SSR module to fuse multi-scale predictions.
After applying SSR with [0.25, 0.5, 1.0, 2.0] inference scales, we gain an impressive improvement of nearly 3.7\% mIoU over using only a single scale.
We also compare with different multi-scale inference approaches under the same training setting.
SSR gains 1.2 mIoU over standard average pooling and further outperforms hierarchical multi-scale attention~\cite{tao2020hierarchical} by nearly 0.6 mean IoU.
In terms of FLOPs, HMA uses around 1433G and SSR consumes 1420G, so SSR does not require additional computational resources.
Furthermore, we report validation performances of HMA and SSR at intermediate checkpoints in Table~\ref{tab:ssr_hma_comparison}.
The results suggest that our scale-space rendering attention can alleviate the gradient vanishing problem and boost the overall performance, while still retaining the flexibility for selecting different training and inference scales.
With both AFA and SSR, we improve the DLA baseline model performance by over 6.3 mIoU.

\parsection{Attention Visualization.}
To understand where our AFA fusion modules attends to, we visualize the generated attention maps for a set of input features in Fig.~\ref{fig:AFA_Attention}.
AFA learns to attend to different regions of the input features depending on the information they contain.
Binary fusion module focuses on object boundaries in shallower features $F_s$ and attends to the rest on deeper features $F_d$.
Our multiple feature fusion module can perform complex selection of features by attending to different regions for each feature level.
$F_1$ aggregates shallower features and thus the module attends to the boundaries, while the rest attend to objects or the background.
Compared to linear fusion operations, AFA provides a more expressive way of combining features.

\begin{figure}[t]
    \centering
    \small
    \setlength\tabcolsep{1.5pt}
    \begin{tabular}{cccc}
        \toprule
        Input & DLA & AFA & GT \\
        \midrule
        \includegraphics[width=0.18\linewidth]{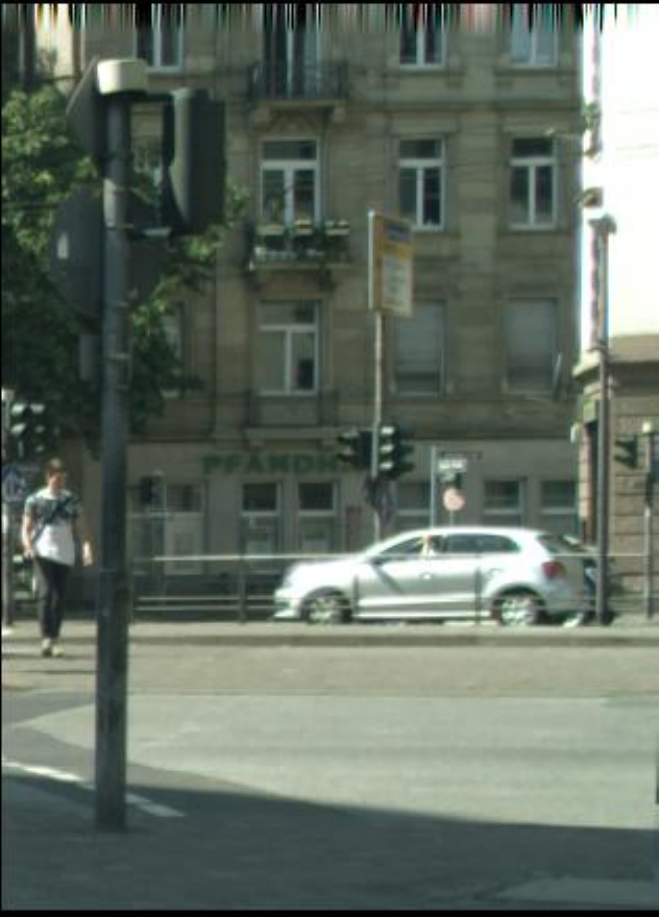}
        &
        \includegraphics[width=0.18\linewidth]{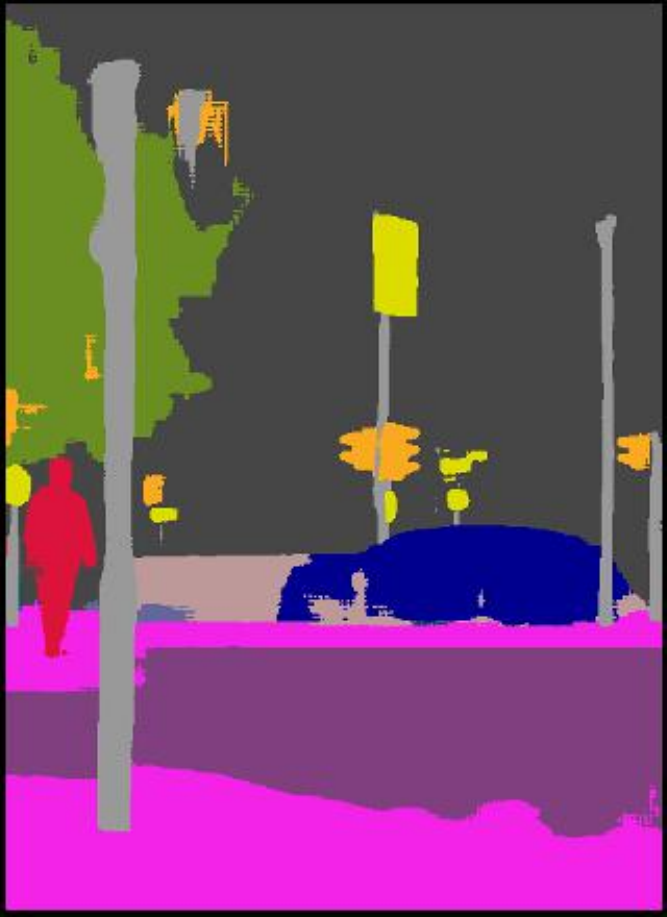}
        &
        \includegraphics[width=0.18\linewidth]{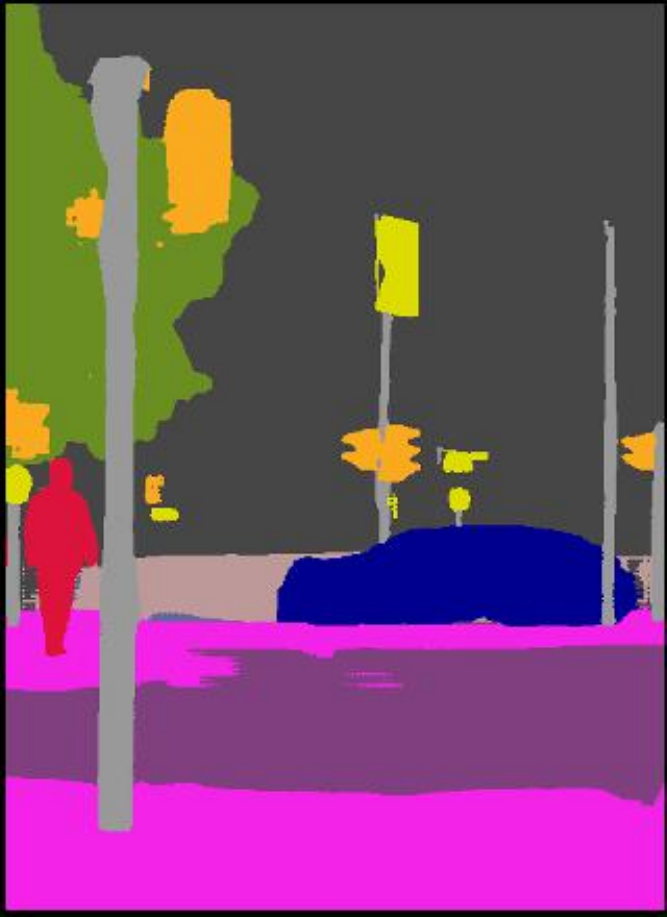}
        &
        \includegraphics[width=0.18\linewidth]{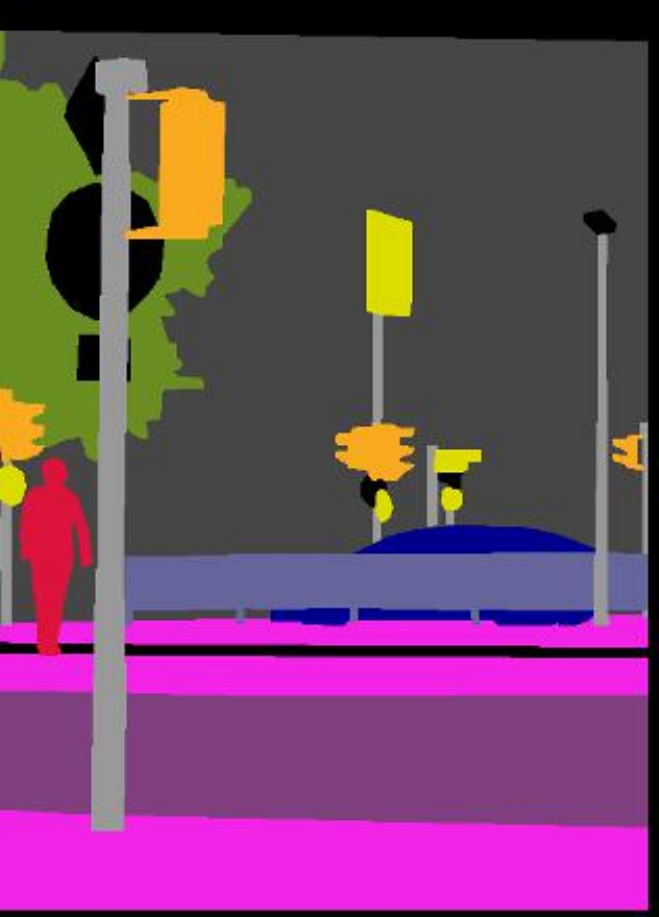} \\
        \includegraphics[width=0.18\linewidth]{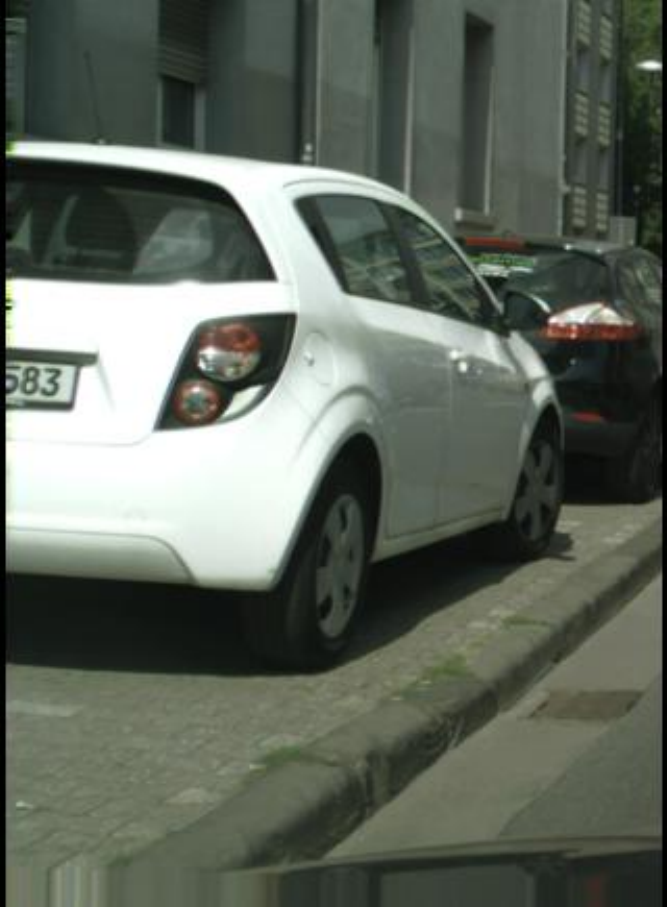}
        &
        \includegraphics[width=0.18\linewidth]{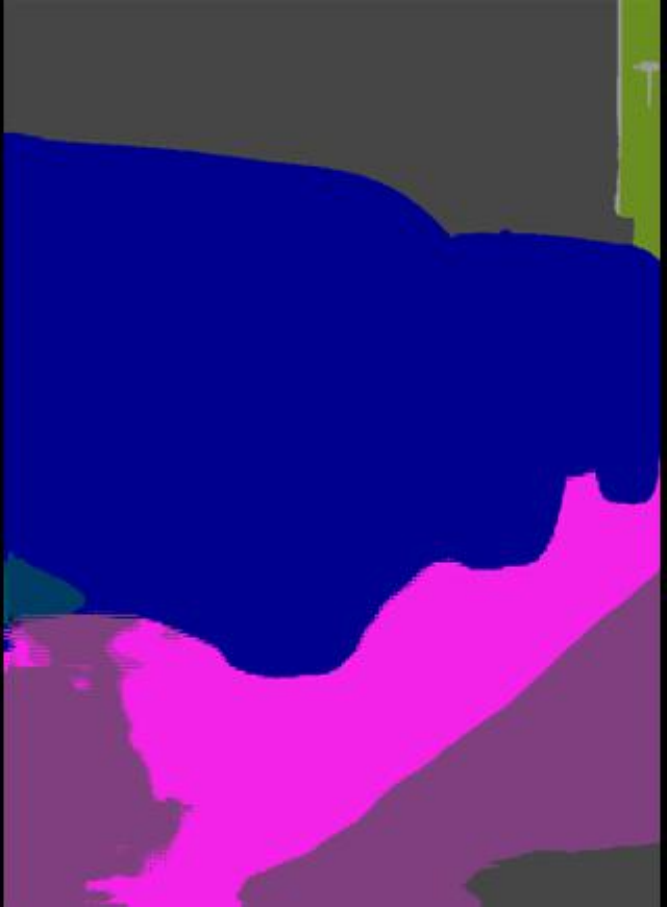}
        &
        \includegraphics[width=0.18\linewidth]{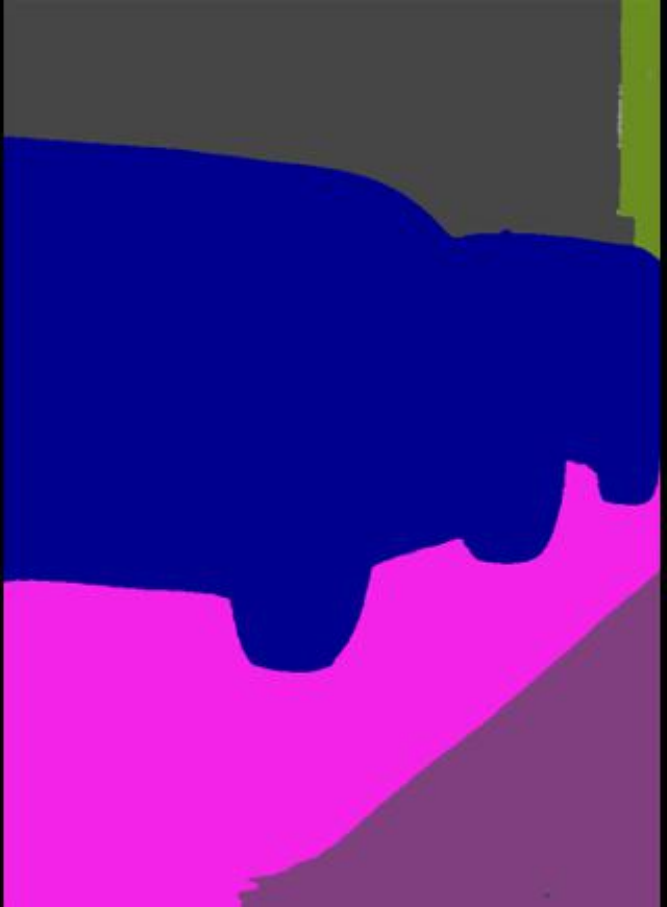}
        &
        \includegraphics[width=0.18\linewidth]{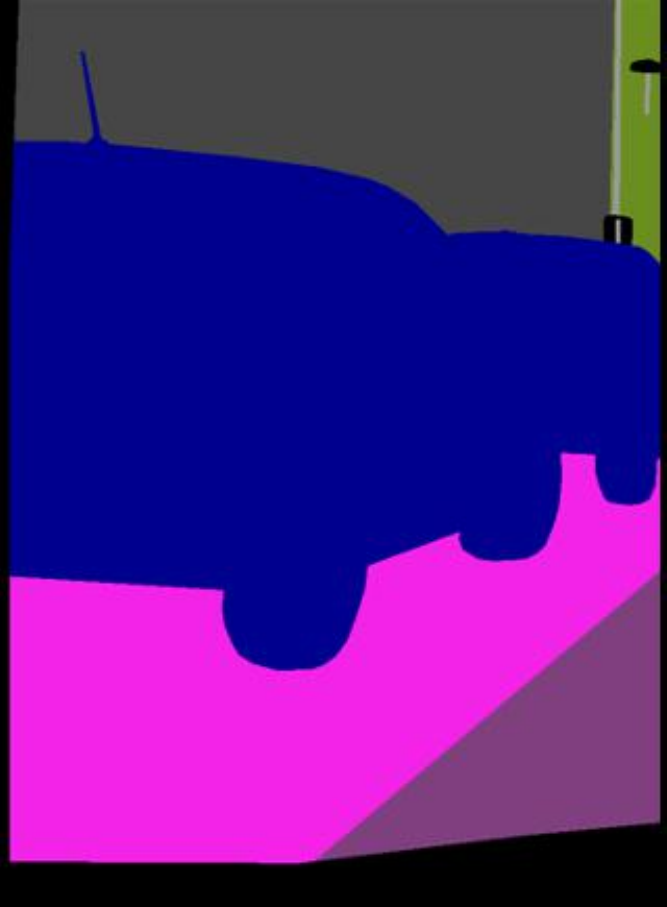} \\
        \bottomrule
    \end{tabular}
    \vspace{0.05in}
    \caption{ 
        Comparison of predictions generated by DLA and AFA-DLA. The black pixels are ignored. AFA-DLA can better distinguish object boundaries and correctly classify object classes.
    }
	\label{fig:DLA_AFA_comparison}
	\vspace{-0.20in}
\end{figure}

\parsection{Segmentation Visualization.}
We take a deeper look at the semantic segmentation results on the Cityscapes produced by AFA-DLA in Fig.~\ref{fig:DLA_AFA_comparison} and compare them to those produced by DLA~\cite{yu2018deep}.
With our AFA module, the model can better leverage spatial and channel information to better distinguish object boundaries and classify object classes.

\section{Conclusion}
We propose a novel attention-based feature aggregation module combined with a new multi-scale inference mechanism to build the competitive AFA-DLA model.
With spatial and channel attention mechanisms, AFA enlarges the receptive field and fuses different network layer features effectively.
% Our SSR attention has the benefit from separate multi-scale training and inference pipelines for saving computational resources while being more robust towards the gradient vanishing problem.
SSR improves existing multi-scale inference methods by being more robust towards the gradient vanishing problem.
Applying all of our components, we improve the DLA baseline model performance by nearly 6.3 mean IoU on Cityscapes.
% Our ablation study shows that the proposed AFA module learns how to fuse features much more effectively by being aware of what information should be preserved.
When combining AFA with existing segmentation models, we found consistent improvements of at least 2.5\% in mean IoU on Cityscapes, with only a small cost in computational and parameter overhead.
% For the boundary detection task, the AFA-DLA model obtains state-of-the-art results on BSDS500 and NYUDv2.
AFA-DLA also establishes new state-of-the-art results on BDD100K and achieves the new best score on Cityscapes when not using external segmentation datasets.
Moreover, for the boundary detection task, AFA-DLA obtains state-of-the-art results on NYUDv2 and BSDS500.

\section{Acknowledgment}
We gratefully acknowledge the support of computer time and facilities from Ministry of Science and Technology of Taiwan (MOST 110-2634-F-002-051).

{\small
\bibliographystyle{ieee_fullname}
\bibliography{egbib}
}

\clearpage

\appendix

This appendix provides additional results on boundary detection benchmarks, ablation study on post-processing effects on semantic segmentation, training settings and more qualitative results of the attention maps and output predictions.

\section{Full Results on Boundary Detection}
To complete the results in Table~5 and Table~6 in the main paper, we provide the full evaluation results on the BSDS500~\cite{arbelaez2010contour} and the NYUDv2~\cite{silberman2012indoor}.

In Table~\ref{tab:bsds500-sup-results}, we show more evaluation results of using the PASCAL VOC Context dataset (PVC)~\cite{mottaghi2014role} as additional training data and multi-scale inference for BSDS500.
When using PVC, we double our training epochs to account for the additional data.
For multi-scale inference, we use standard average pooling for fair comparison with other methods.
AFA-DLA achieves state-of-the-art results on single-scale inference when not training with additional data.
Surprisingly, using PVC does not further improve the results.
Nevertheless, AFA-DLA achieves the same performance with the state-of-the-art method BDCN~\cite{he2019bi} when using multi-scale inference.

In Table~\ref{tab:nyud-sup-results}, we report more evaluation results of using three different types of inputs.
Our AFA-DLA model outperforms all other methods by a large margin across
all three types of inputs, achieving state-of-the-art performances.
When only using RGB images as input, AFA-DLA already outperforms some other methods using both RGB and HHA images.

\section{Ablation Study on Post-processing of Semantic Segmentation}
To have a fair competition with other methods, we exploit several post-processing techniques to pursue higher performance.
We conduct an ablation study on how each technique affects the final performance on the Cityscapes~\cite{cordts2016cityscapes} validation set in Table~\ref{ablation-post-processing}.
The main improvement gains are from our Scale Space Rendering (SSR) for multi-scale inference, and the other techniques only bring minor improvements.

\begin{table}[t]
    \caption{Boundary detection results on BSDS500. PVC indicates training with additional PASCAL VOC Context dataset. MS indicates multi-scale inference. AFA-DLA achieves state-of-the-art results on single-scale images without using additional data, and competitive results when using both PVC and MS.
    }
    \vspace{0.05in}
    \centering
    \begin{tabular}{l|cc|cc}
        \toprule
        Method & PVC & MS & ODS & OIS \\
        \midrule
        Human & & & 0.803 & 0.803 \\ \midrule
        DLA~\cite{yu2018deep} &  &  & 0.803 & 0.813 \\
        LPCB~\cite{deng2018learning} &  &  & 0.800 & 0.816 \\
        BDCN~\cite{he2019bi} &  &  & 0.806 & \textbf{0.826} \\
        AFA-DLA (Ours) &  &  & \textbf{0.812} & \textbf{0.826} \\
        \midrule
        RCF~\cite{liu2017richer} & \checkmark &   & 0.808 & 0.825 \\
        LPCB~\cite{deng2018learning} & \checkmark &   & 0.808 & 0.824 \\
        BDCN~\cite{he2019bi} & \checkmark &   & \textbf{0.820} & \textbf{0.838} \\
        PiDiNet~\cite{su2021pixel} & \checkmark &   & 0.807 & 0.823 \\
        AFA-DLA (Ours) & \checkmark &  & 0.810 & 0.826 \\ \midrule
        RCF~\cite{liu2017richer} & \checkmark & \checkmark & 0.814 & 0.833 \\
        LPCB~\cite{deng2018learning} & \checkmark & \checkmark & 0.815 & 0.834 \\
        BDCN~\cite{he2019bi} & \checkmark & \checkmark  & 0.828 & 0.844 \\
        AFA-DLA (Ours) & \checkmark & \checkmark  & \textbf{0.828} & \textbf{0.844} \\
        \bottomrule
    \end{tabular}
    \label{tab:bsds500-sup-results}
\end{table}

\section{Training Losses}
In this section, we describe in more detail the formulation of our loss function for AFA-DLA for both semantic segmentation and boundary detection.

\begin{table}[t]
    \caption{Boundary detection results on NYUDv2 using three different types of inputs. AFA-DLA achieves state-of-the-art results across all three settings.}
    \vspace{0.05in}
    \centering
    \begin{tabular}{l|c|cc}
    \toprule
        Method & Input & ODS & OIS \\
        \midrule
        AMH-Net~\cite{liu2017richer} & \multirow{4}{*}{RGB} & 0.744 & 0.758 \\
        BDCN~\cite{he2019bi} &  & 0.748 & 0.763 \\
        PiDiNet~\cite{su2021pixel} &  & 0.733 & 0.747 \\
        AFA-DLA (Ours) &  & \textbf{0.762} & \textbf{0.775} \\ \midrule
        AMH-Net~\cite{liu2017richer} & \multirow{4}{*}{HHA} & 0.716 & 0.729 \\
        BDCN~\cite{he2019bi} &  & 0.707 & 0.719 \\
        PiDiNet~\cite{su2021pixel} &  & 0.715 & 0.728 \\
        AFA-DLA (Ours) &  & \textbf{0.718} & \textbf{0.730} \\ \midrule
        AMH-Net~\cite{liu2017richer} & \multirow{4}{*}{RGB+HHA} & 0.771 & 0.786 \\
        BDCN~\cite{he2019bi} &  & 0.765 & 0.781 \\
        PiDiNet~\cite{su2021pixel} &  & 0.756 & 0.773 \\
        AFA-DLA (Ours) &  & \textbf{0.780} & \textbf{0.792} \\
        \bottomrule
    \end{tabular}
    \label{tab:nyud-sup-results}
\end{table}

\subsection{Semantic Segmentation}
We use $k$ scales for training and RMI~\cite{zhao2019region} to be the primary loss for our final prediction $P_{\mathtt{final}}$, i.e.,
\begin{equation}
    L_{\mathtt{primary}}\triangleq L_{\mathtt{rmi}}(\hat P, P_{\mathtt{final}})\,,
\end{equation}

where $\hat P$ is the ground truth and $L_\mathtt{rmi}$ is the RMI loss function.
The first auxiliary cross-entropy loss is computed by using the generated scale-space rendering (SSR) attention to fuse the auxiliary per-scale predictions from the OCR~\cite{yuan2020object} module, yielding
\begin{equation}
L_{\mathtt{ocr}}\triangleq L_{\mathtt{ce}}(\hat P, P^{\mathtt{aux}}_{\mathtt{ocr}})\,,
\end{equation}
where $L_\mathtt{ce}$ denotes the cross-entropy loss.
For the second auxiliary loss, we compute and sum up cross-entropy losses for each scale prediction $P_i$, where $1\leq i\leq k$, yielding
\begin{equation}
    L_{\mathtt{scale}}\triangleq \sum_{i=1}^k L_{\mathtt{ce}}(\hat P, P_i)\,.
\end{equation}
Lastly, for the auxiliary loss inside AFA-DLA, we fuse the predictions of each auxiliary segmentation head with SSR across scales and get $P^{\mathtt{aux}}_j$, where $1\leq j\leq 4$. We compute the auxiliary loss for each prediction and sum them up as
\begin{equation}
    L_{\mathtt{aux}}\triangleq\sum_{j=1}^4 L_{\mathtt{ce}}(\hat P, P^{\mathtt{aux}}_j))\,.
\end{equation}
Accordingly, the total loss function is the weighted sum as
\begin{equation}
    L_{\mathtt{seg}}\triangleq L_{\mathtt{primary}}+\beta_o L_{\mathtt{ocr}}+\beta_s L_{\mathtt{scale}}+\beta_a L_{\mathtt{aux}},
\end{equation}
where we set $\beta_o = 0.4$, $\beta_s = 0.05$, and $\beta_a = 0.05$.

\subsection{Boundary Detection}
For boundary detection, we opted to using a simpler version of the loss function for semantic segmentation.
We use standard binary cross entropy (BCE) to be the primary loss for our final prediction $P_{\mathtt{final}}$, i.e.,
\begin{equation}
    L_{\mathtt{primary}}\triangleq L_{\mathtt{bce}}(\hat P, P_{\mathtt{final}})\,,
\end{equation}
where $\hat P$ is the ground truth and $L_\mathtt{bce}$ is the BCE loss function.

We also use auxiliary segmentation heads to make predictions at each feature level.
Each prediction $P^{\mathtt{aux}}_j$, where $1\leq j\leq 4$, is upsampled to the original scale and the BCE loss is used to compute the auxiliary loss, i.e.,
\begin{equation}
    L_{\mathtt{aux}}\triangleq\sum_{j=1}^4 L_{\mathtt{bce}}(\hat P, P^{\mathtt{aux}}_j))\,.
\end{equation}
Accordingly, the total loss function is the weighted sum as
\begin{equation}
    L_{\mathtt{bd}}\triangleq L_{\mathtt{primary}}+\beta_a L_{\mathtt{aux}},
\end{equation}
where we set $\beta_a = 0.05$.

\begin{table}[t]
  \caption{Ablation study on Cityscapes validation set with AFA-DLA-X-102 for validating each post-processing technique. SSR indicates our Scale Space Rendering.}
  \label{ablation-post-processing}
  \centering
  \setlength\tabcolsep{15pt}
  \begin{tabular}{ccc|c}
    \toprule
    SSR & Flip & Seg-Fix~\cite{yuan2020segfix} & mIoU \\
    \midrule
    - & - & - & 83.06 \\
    \checkmark & - & - & 84.81 \\
    \checkmark & \checkmark & - &  85.00 \\
    \checkmark & \checkmark & \checkmark & 85.14 \\
    \bottomrule
  \end{tabular}
\end{table}

\begin{table}[t]
    \caption{Specific training settings for each dataset. BS stands for training batch size.}
    \label{training-setting}
    \centering
    \begin{tabular}{lccc}
        \toprule
        Dataset & Crop Size & BS & Training Epochs \\
        \midrule
        Cityscapes & $2048 \times 1024$ & 8 & 375 \\
        % Mapillary Vistas & $832 \times 832$ & 16 & 200 \\
        BDD100K & $1280 \times 720$ & 16 & 200 \\
        BSDS500 & $416 \times 416$ & 16 & 10 \\
        NYUDv2 & $480 \times 480$ & 16 & 54 \\
        \bottomrule
    \end{tabular}
\end{table}

\section{Implementation Details}
We provide the general training setting and procedure used for training on Cityscapes~\cite{cordts2016cityscapes}, BDD100K~\cite{yu2020bdd100k}, BSDS500~\cite{arbelaez2010contour}, and NYUDv2~\cite{silberman2012indoor}.

We use PyTorch~\cite{paszke2019pytorch} as our framework and develop based on the NVIDIA semantic segmentation codebase~\footnote[1]{NVIDIA license: \url{https://github.com/NVIDIA/semantic-segmentation/blob/main/LICENSE}}. The general training procedure is using SGD~\cite{robbins1951stochastic} with momentum of 0.9 and weight decay of $10^{-4}$.
Specific settings for each dataset are shown in Table~\ref{training-setting}.

\subsection{Semantic Segmentation}
We use an initial learning rate of $1.0 \times 10^{-2}$.
We use the learning rate warm-up over the initial 1K training iterations and the polynomial decay schedule, which decays the initial learning rate by multiplying $(1 - \frac{\mathtt{epoch}}{\mathtt{max\_epochs}})^{0.9}$ every epoch.
We apply random image horizontal flipping, randomly rotating within 10 degrees, random scales from 0.5 to 2.0, random image color augmentation, and random cropping.
As in \cite{zhu2019improving}, we also use class uniform sampling in the data loader to overcome the data class distribution unbalance problem.
Due to limitations in computational power, we further use Inplace-ABN~\cite{bulo2018place} to replace the batch norm and ReLU function to acquire the largest possible training crop size and batch size on 8 Tesla V-100 32G GPUs.

\subsection{Boundary Detection}
We use an initial learning rate of $1.0 \times 10^{-2}$ and a batch size of 16 for both BSDS500~\cite{arbelaez2010contour} and NYUDv2~\cite{silberman2012indoor}.
We use the step decay schedule and drop the learning rate by 10 times at around $0.55\times\mathtt{max\_epochs}$ and then again at $0.85\times\mathtt{max\_epochs}$.
For augmentation, we follow the standard protocol in literature~\cite{xie2015holistically,liu2017richer} and apply random flipping, scaling by 0.5 and 1.5, and rotation by 16 different angles.
We train all our models on a single GeForce RTX 2080Ti GPU.

\begin{figure}[t]
	\centering
	\includegraphics[width=0.47\textwidth]{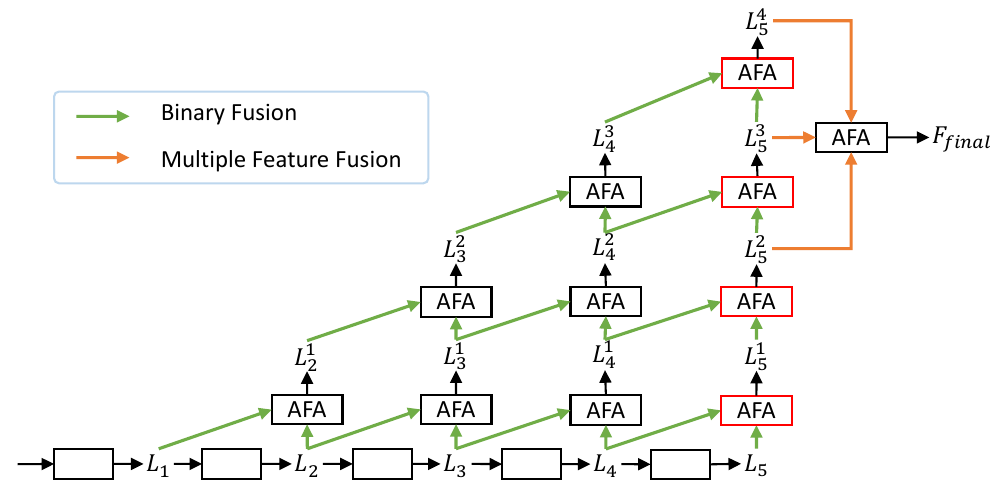}
	\caption{ 
        Notation of aggregated features of AFA-DLA.
	}
	\label{fig:AFA-DLA_detail}
\end{figure}

\section{Visualization of Attention Maps}
In this section, we provide more visualizations of attention maps generated by our proposed AFA module.
For reference, we provide a detailed architecture of AFA-DLA and denote the notation of aggregated features of different levels in Figure~\ref{fig:AFA-DLA_detail}.

\begin{figure*}[t]
    \centering
    \begin{tabular}{cc}
        \toprule
        \includegraphics[width=0.4\linewidth]{images/AFA/input_1.pdf}
        &
        \includegraphics[width=0.4\linewidth]{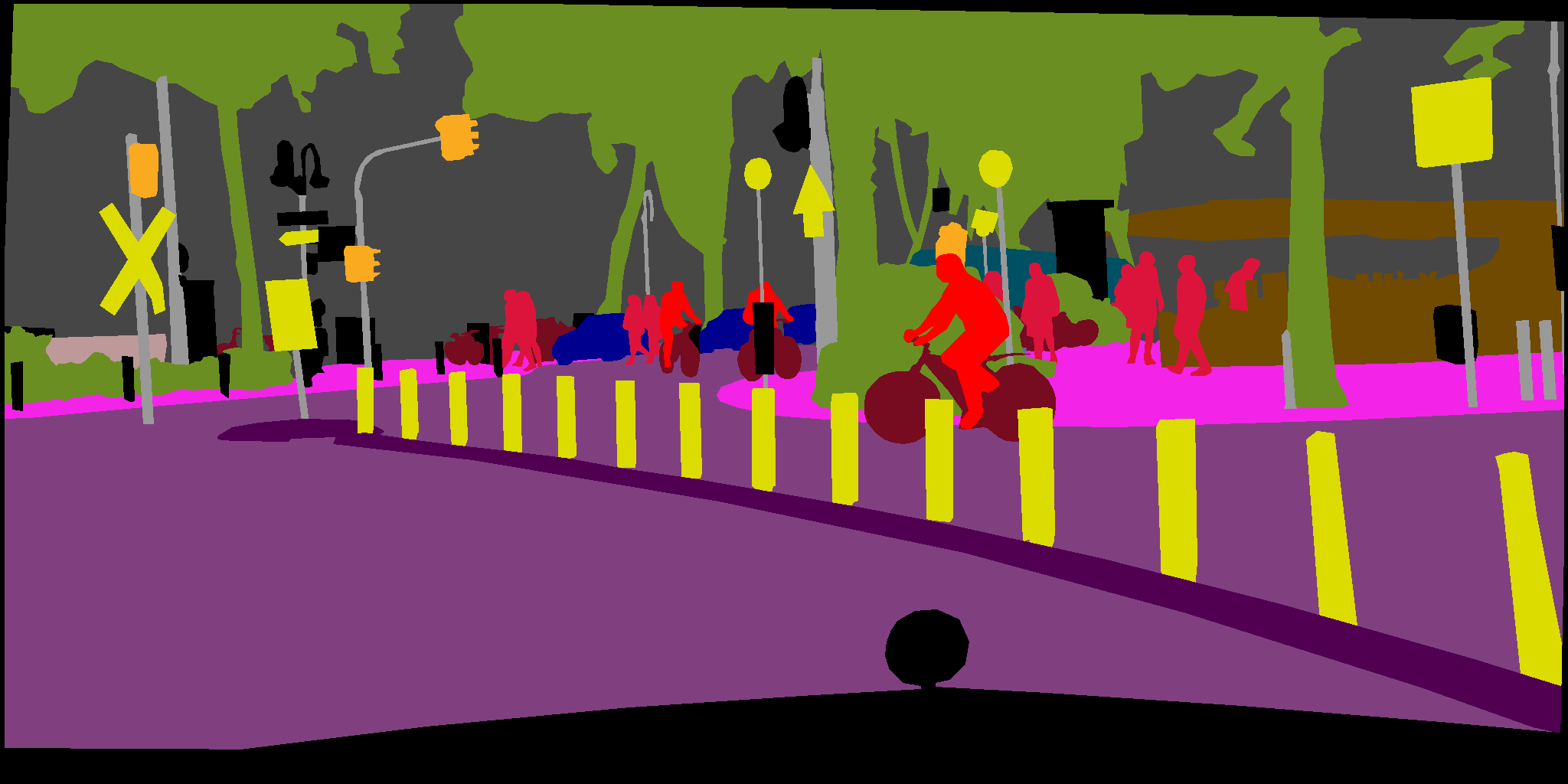} \\
        Input
        &
        Ground Truth \\
        \includegraphics[width=0.4\linewidth]{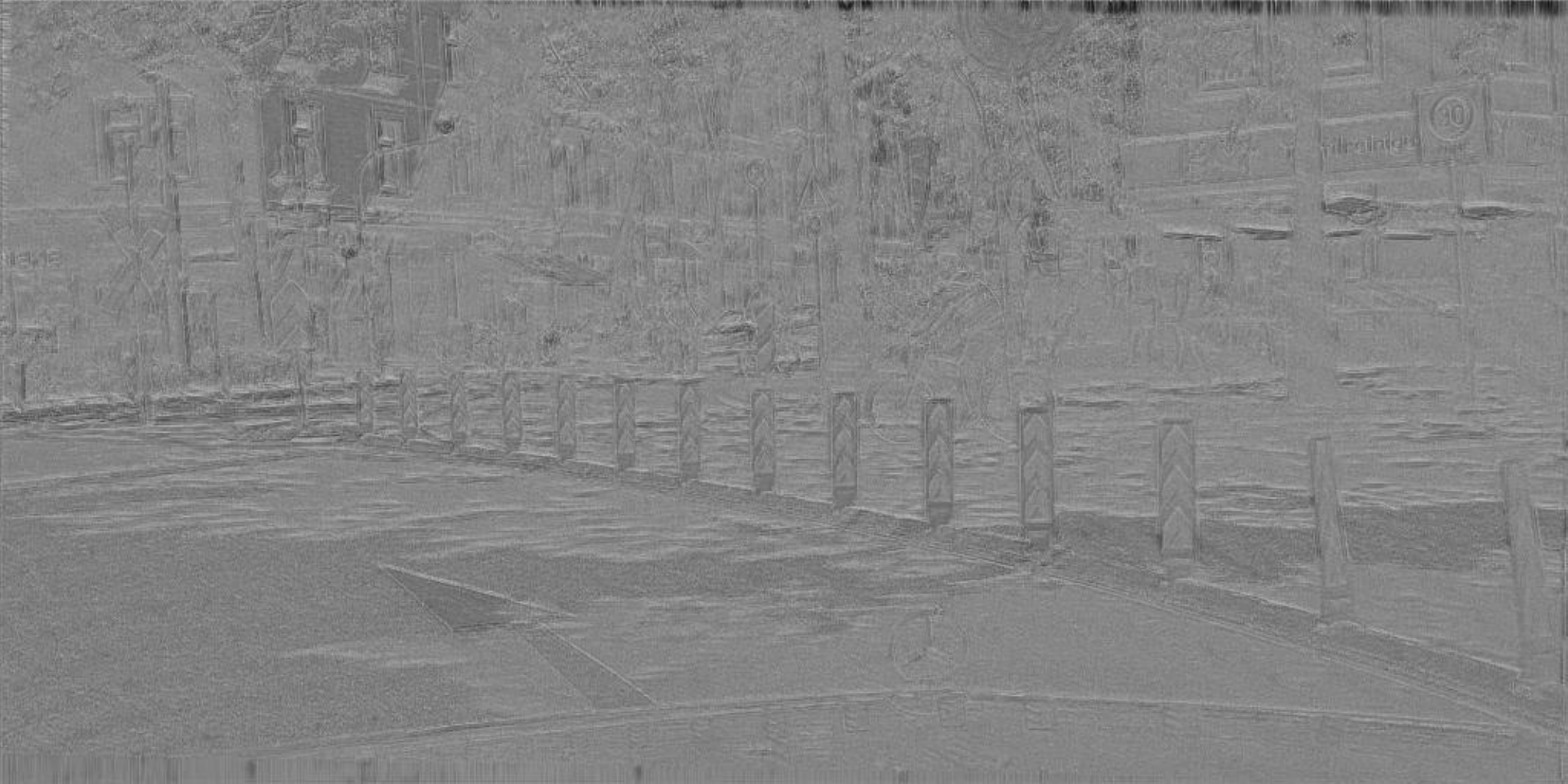}
        &
        \includegraphics[width=0.4\linewidth]{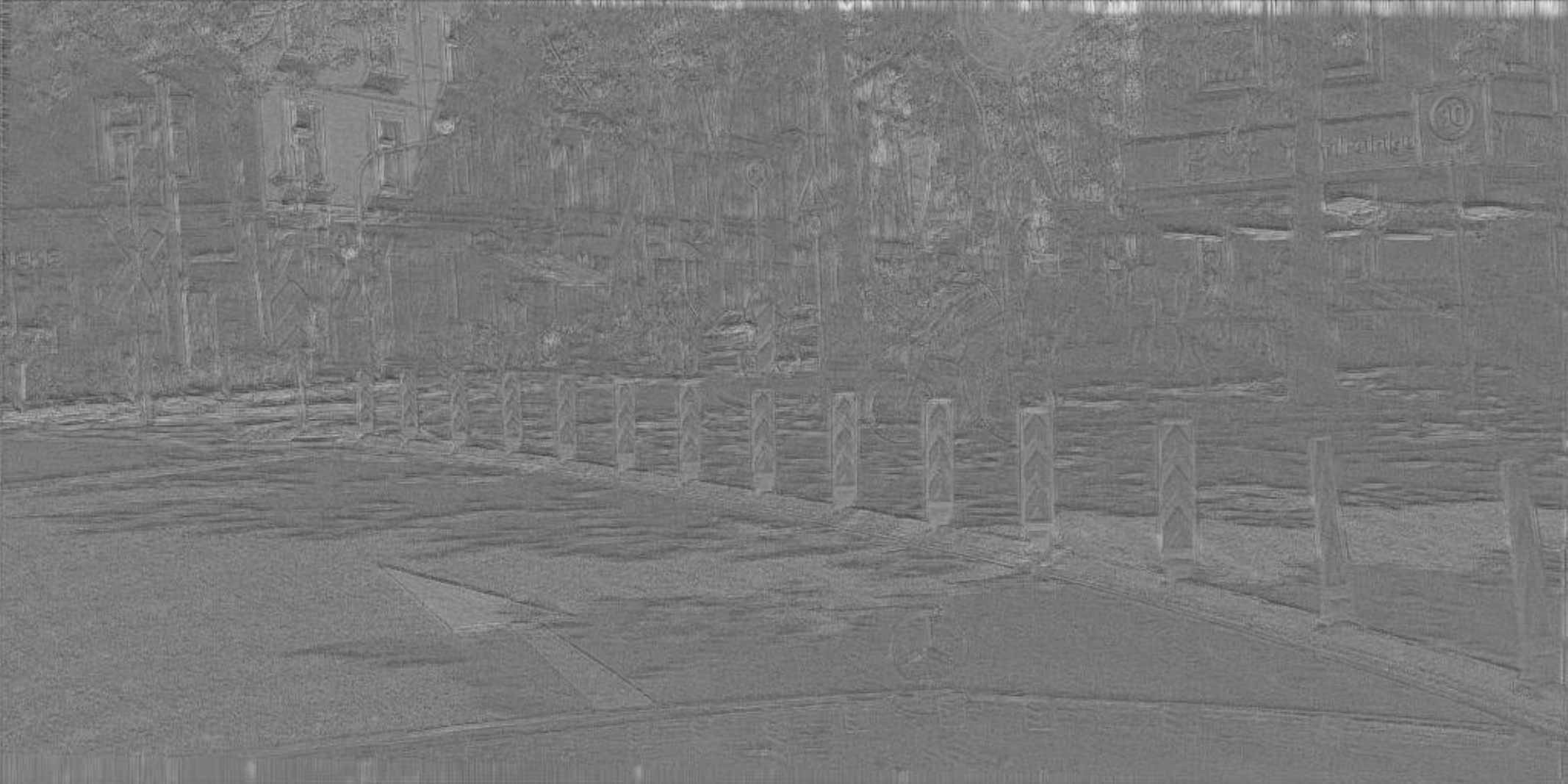} \\
        spatial attention for $L_1$
        &
        spatial attention for $L_2$ \\
        \includegraphics[width=0.4\linewidth]{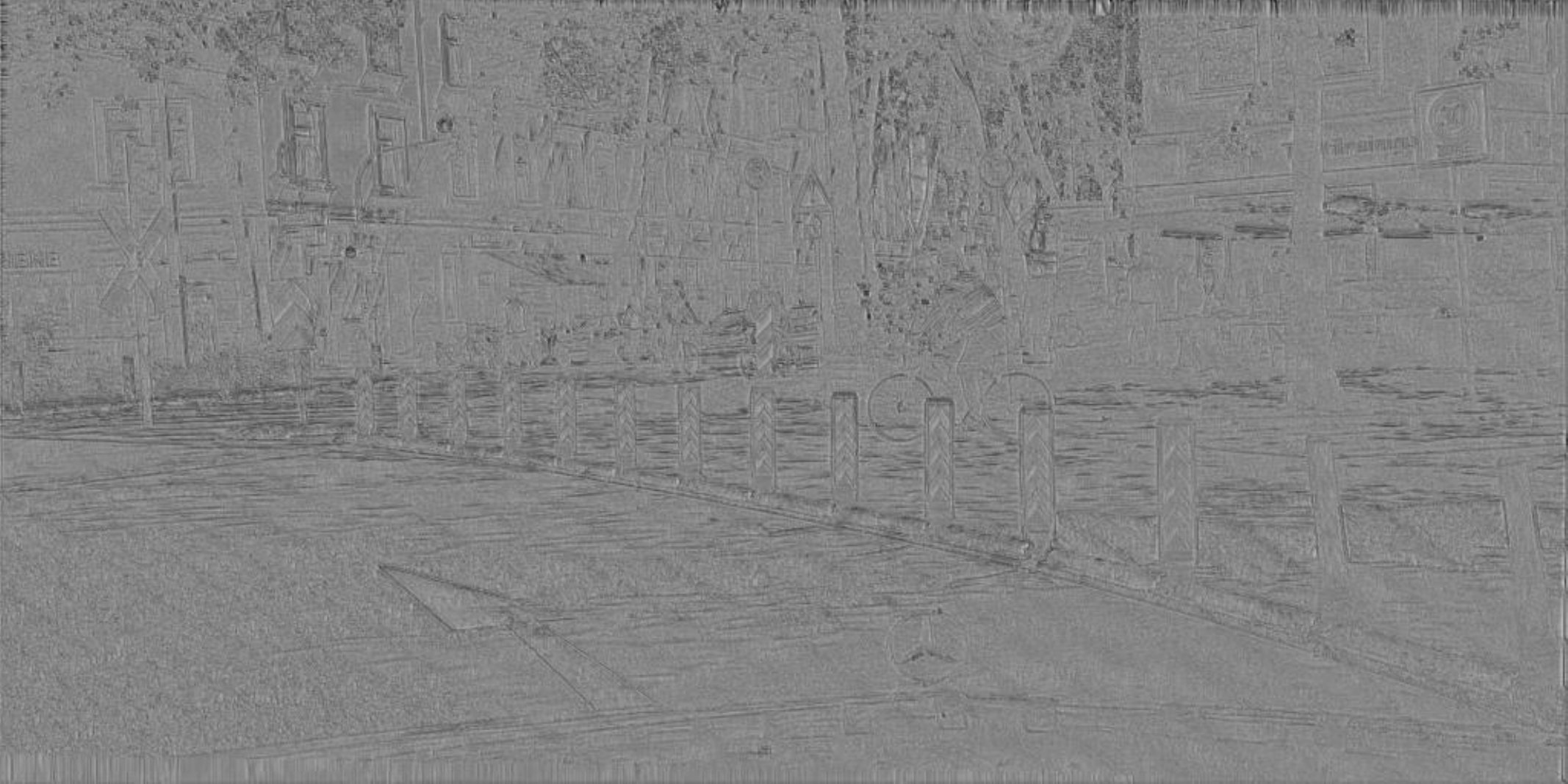}
        &
        \includegraphics[width=0.4\linewidth]{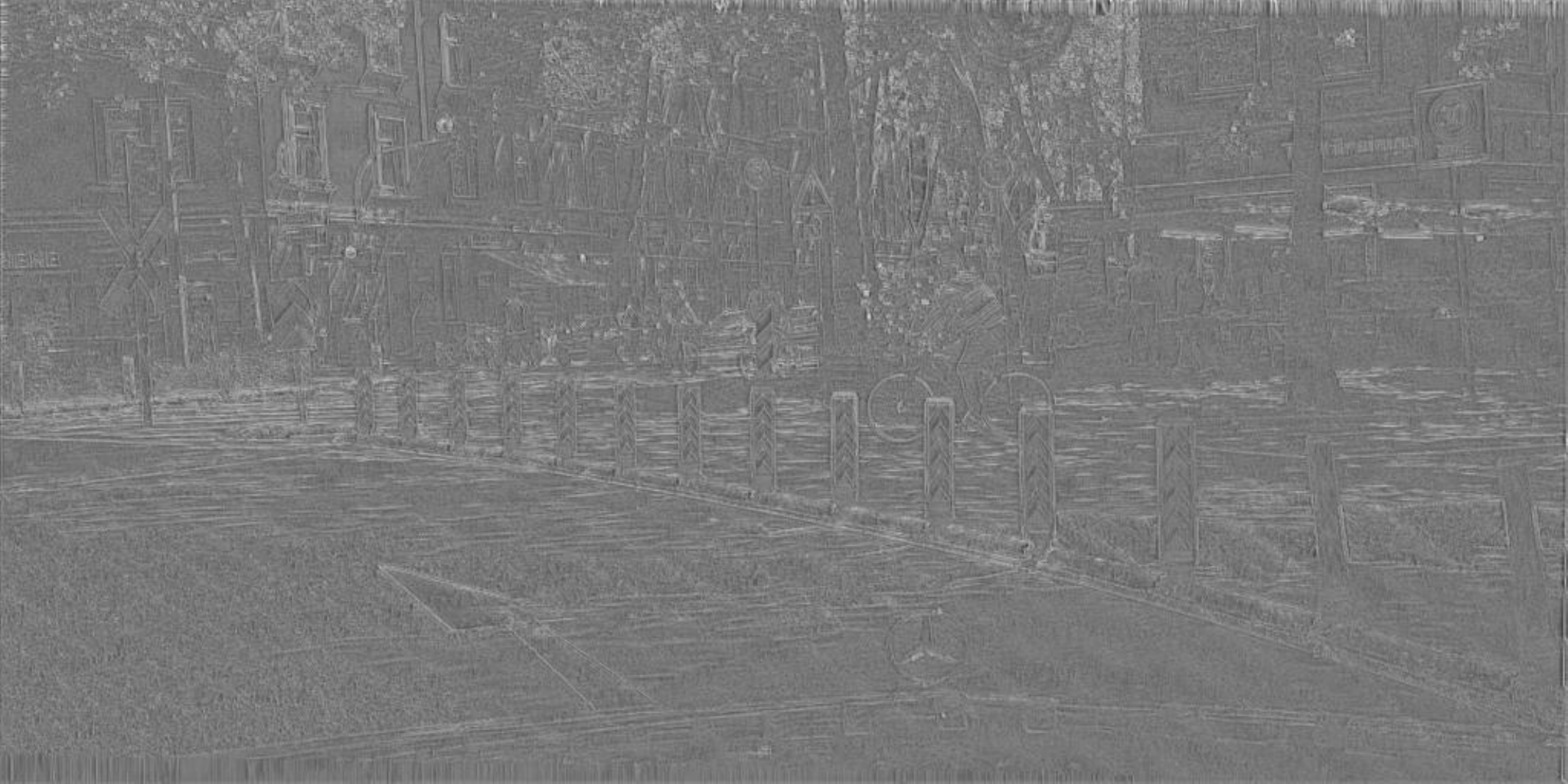} \\
        spatial attention for $L^1_2$ 
        &
        spatial attention for $L^1_3$ \\
        \includegraphics[width=0.4\linewidth]{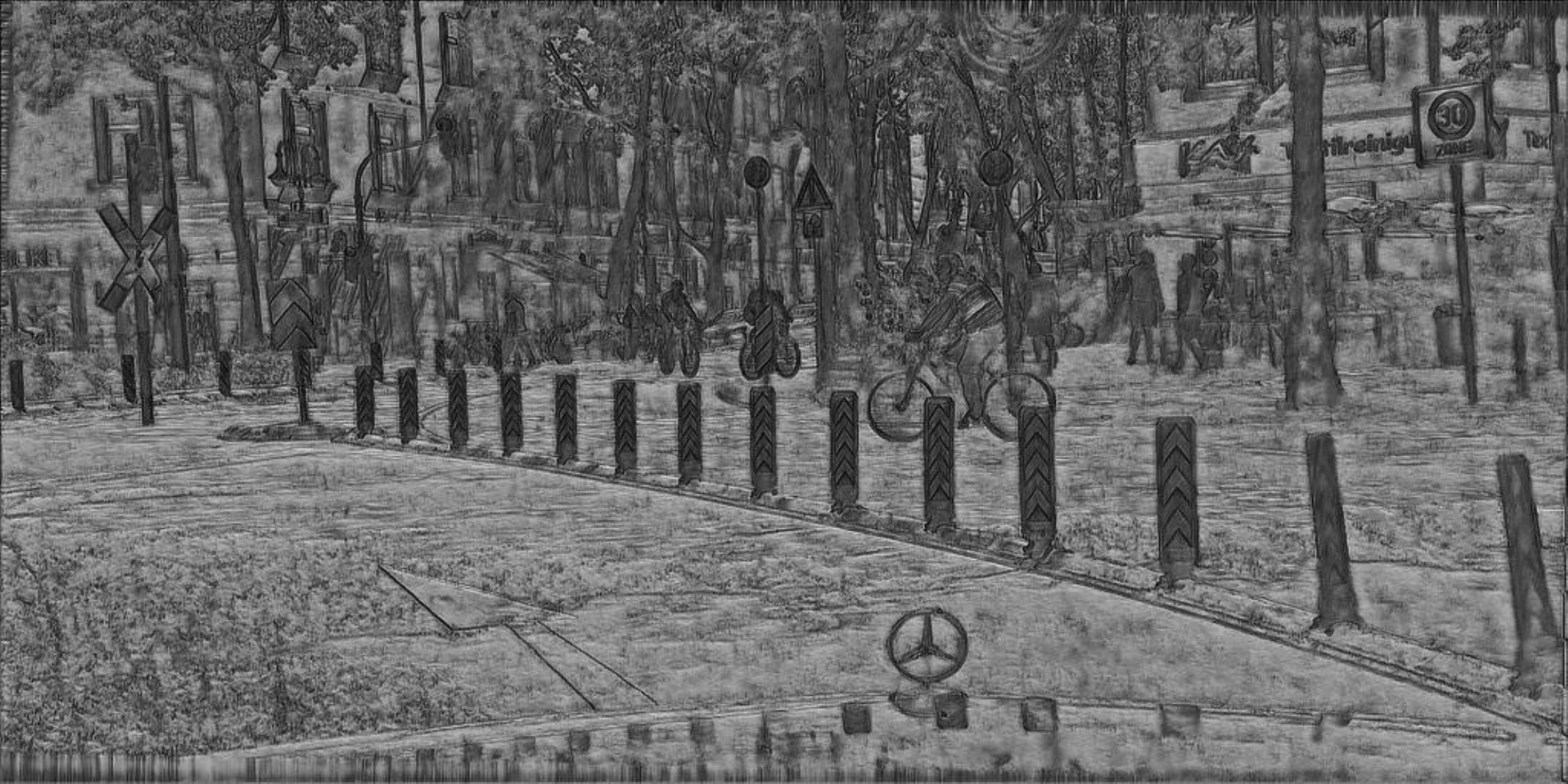}
        &
        \includegraphics[width=0.4\linewidth]{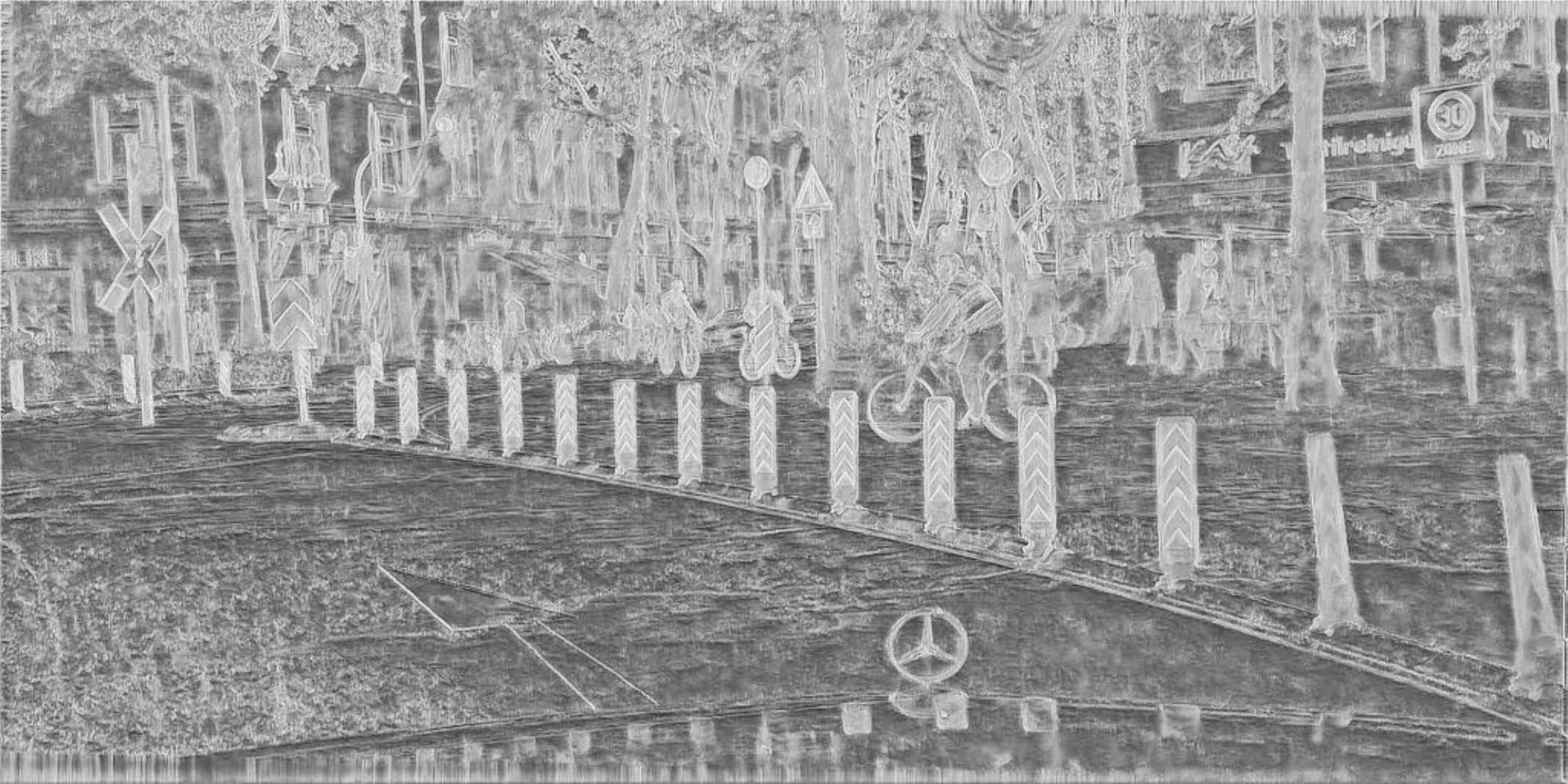} \\
        spatial attention for $L^2_3$ 
        &
        spatial attention for $L^2_4$ \\
        \includegraphics[width=0.4\linewidth]{images/AFA/L4_L5_4.pdf}
        &
        \includegraphics[width=0.4\linewidth]{images/AFA/L5_L4_4.pdf} \\
        spatial attention for $L^3_4$ 
        &
        spatial attention for $L^3_5$ \\
        \bottomrule
    \end{tabular}
    \vspace{0.05in}
    \caption{Spatial attention maps at four different levels generated by our binary fusion module which aggregates two features. Whiter regions denote higher attention. Compared to linear fusion operations, our AFA module provides a more expressive way of combining features.}
    \label{fig:AFA_GATE_1}
\end{figure*}

\begin{figure*}[t]
    \centering
    \begin{tabular}{cc}
        \toprule
        \includegraphics[width=0.4\linewidth]{images/AFA/input_2.pdf}
        &
        \includegraphics[width=0.4\linewidth]{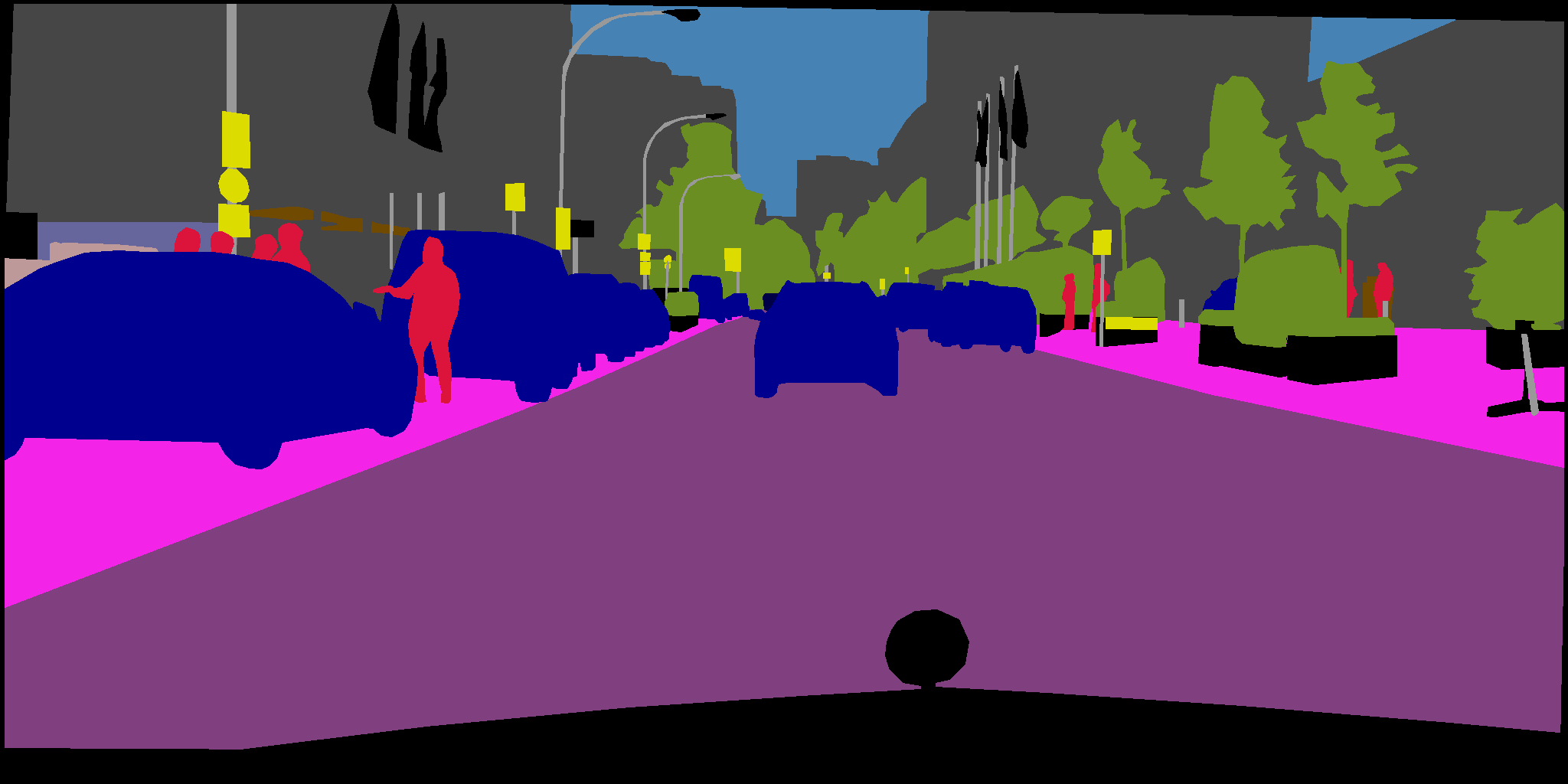} \\
        Input
        &
        Ground Truth \\
        \includegraphics[width=0.4\linewidth]{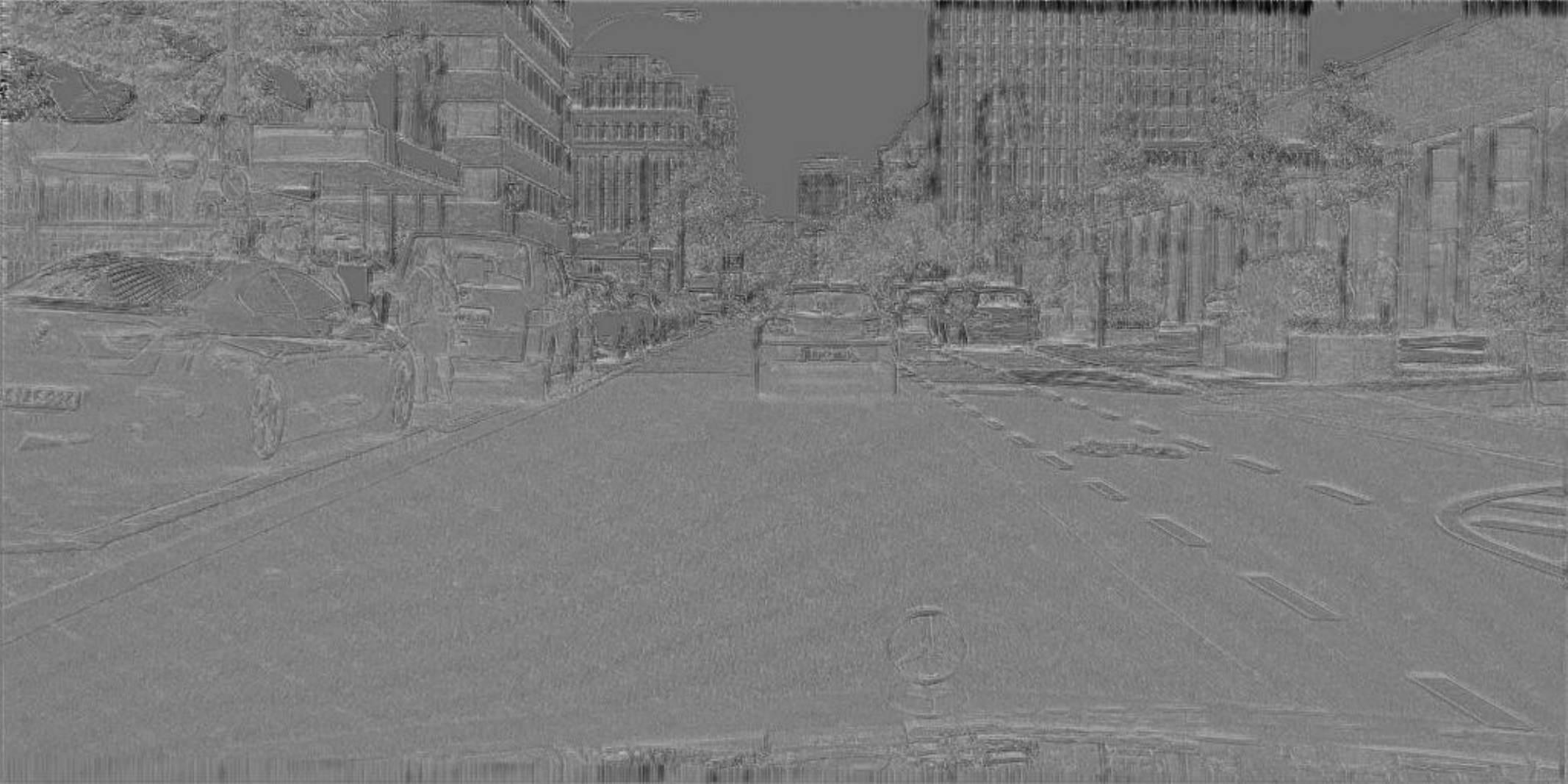}
        &
        \includegraphics[width=0.4\linewidth]{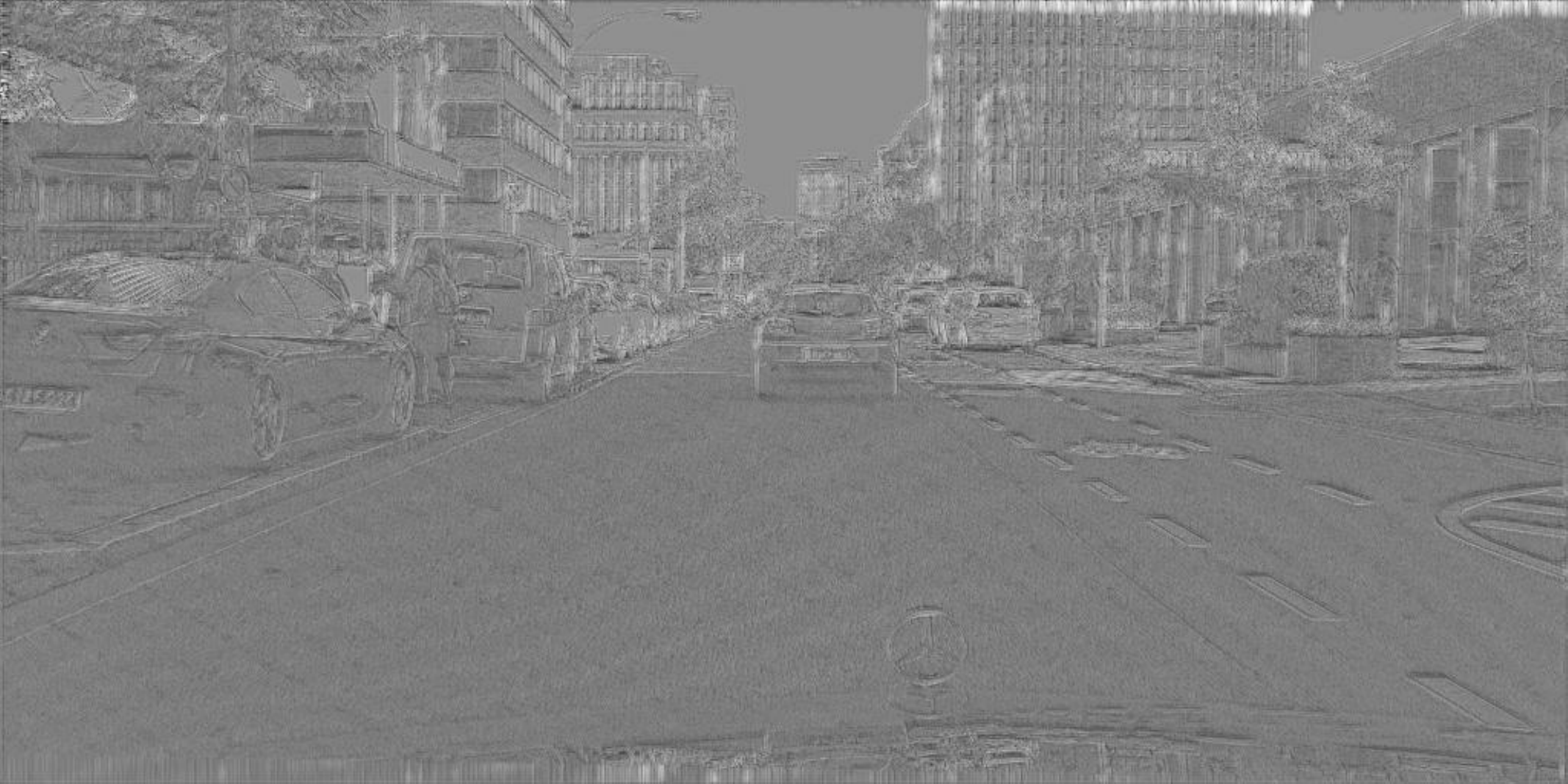} \\
        spatial attention for $L_1$
        &
        spatial attention for $L_2$ \\
        \includegraphics[width=0.4\linewidth]{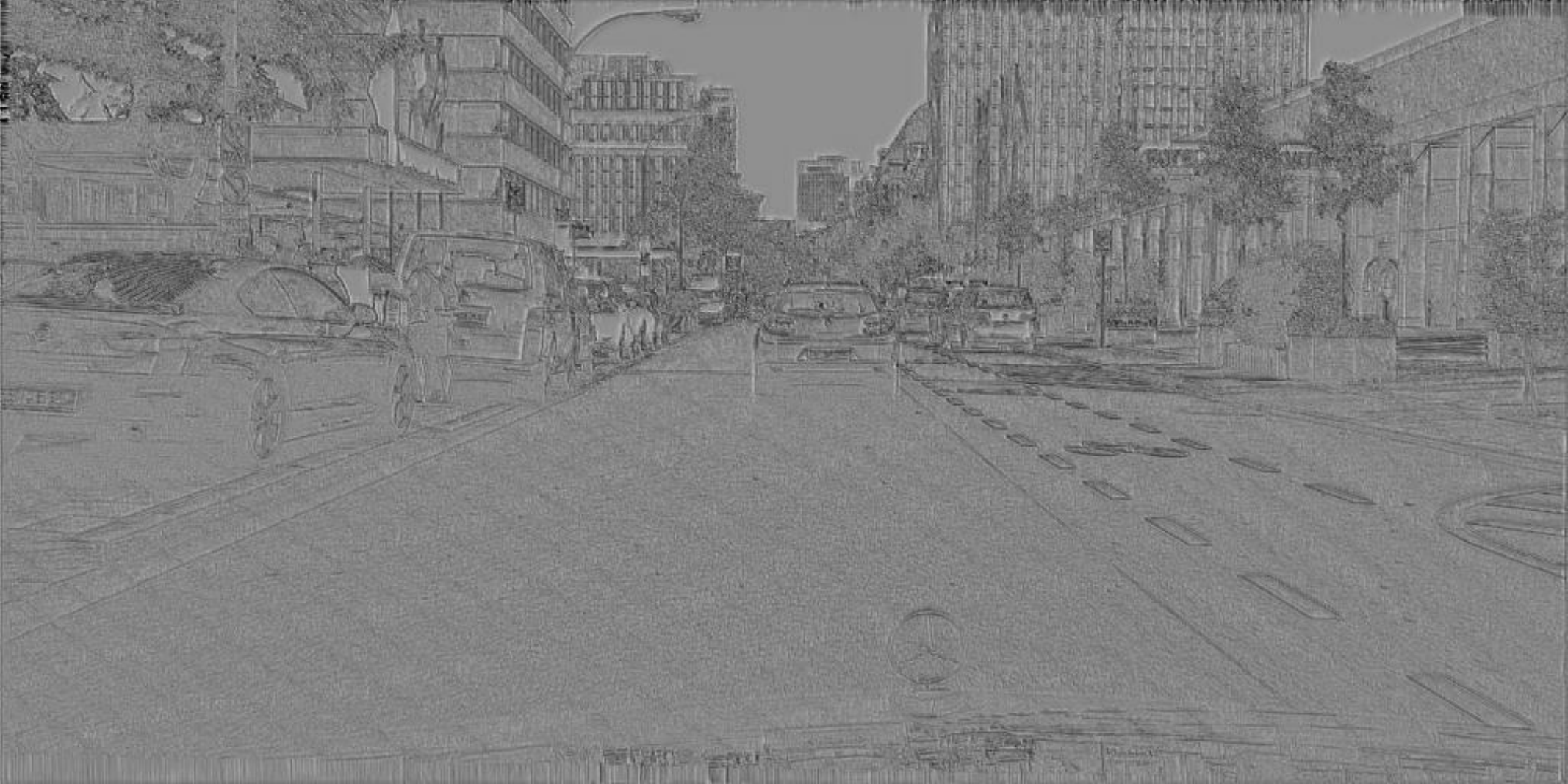}
        &
        \includegraphics[width=0.4\linewidth]{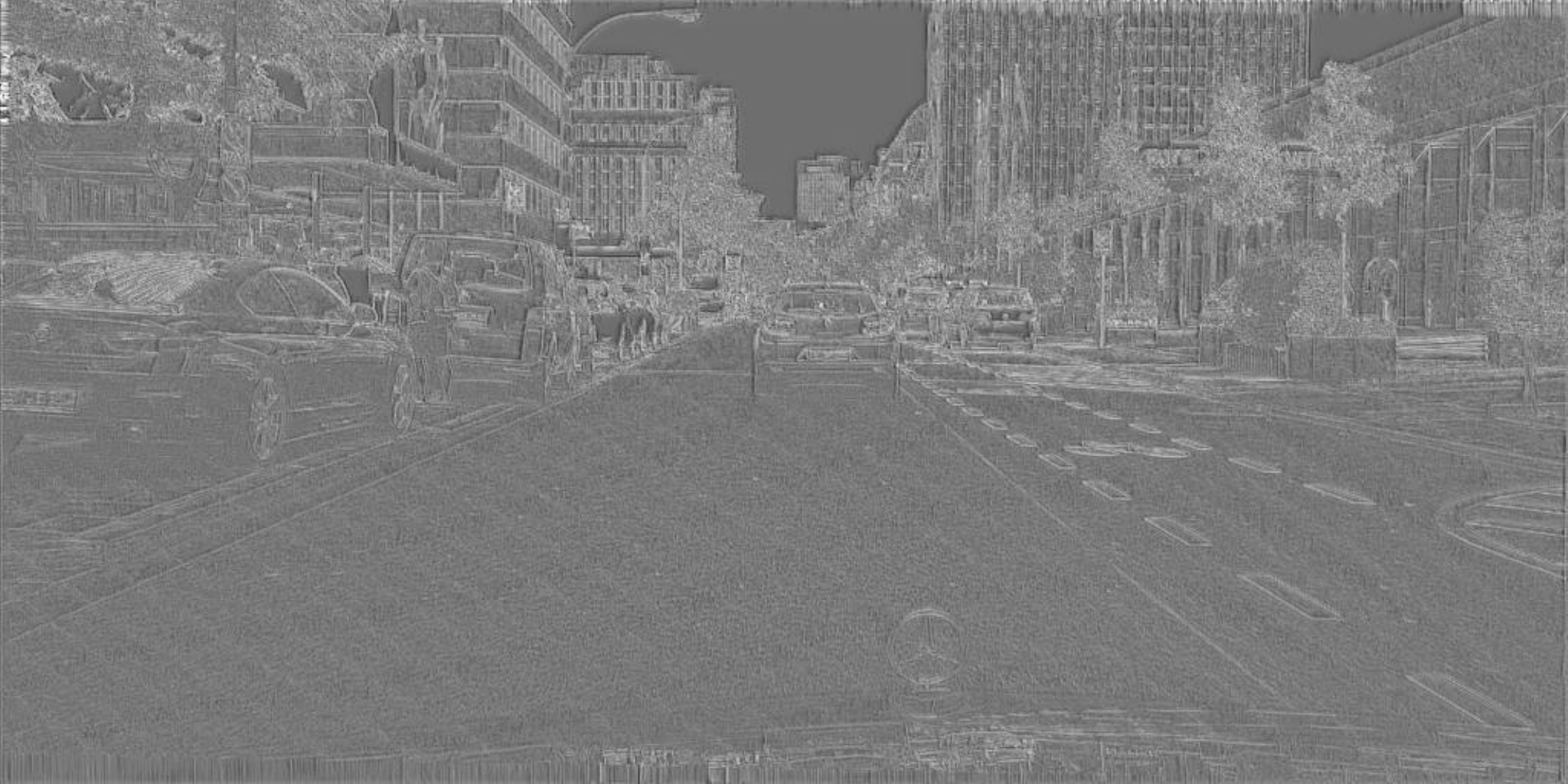} \\
        spatial attention for $L^1_2$ 
        &
        spatial attention for $L^1_3$ \\
        \includegraphics[width=0.4\linewidth]{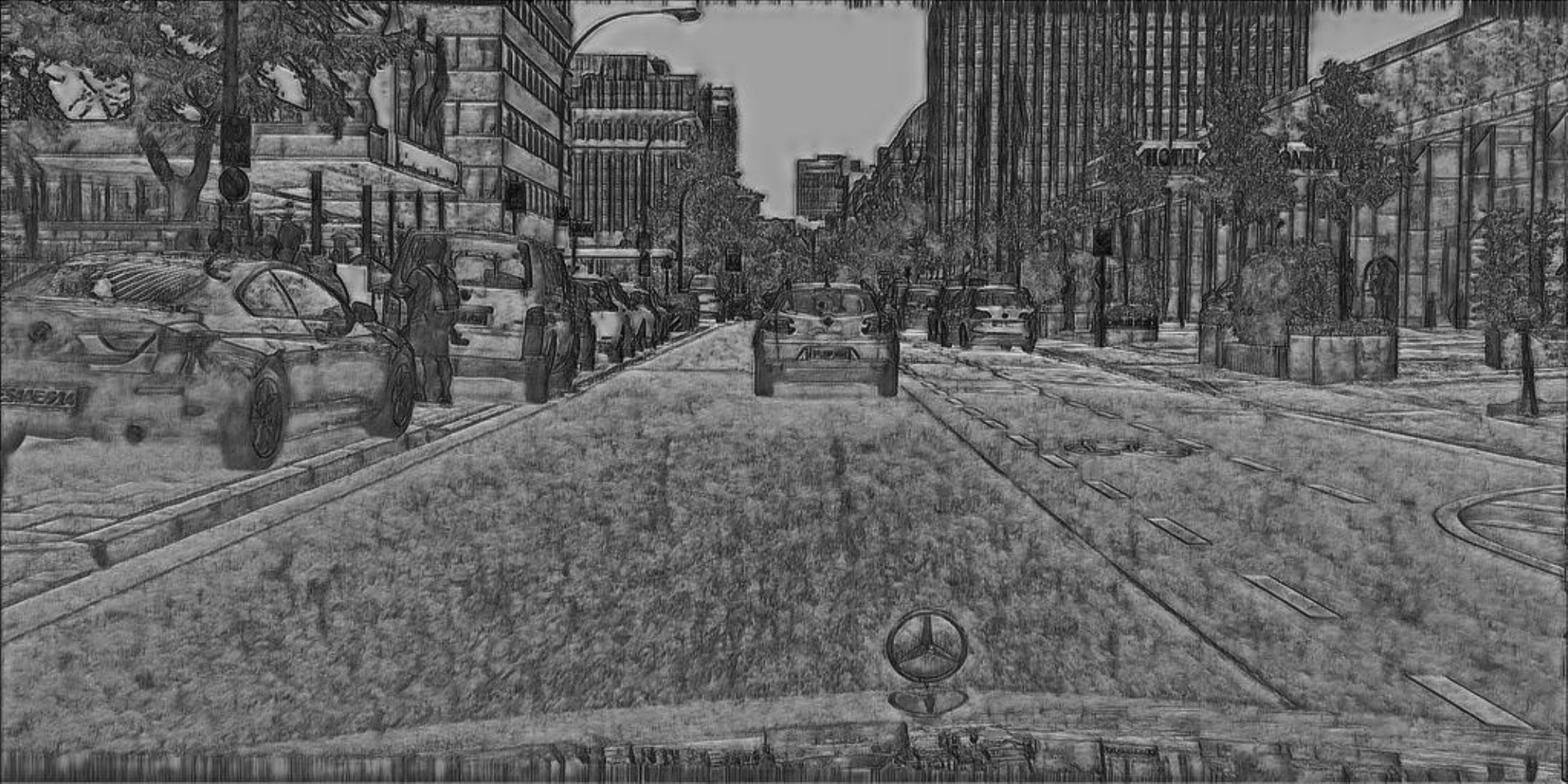}
        &
        \includegraphics[width=0.4\linewidth]{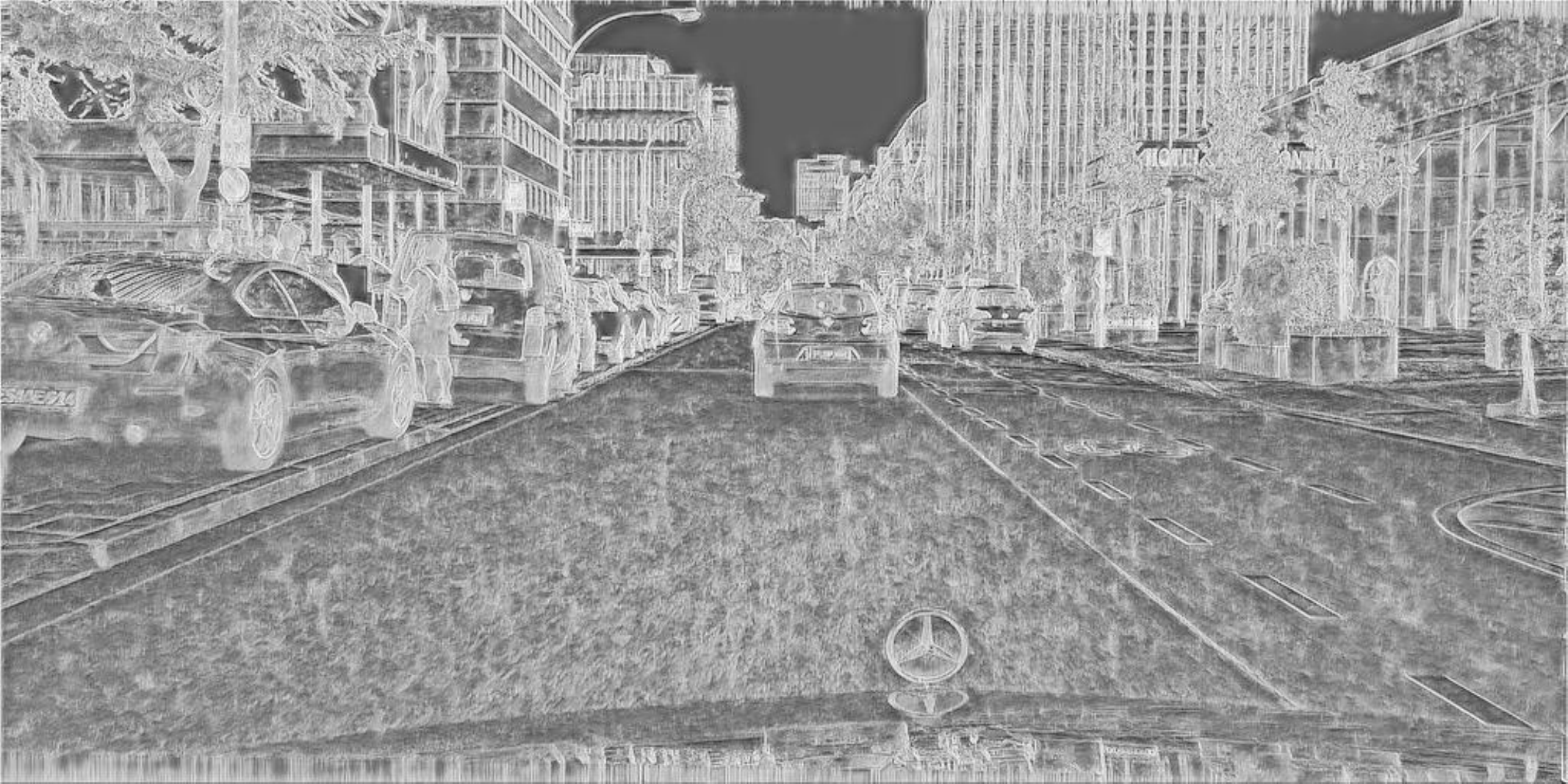} \\
        spatial attention for $L^2_3$ 
        &
        spatial attention for $L^2_4$ \\
        \includegraphics[width=0.4\linewidth]{images/AFA/L4_L5.pdf}
        &
        \includegraphics[width=0.4\linewidth]{images/AFA/L5_L4.pdf} \\
        spatial attention for $L^3_4$ 
        &
        spatial attention for $L^3_5$ \\
        \bottomrule
    \end{tabular}
    \vspace{0.05in}
    \caption{Spatial attention maps at four different levels generated by our binary fusion module which aggregates two features. Whiter regions denote higher attention. Compared to linear fusion operations, our AFA module provides a more expressive way of combining features.}
    \label{fig:AFA_GATE_2}
\end{figure*}

We first look at the attention maps generated by our binary fusion module which aggregates two features in Figure~\ref{fig:AFA_GATE_1} and Figure~\ref{fig:AFA_GATE_2}.
We provide the spatial attention maps for binary fusion at four different levels.
When the difference of the level information between two input features is larger (e.g., $L^3_4$ and $L^3_5$), our attention mask will become more specific and be able to focus on the right place to be fused.
Take the fusion of $L^3_4$ and $L^3_5$ as example.
Since $L^3_4$ contains the information of the $L_1$ feature, our attention focuses on object boundaries on it and attend to the rest on $L^3_5$, which has richer semantic information.
Compared to linear fusion operations, our AFA module provides a more expressive way of combining features.
% Since $L^3_4$ contains the information of $L_1$ feature, so our attention focus on the object border on it and attend the rest of part on $L^3_5$, which has richer semantic information.

\begin{figure*}[t]
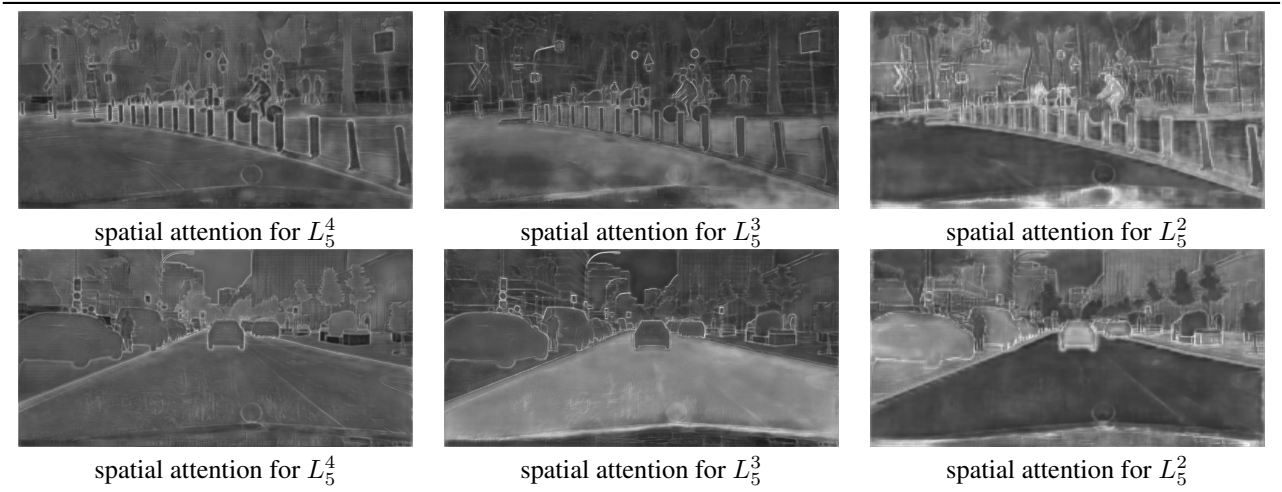

    \centering
    \begin{tabular}{ccc}
        \toprule
        \includegraphics[width=0.3\linewidth]{images/AFA/L5_4.pdf}
        &
        \includegraphics[width=0.3\linewidth]{images/AFA/L5_3.pdf}
        &
        \includegraphics[width=0.3\linewidth]{images/AFA/L5_2.pdf} \\
        spatial attention for $L^4_5$
        &
        spatial attention for $L^3_5$
        & 
        spatial attention for $L^2_5$ \\
        \includegraphics[width=0.3\linewidth]{images/AFA/L5_4_2.pdf}
        &
        \includegraphics[width=0.3\linewidth]{images/AFA/L5_3_2.pdf}
        &
        \includegraphics[width=0.3\linewidth]{images/AFA/L5_2_2.pdf} \\
        spatial attention for $L^4_5$
        &
        spatial attention for $L^3_5$
        & 
        spatial attention for $L^2_5$ \\
        \bottomrule
    \end{tabular}
    \vspace{0.05in}
    \caption{
        Spatial attention maps generated by our multiple feature fusion module which aggregates multiple features. Whiter regions denote higher attention. With our multiple feature fusion module, our model can strike a balance between the low-level and the high-level information and perform fusion accordingly.
    }
	\label{fig:AFA_Final}
\end{figure*}

We additionally look at the spatial attention maps generated by our multiple feature fusion module in Figure~\ref{fig:AFA_Final}.
% Also, for the multiple feature aggregation, we took a deep analysis on it without the effect of our scale-space attention involved.
% We plot the visualization result in Figure~\ref{fig:AFA_Final}.
Only using the final aggregated features for prediction may cause our model to overly focus on low level features.
Thus, our multiple feature fusion module provides the model with more flexibility to select between the features that contain different low level information.
For input features that contain $L_1$ information like $L^4_5$ and $L^3_5$, the attention focuses more on the object boundaries, similar to our binary fusion module.
For other input features like $L^2_5$, the attention can focus on objects or the background.
With our multiple feature fusion module, our model can strike a balance between the low-level and the high-level information and perform fusion accordingly.

\section{Qualitative Results}
\begin{figure*}[t]
    \centering
    \begin{tabular}{ccc}
        \toprule
        Input & Ground Truth & Prediction \\
        \midrule
        \includegraphics[width=0.3\linewidth]{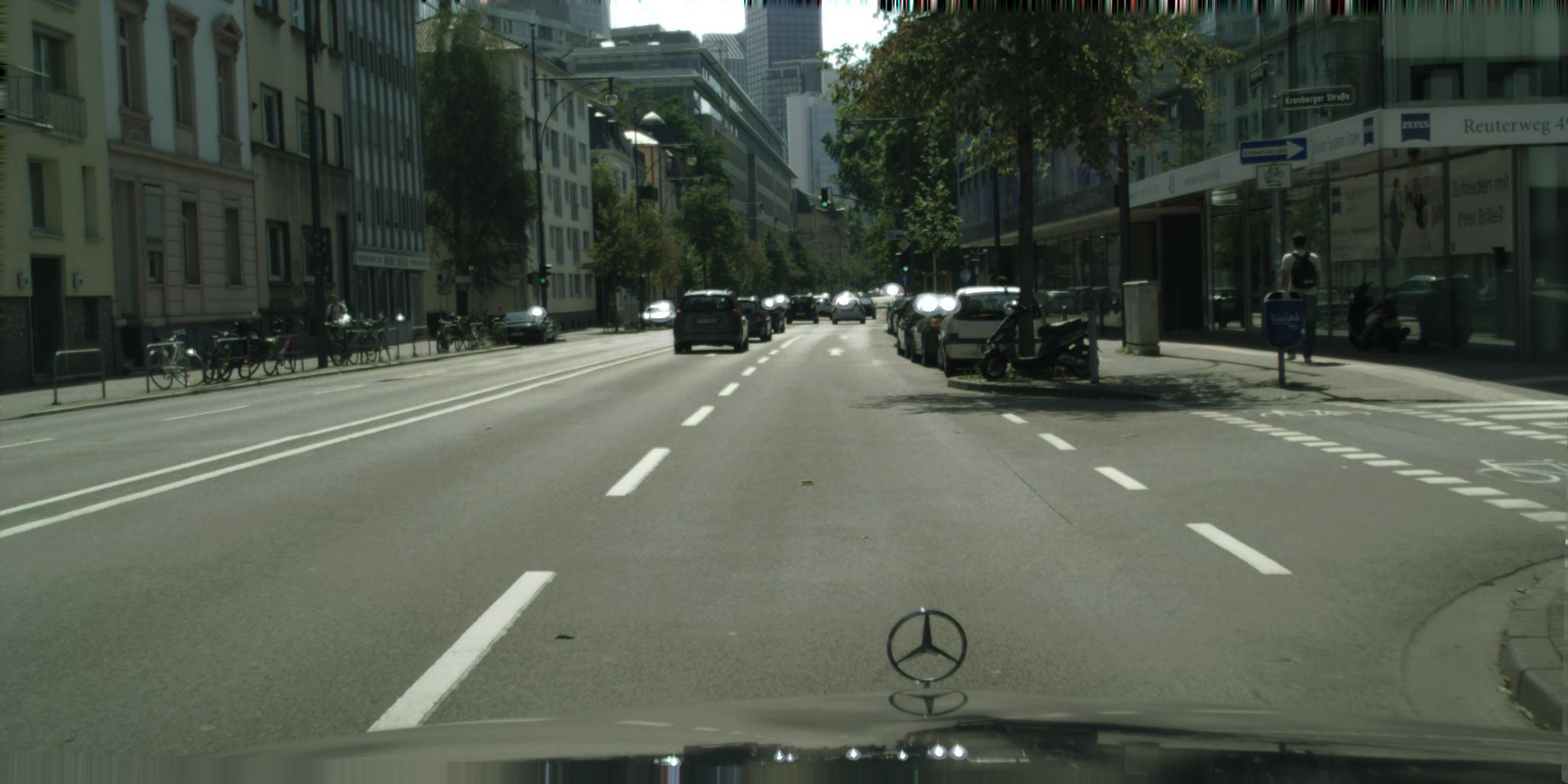}
        &
        \includegraphics[width=0.3\linewidth]{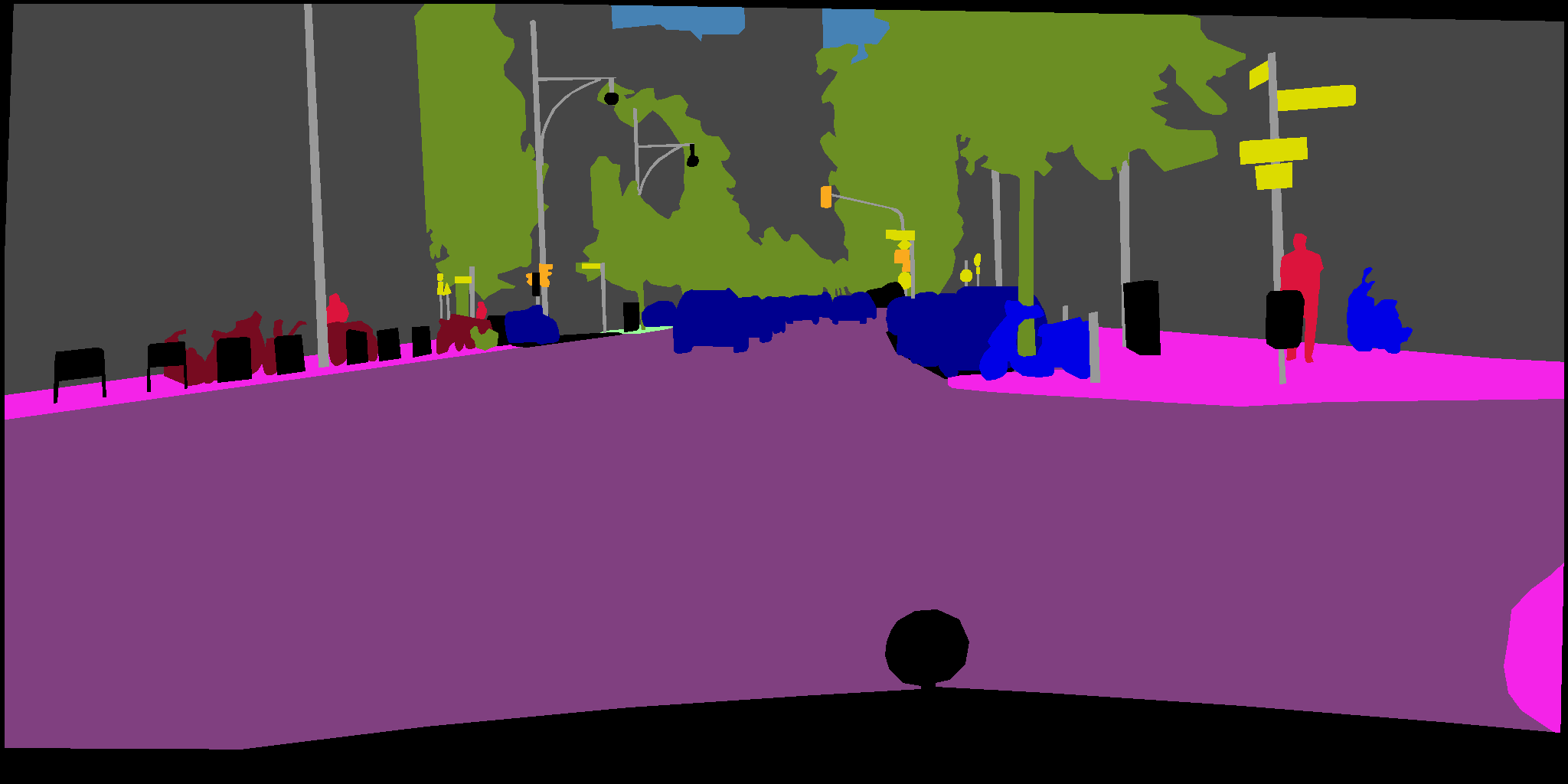}
        &
        \includegraphics[width=0.3\linewidth]{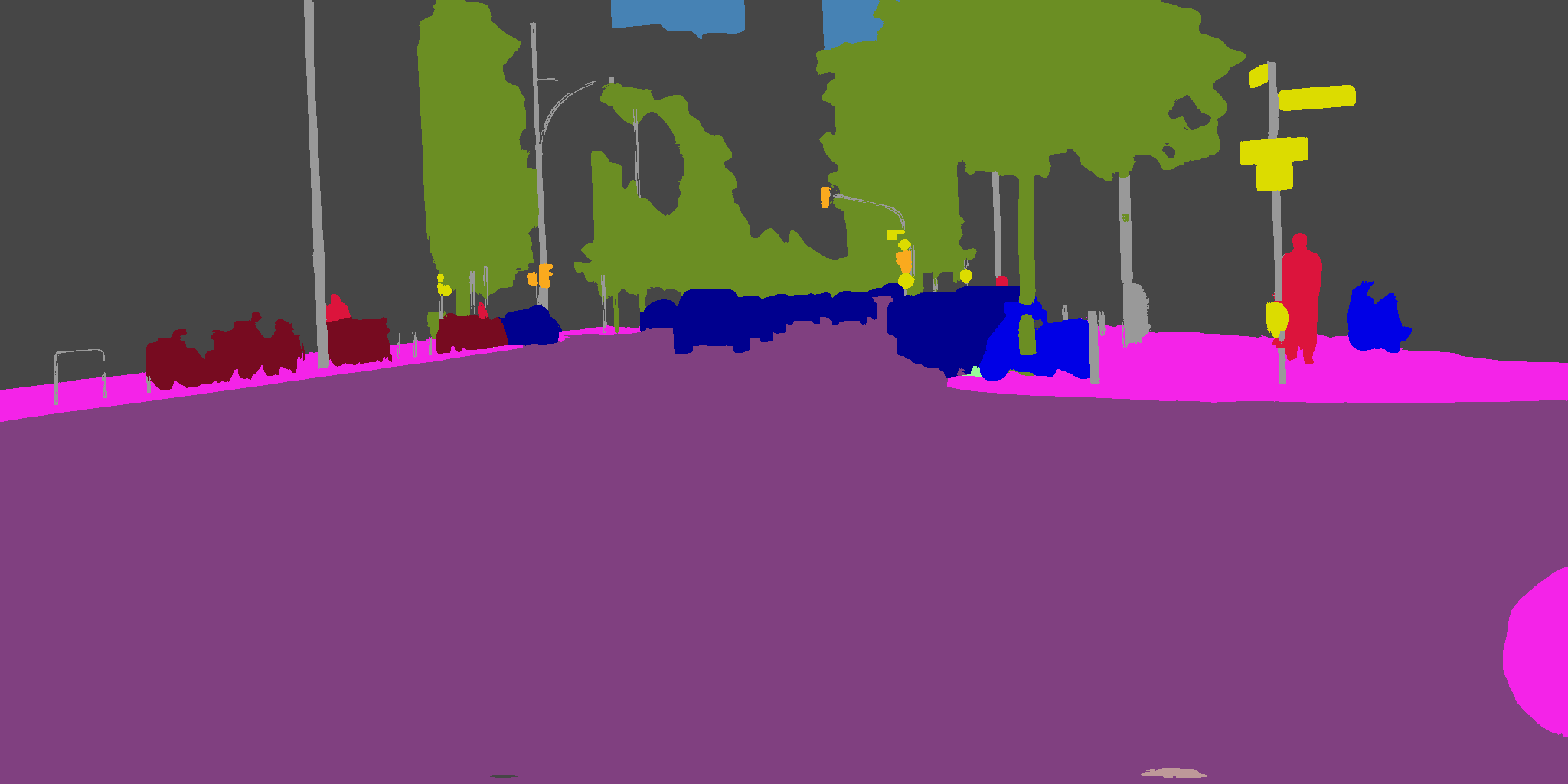} \\
        \includegraphics[width=0.3\linewidth]{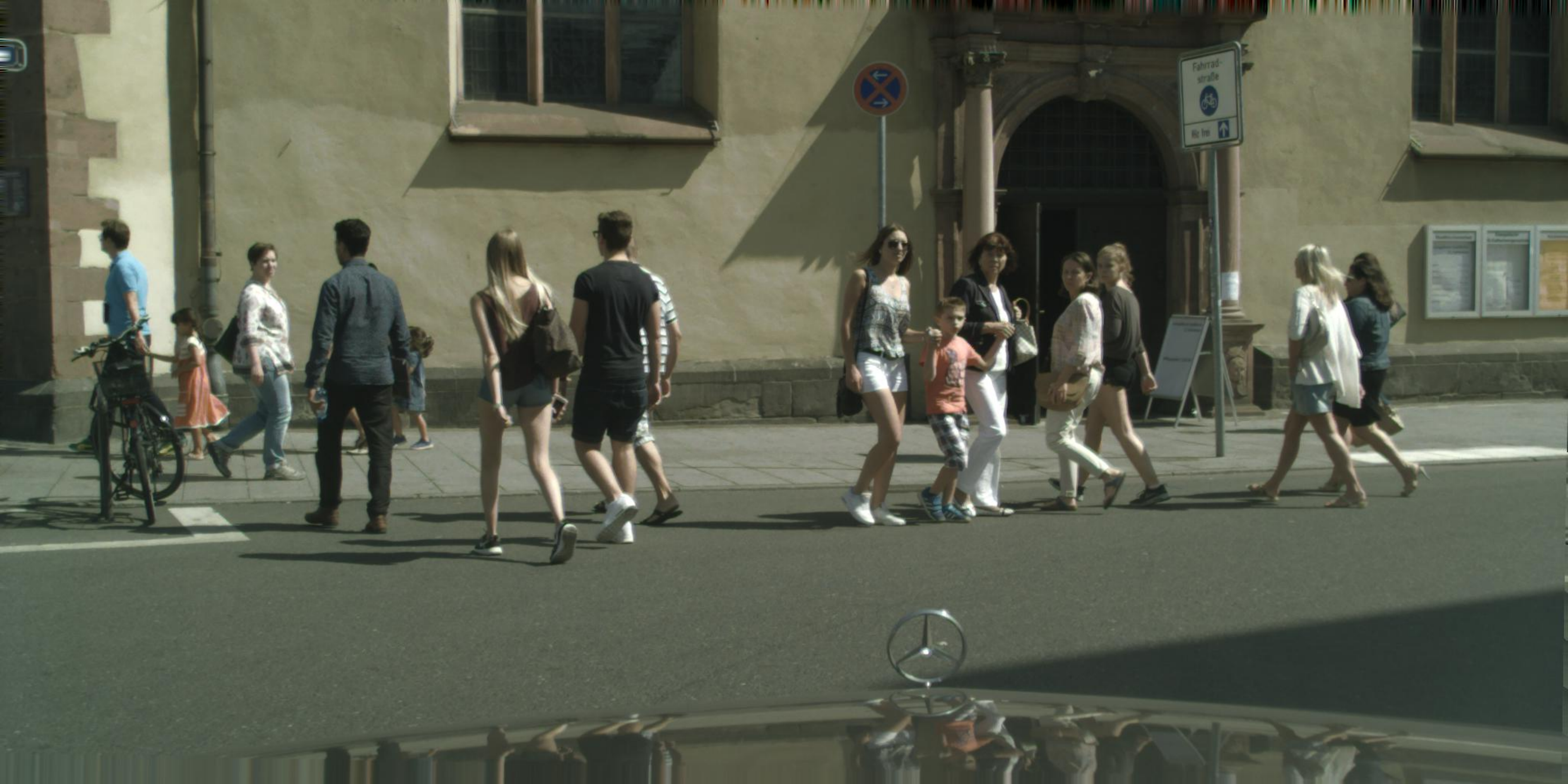}
        &
        \includegraphics[width=0.3\linewidth]{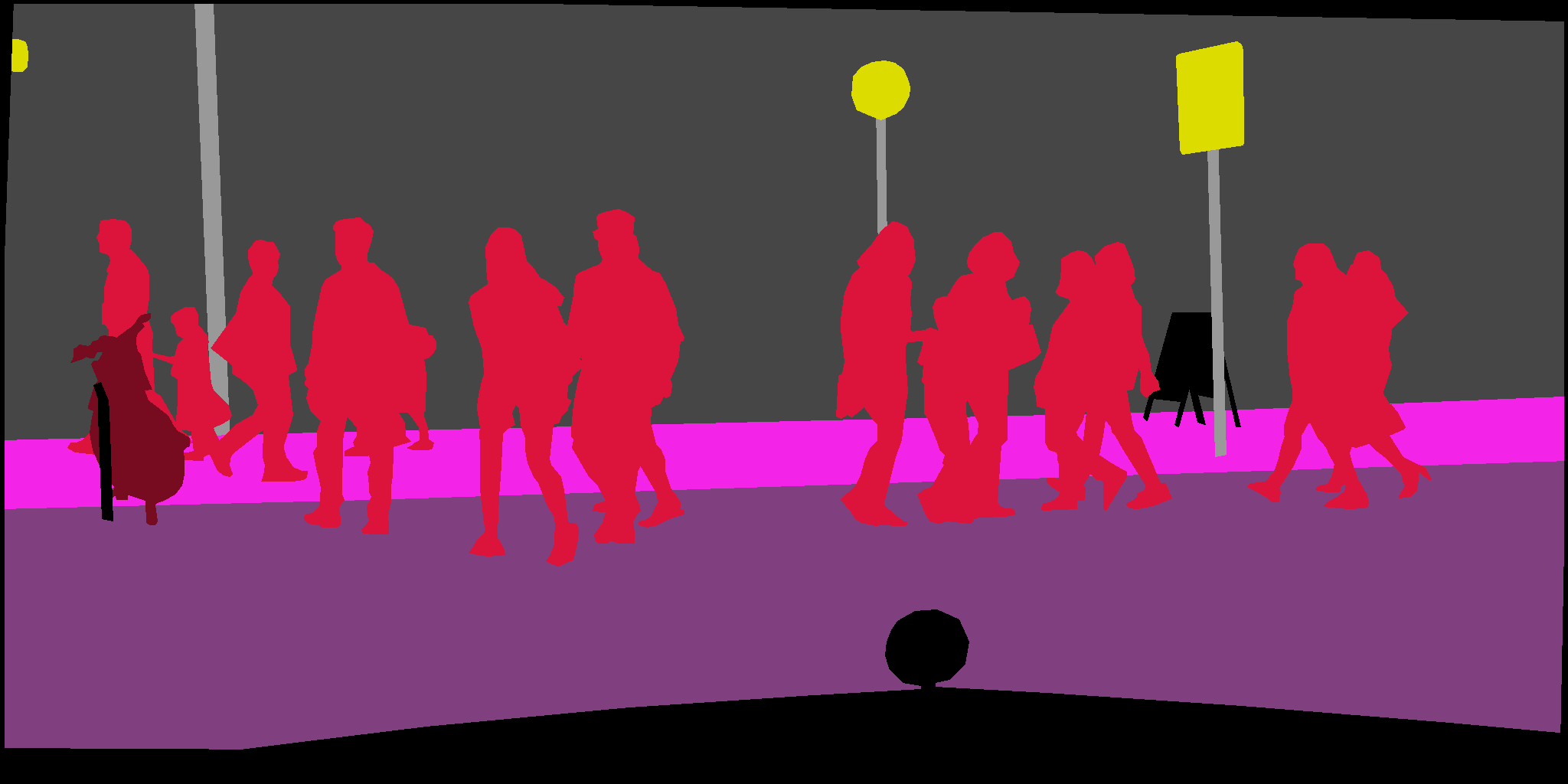}
        &
        \includegraphics[width=0.3\linewidth]{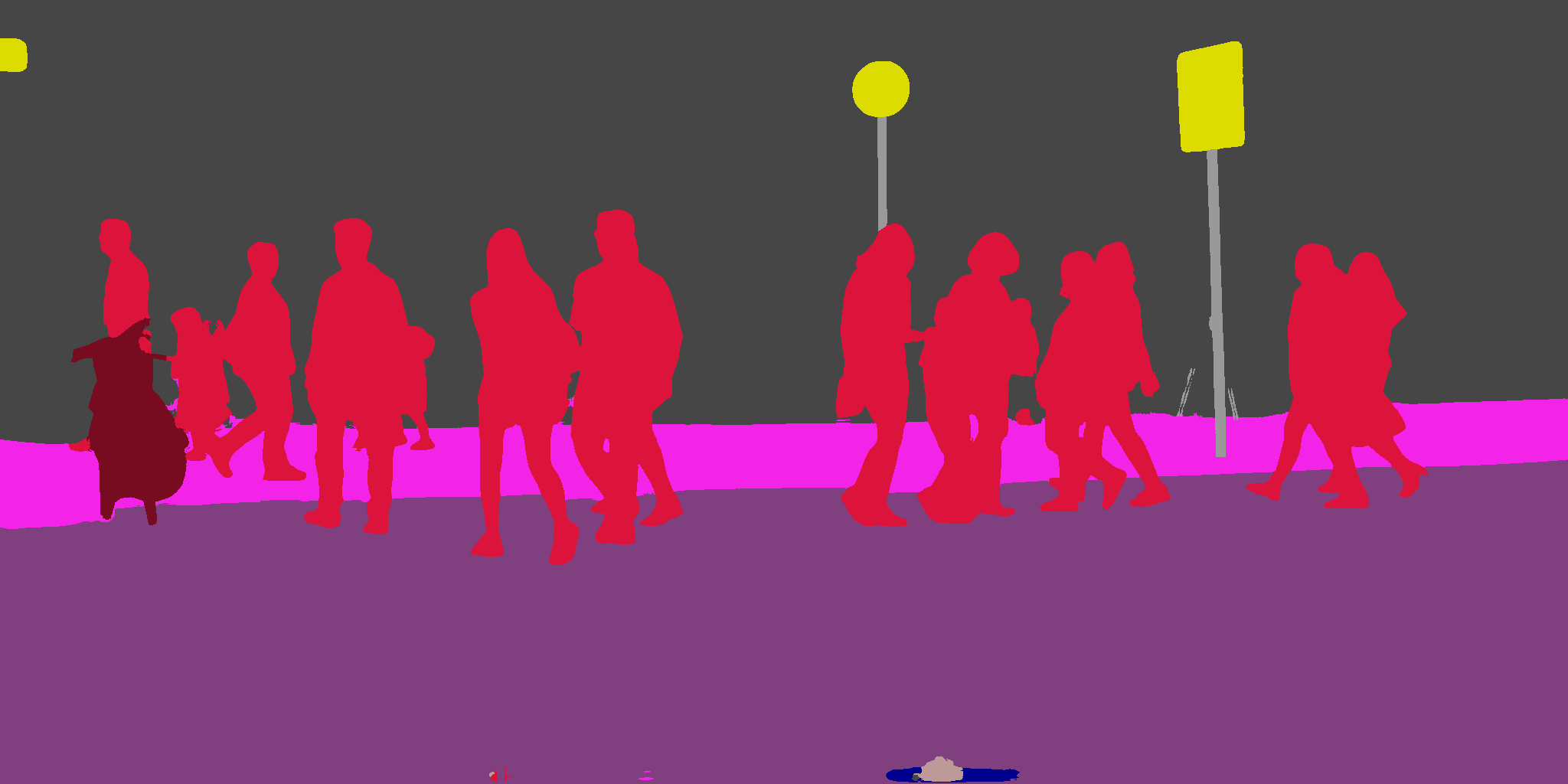} \\
        \includegraphics[width=0.3\linewidth]{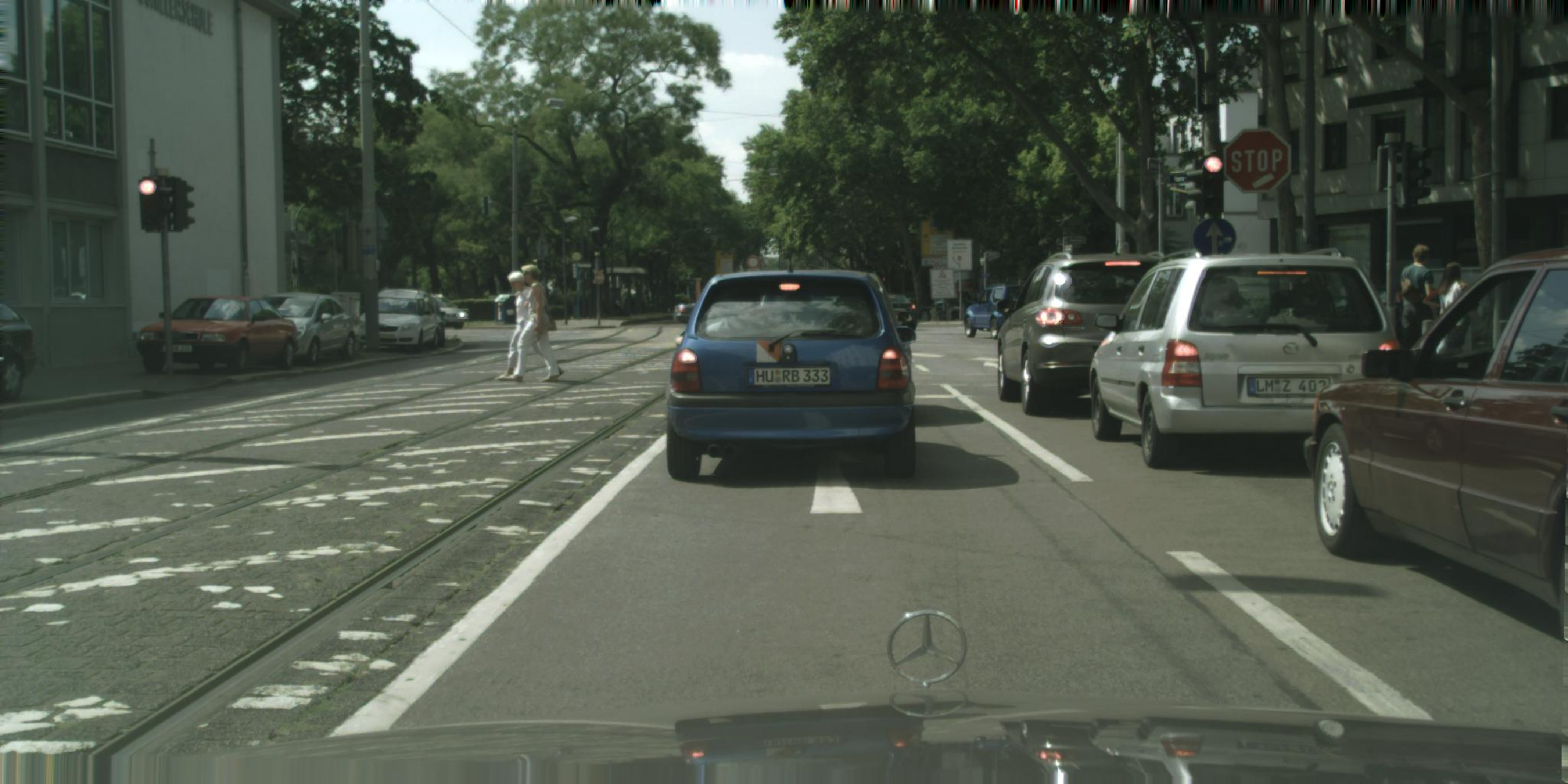}
        &
        \includegraphics[width=0.3\linewidth]{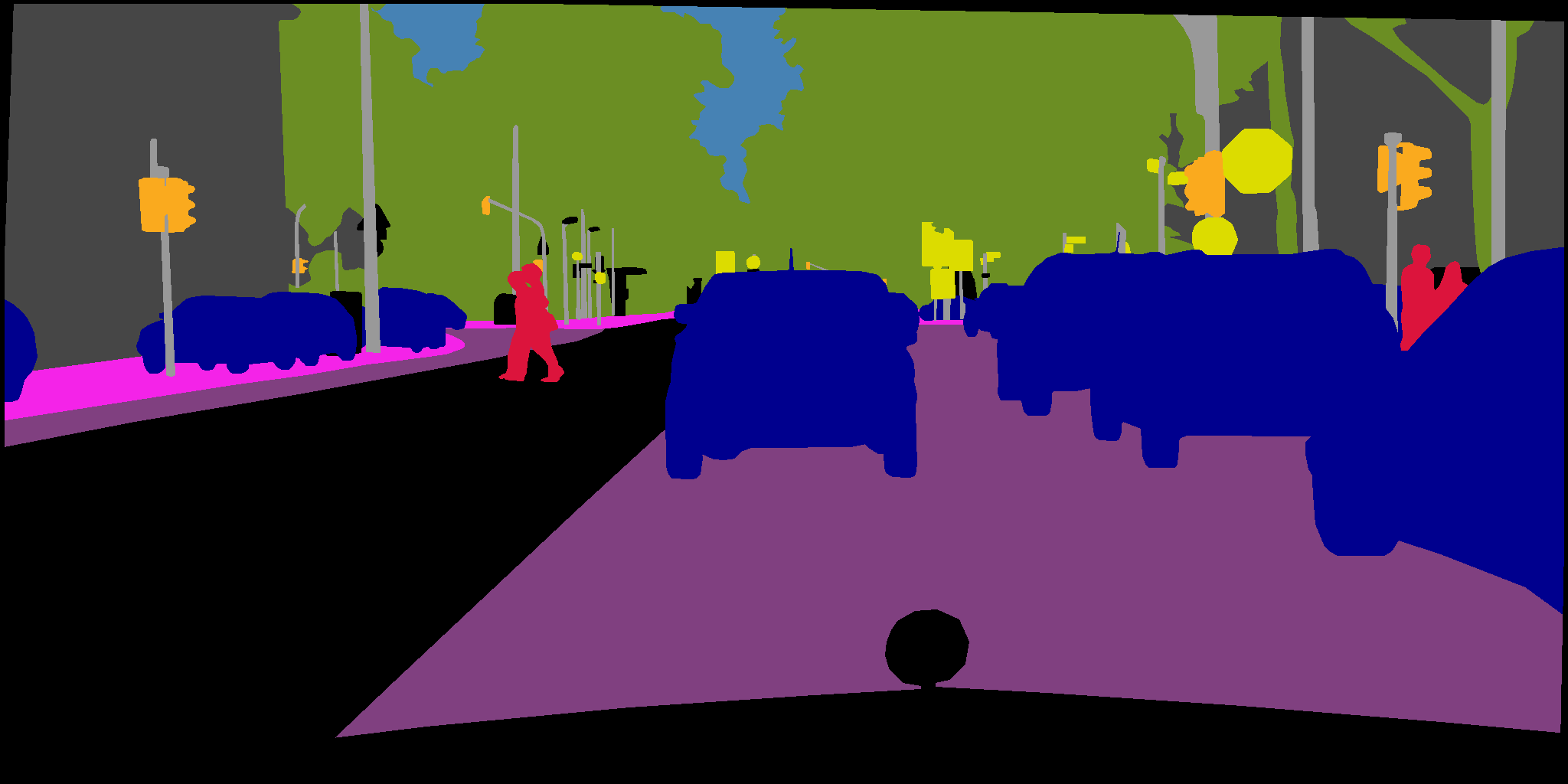}
        &
        \includegraphics[width=0.3\linewidth]{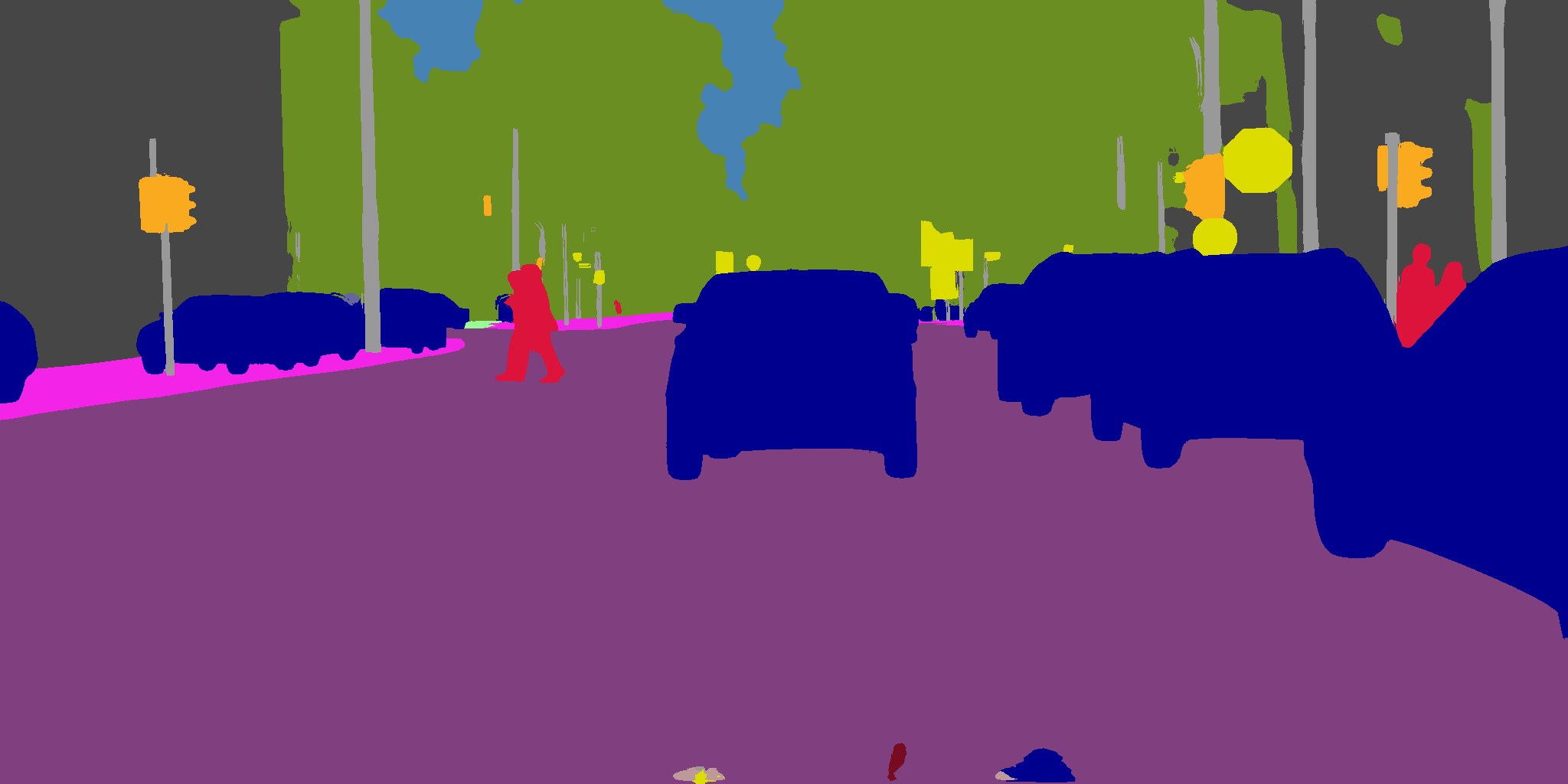} \\
        \includegraphics[width=0.3\linewidth]{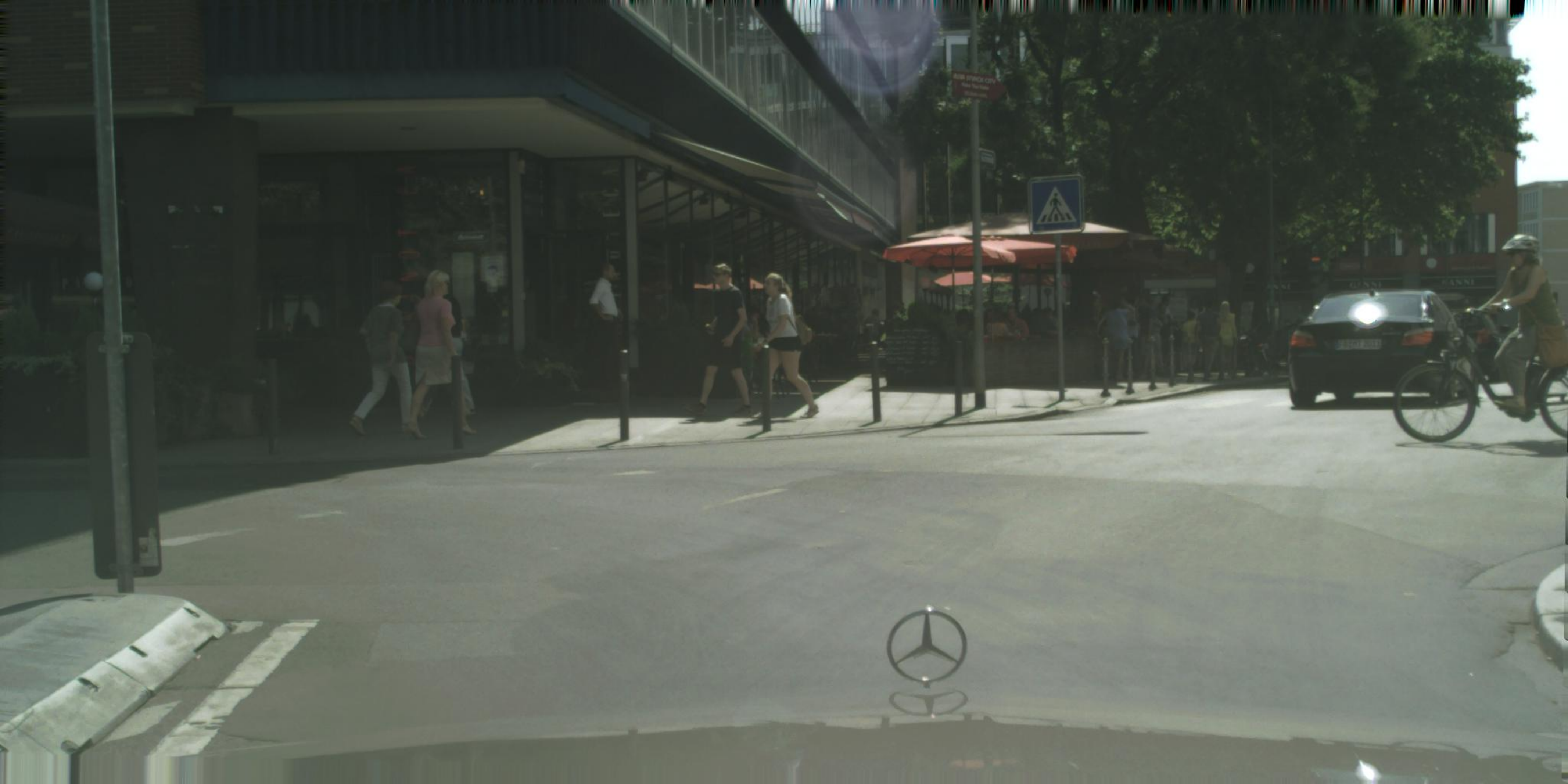}
        &
        \includegraphics[width=0.3\linewidth]{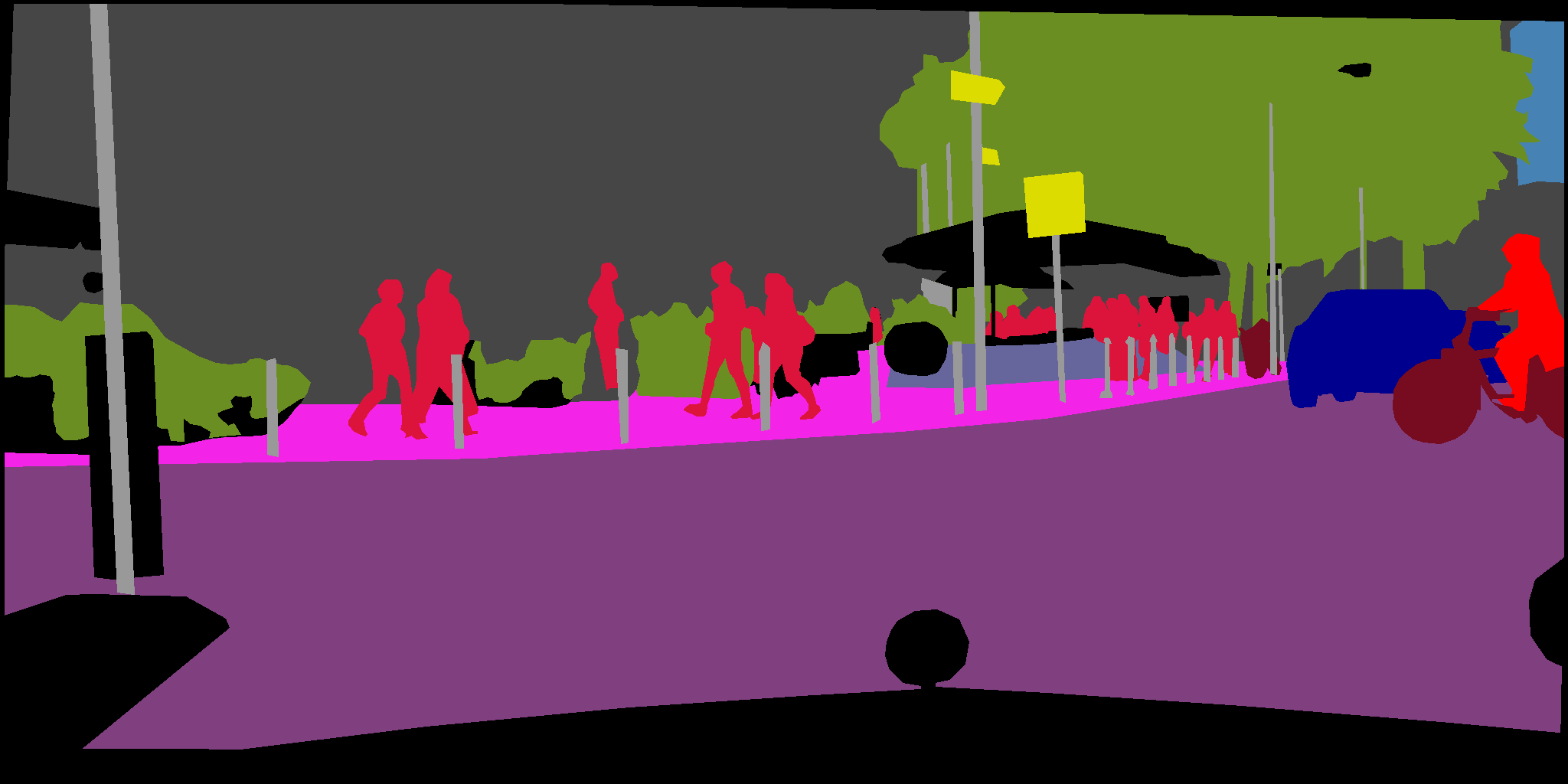}
        &
        \includegraphics[width=0.3\linewidth]{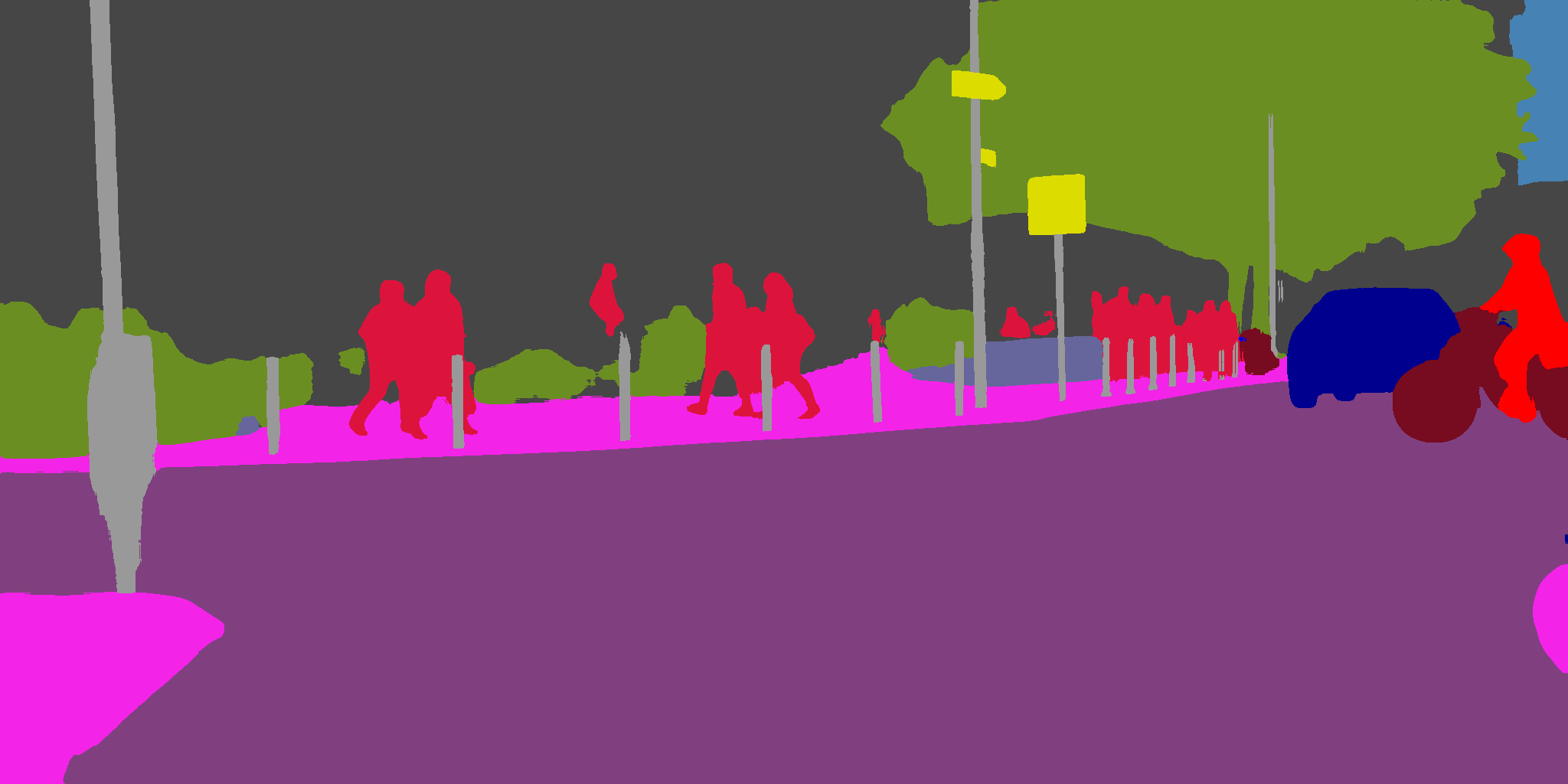} \\
        \bottomrule
    \end{tabular}
    \vspace{0.05in}
    \caption{
        The qualitative results of AFA-DLA-X-102 on the Cityscapes validation set. Our model can handle both fine and coarse details well and is robust towards different input scenes.
    }
	\label{fig:Cityscapes}
\end{figure*}

\begin{figure*}
    \centering
    \begin{tabular}{ccc}
        \toprule
        Input & Ground Truth & Prediction \\
        \midrule
        \includegraphics[width=0.3\linewidth]{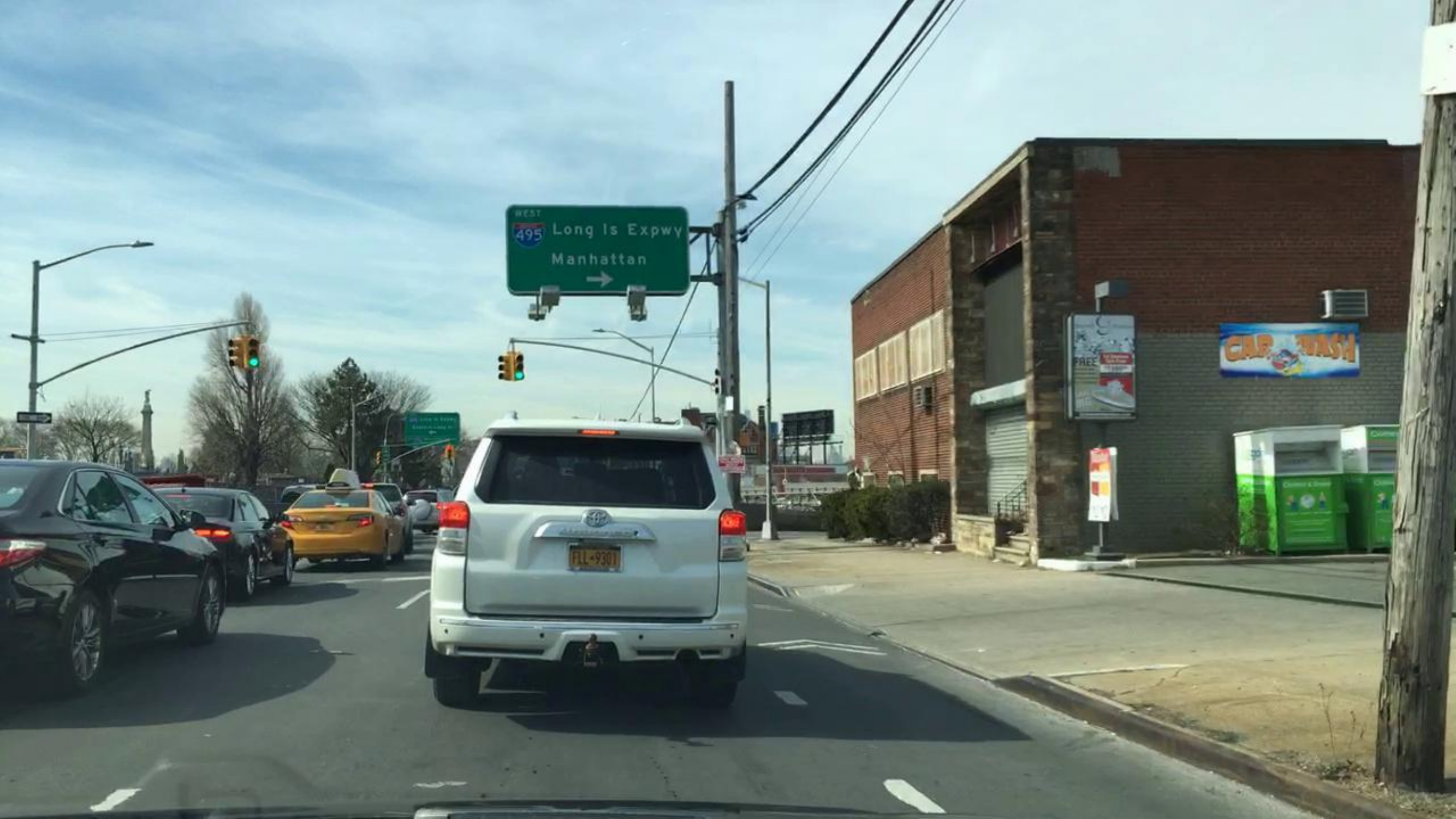}
        &
        \includegraphics[width=0.3\linewidth]{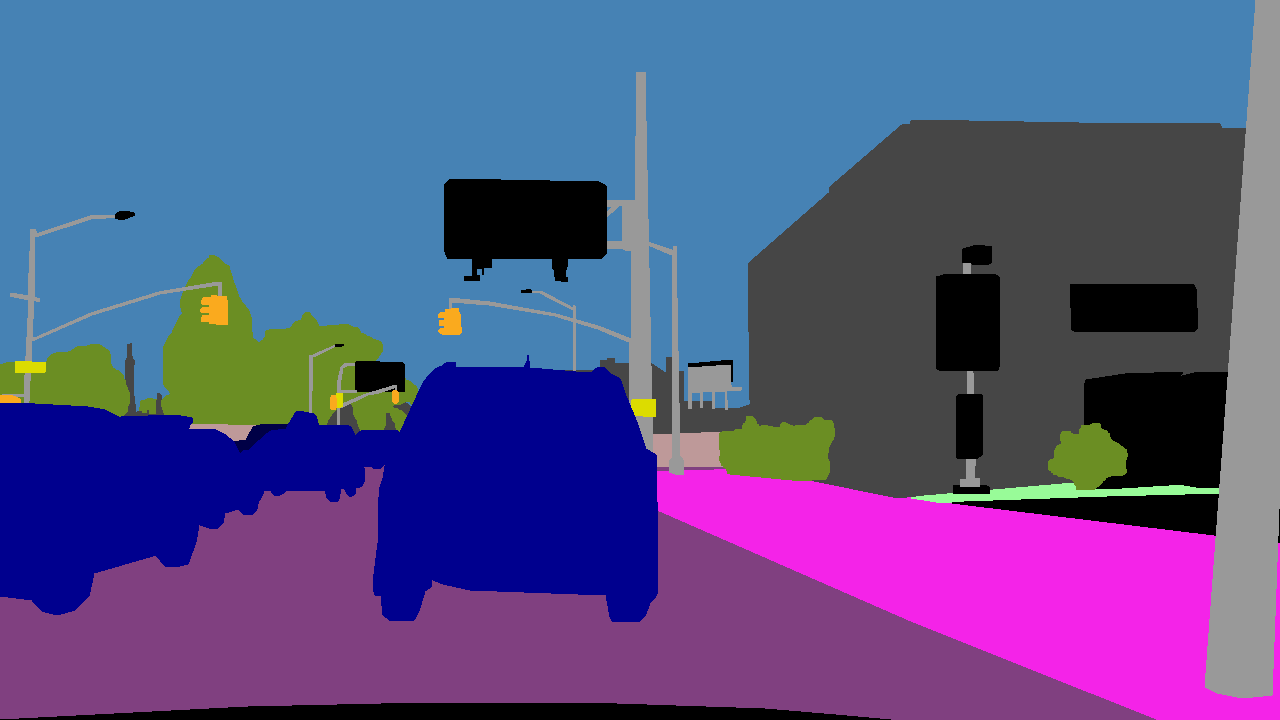}
        &
        \includegraphics[width=0.3\linewidth]{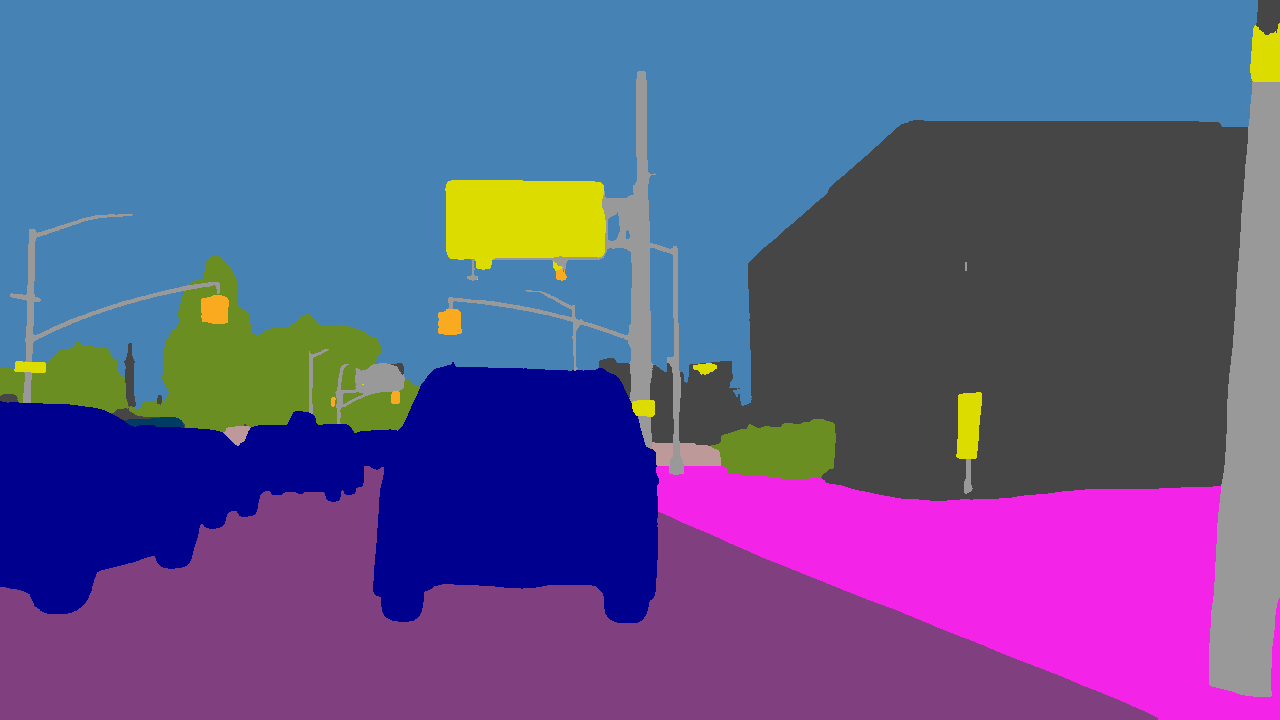} \\
        \includegraphics[width=0.3\linewidth]{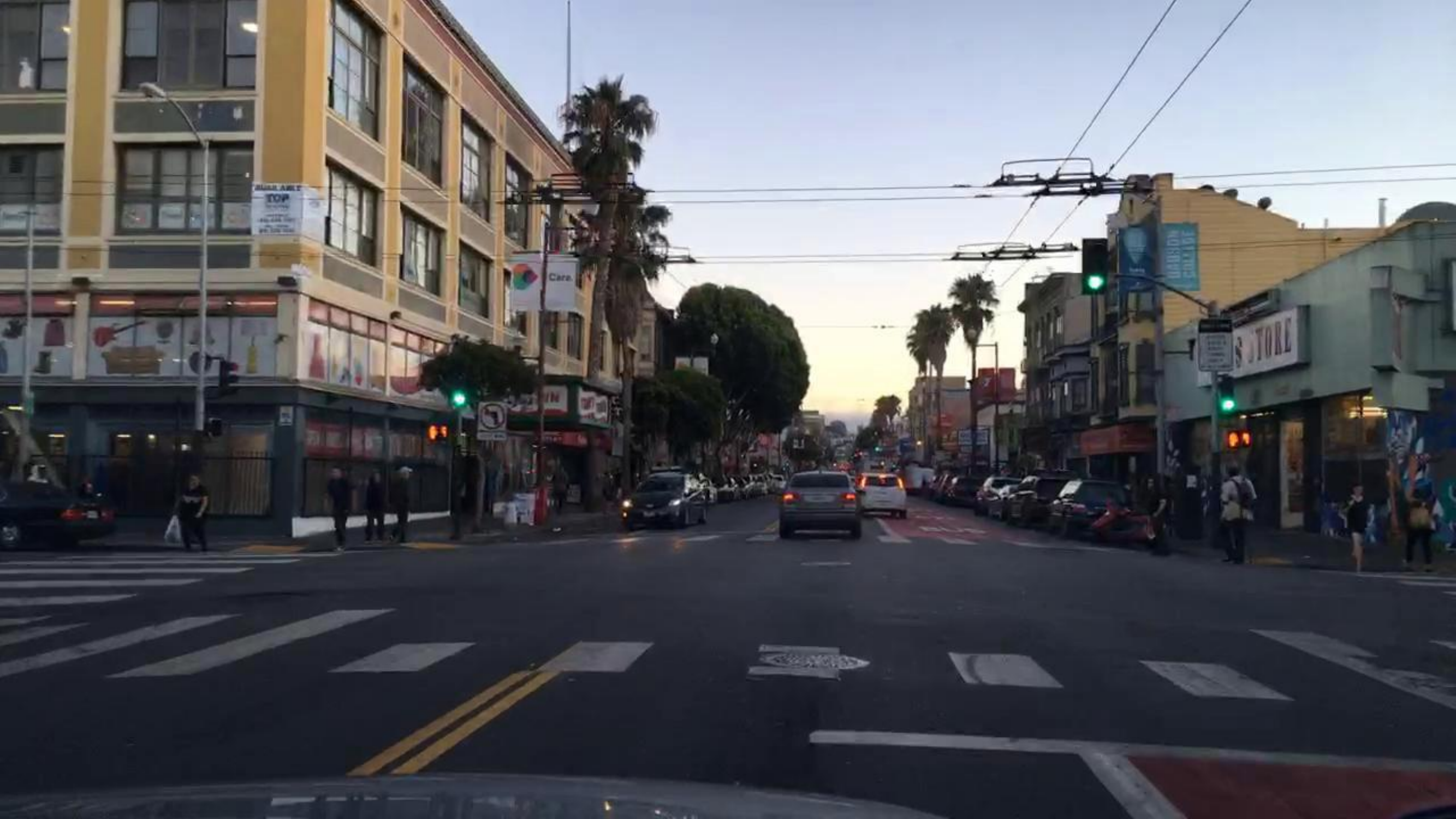}
        &
        \includegraphics[width=0.3\linewidth]{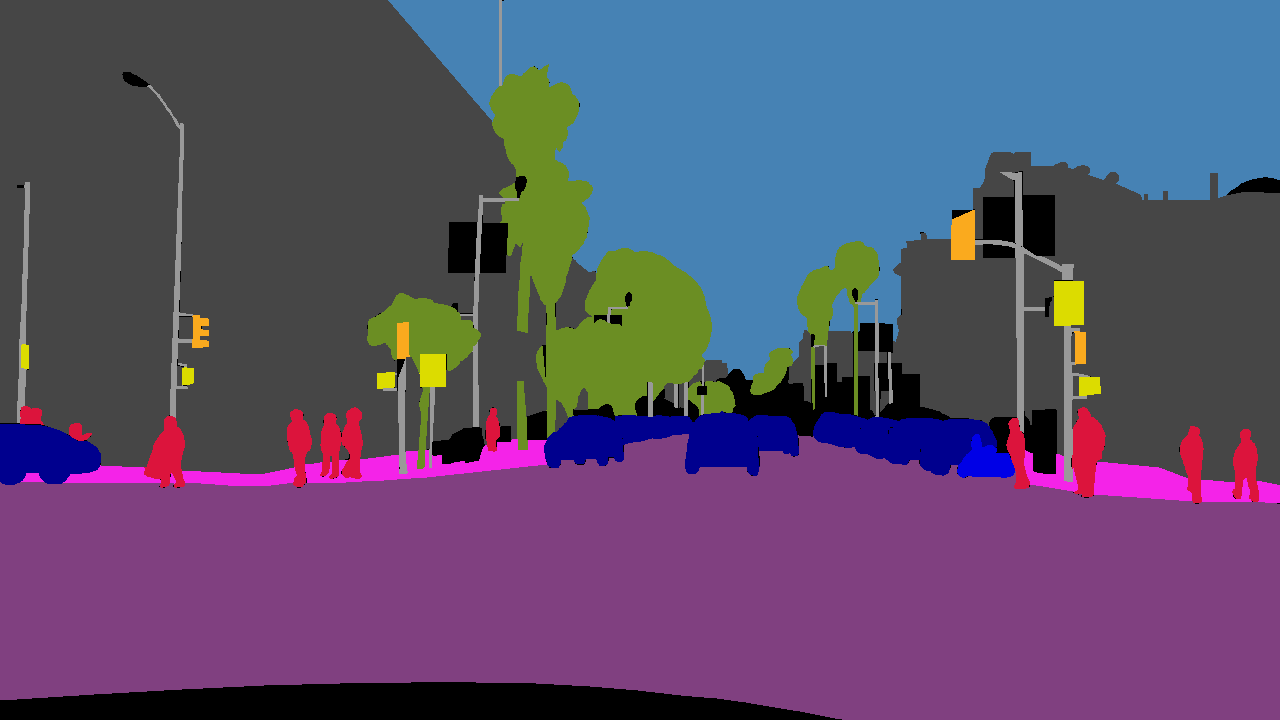}
        &
        \includegraphics[width=0.3\linewidth]{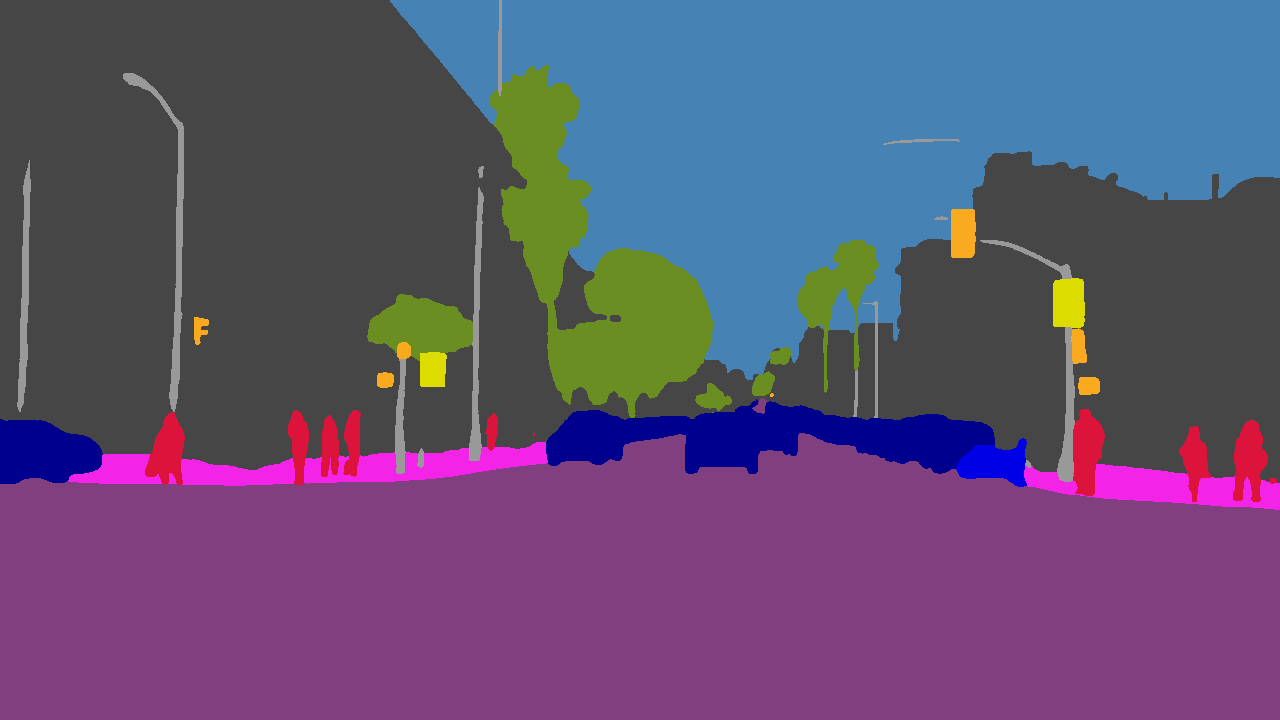} \\
        \includegraphics[width=0.3\linewidth]{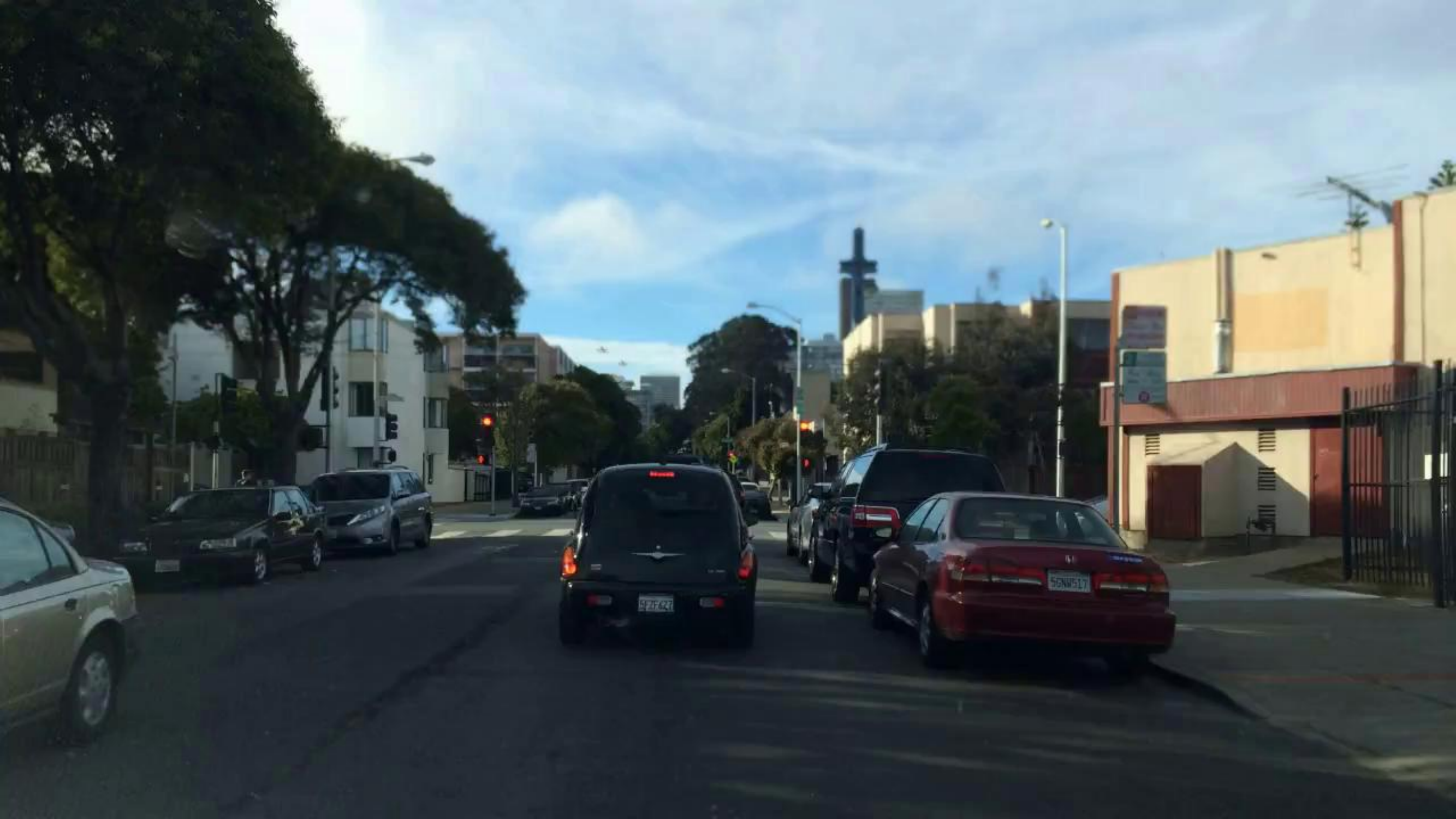}
        &
        \includegraphics[width=0.3\linewidth]{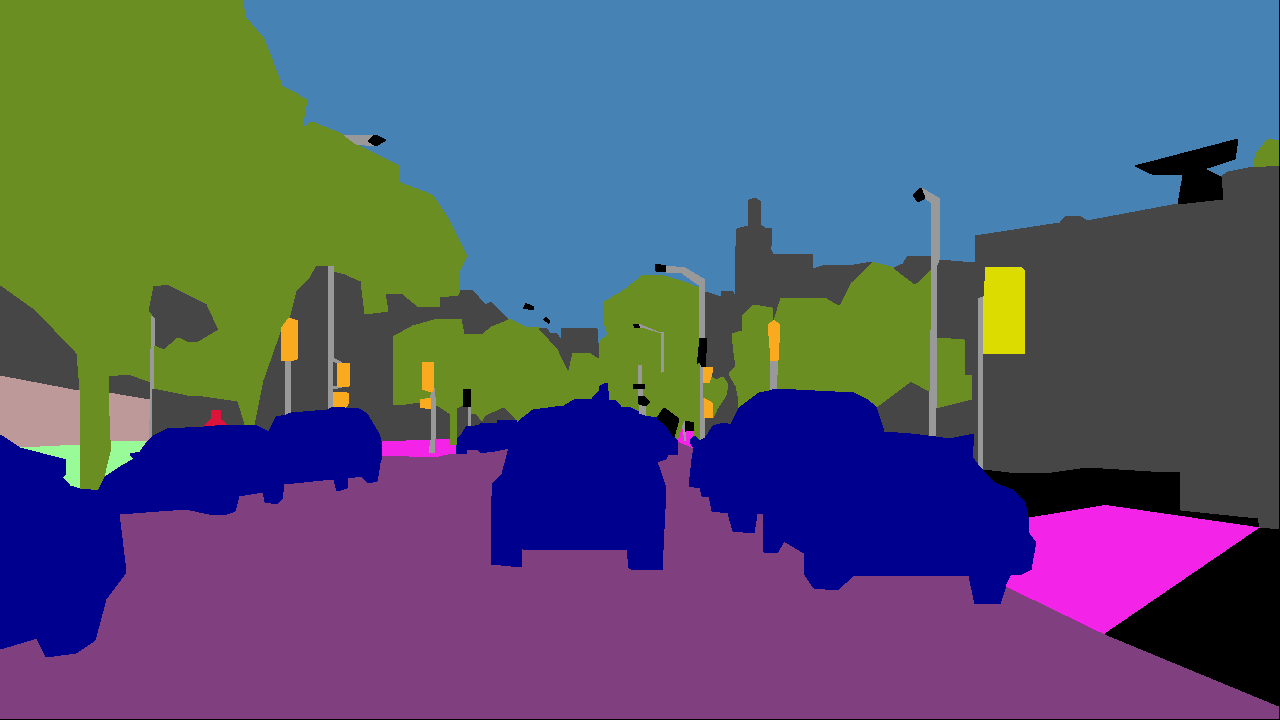}
        &
        \includegraphics[width=0.3\linewidth]{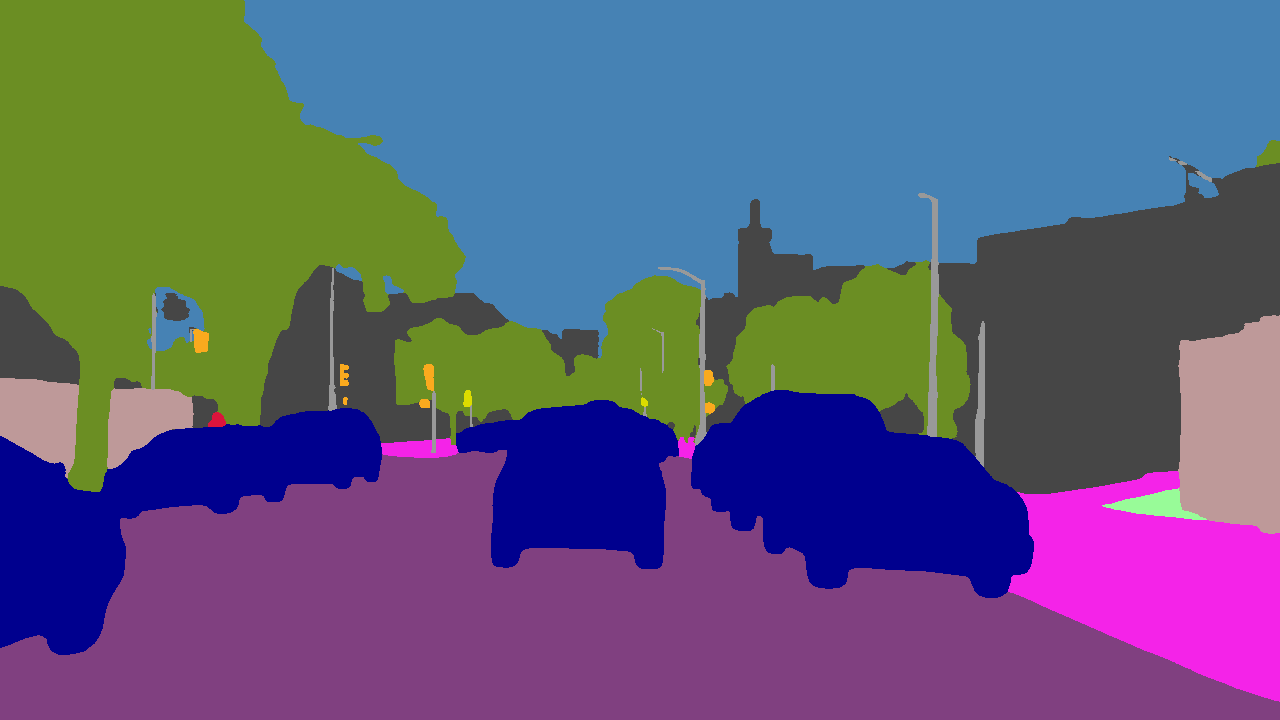} \\
        \includegraphics[width=0.3\linewidth]{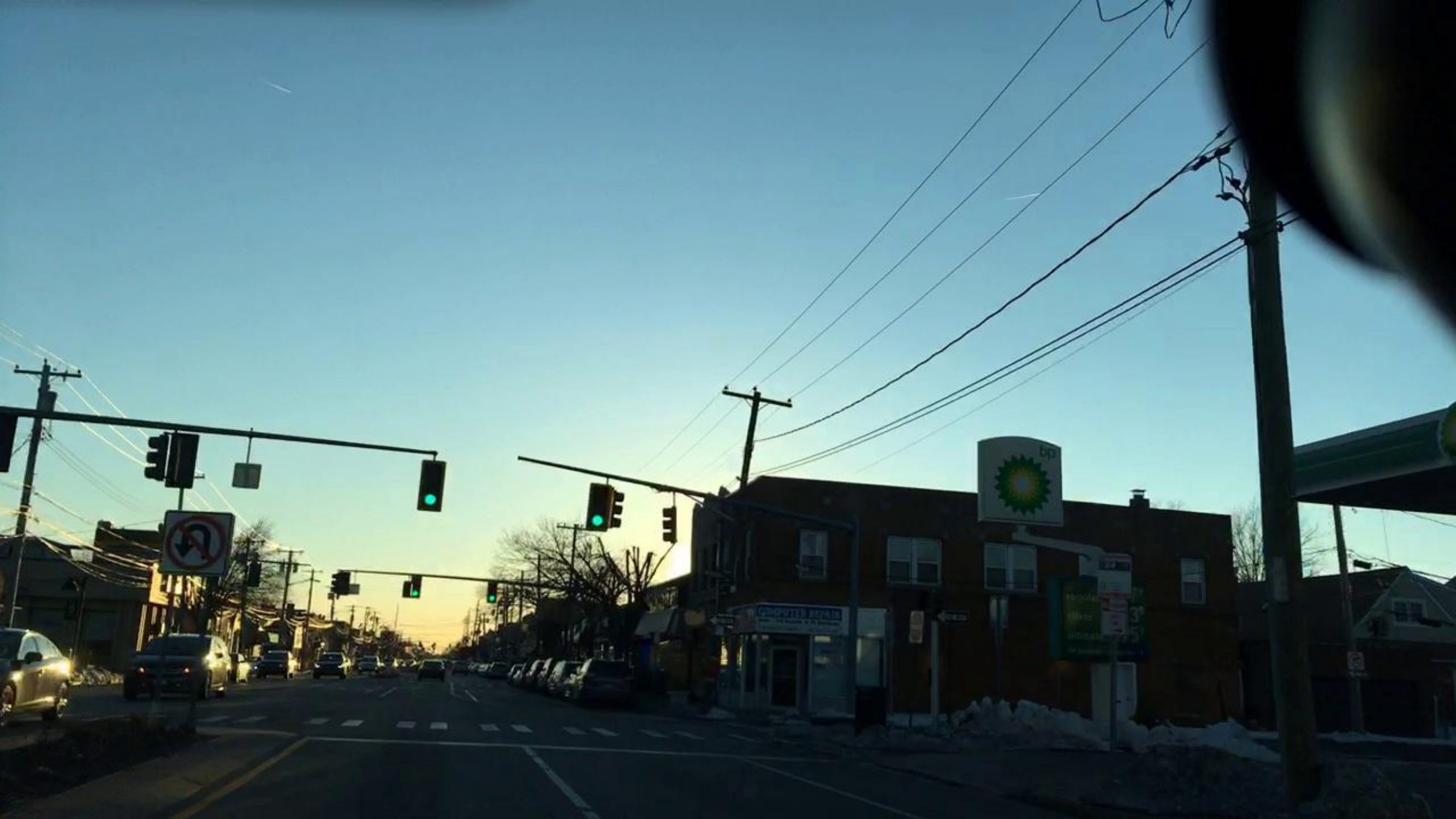}
        &
        \includegraphics[width=0.3\linewidth]{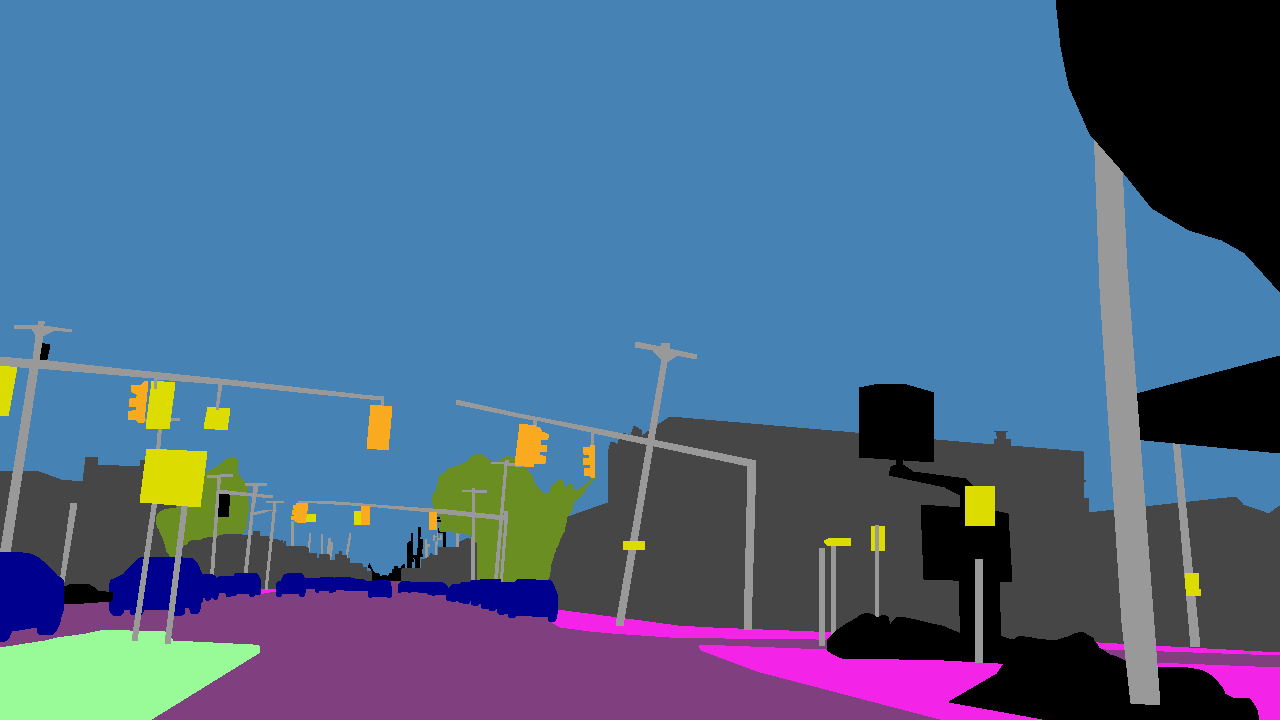}
        &
        \includegraphics[width=0.3\linewidth]{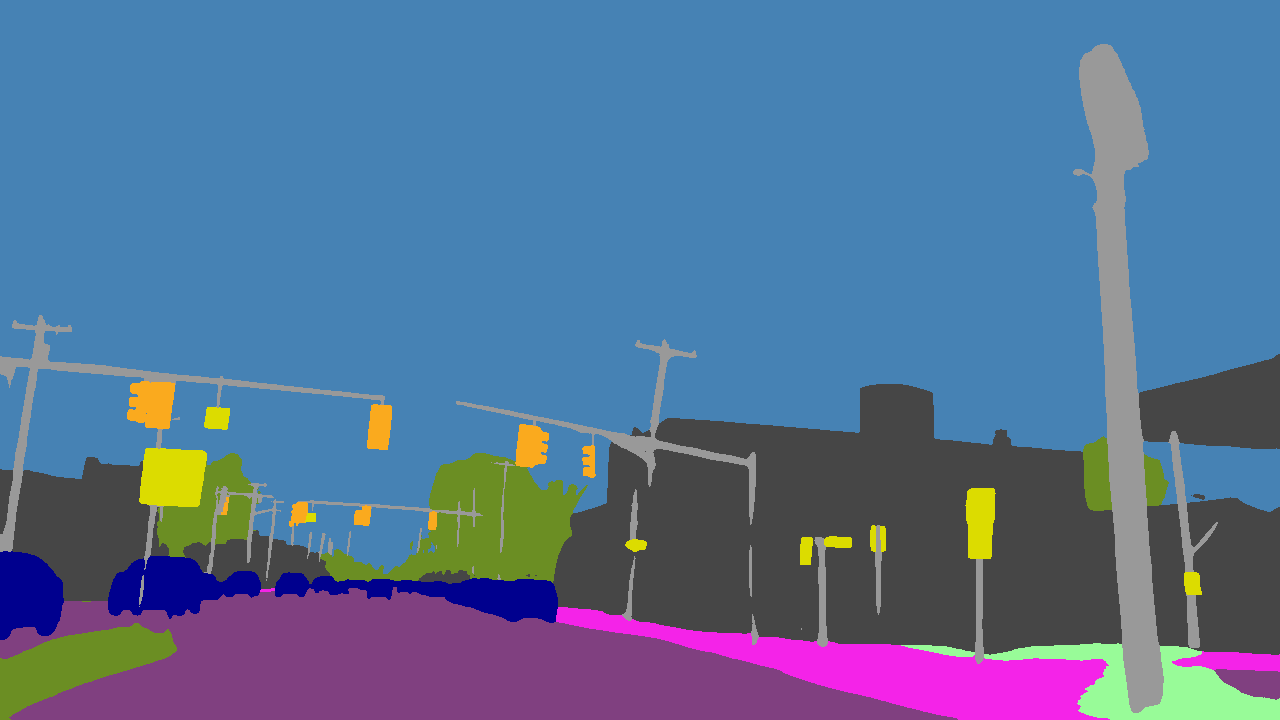} \\
        \bottomrule
    \end{tabular}
    \vspace{0.05in}
	\caption{
	    Qualitative results of AFA-DLA-169 on the BDD100K validation set. Our model can handle diverse urban scenes, with varying weather conditions and times of the day.
	}
	\label{fig:BDD}
\end{figure*}

\begin{figure*}[t]
    \centering
    \begin{tabular}{ccc}
        \toprule
        Input & Ground Truth & Prediction \\
        \midrule
        \includegraphics[width=0.3\linewidth]{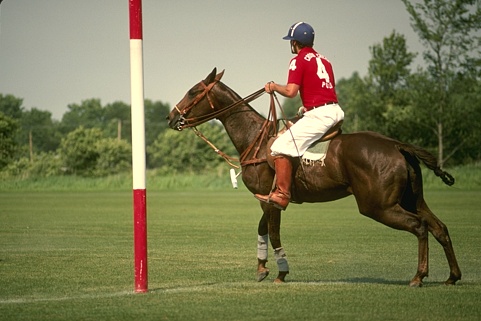}
        &
        \includegraphics[width=0.3\linewidth]{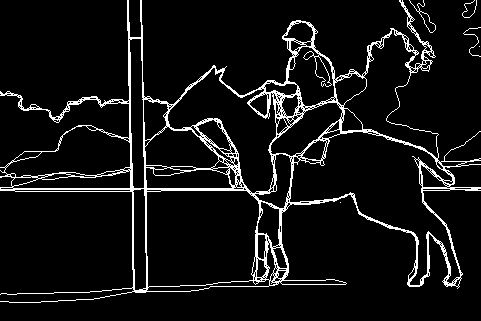}
        &
        \includegraphics[width=0.3\linewidth]{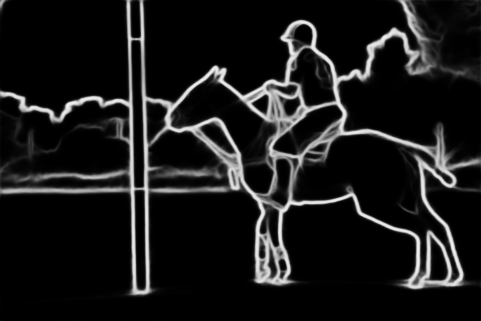} \\
        \includegraphics[width=0.3\linewidth]{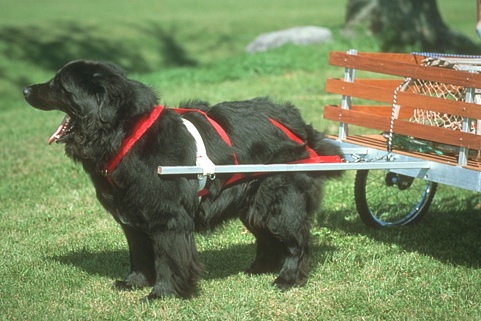}
        &
        \includegraphics[width=0.3\linewidth]{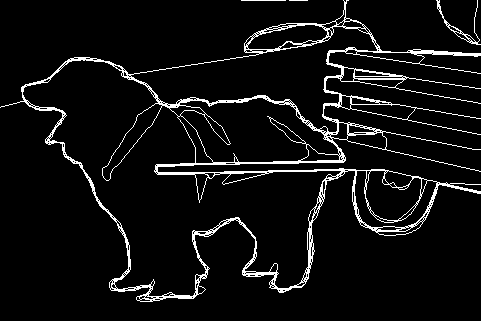}
        &
        \includegraphics[width=0.3\linewidth]{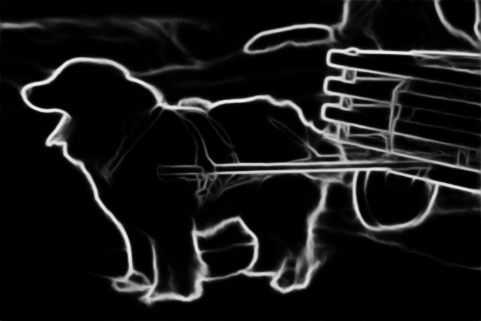} \\
        \includegraphics[width=0.3\linewidth]{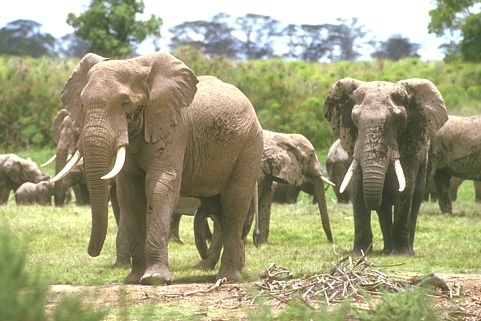}
        &
        \includegraphics[width=0.3\linewidth]{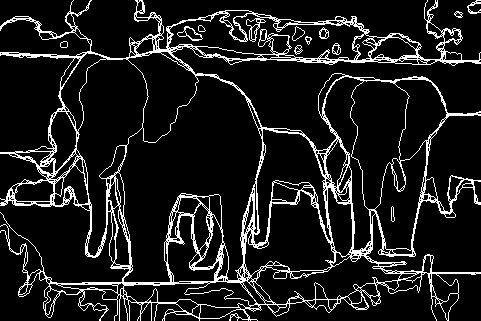}
        &
        \includegraphics[width=0.3\linewidth]{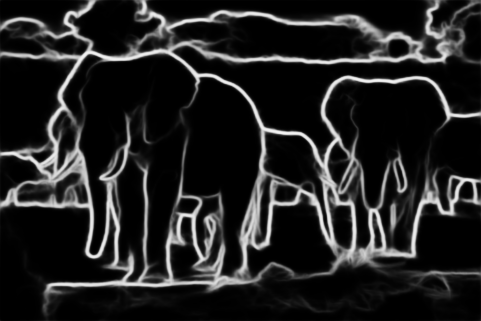} \\
        \includegraphics[width=0.3\linewidth]{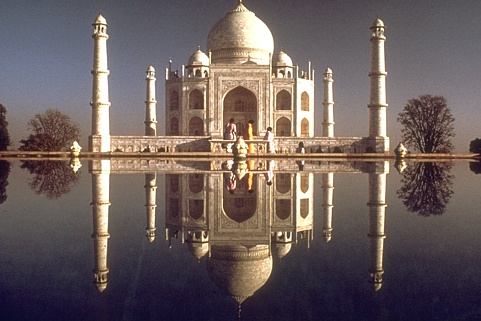}
        &
        \includegraphics[width=0.3\linewidth]{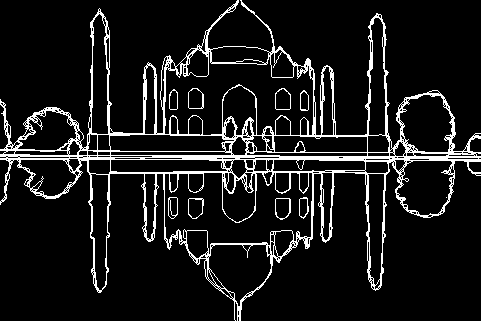}
        &
        \includegraphics[width=0.3\linewidth]{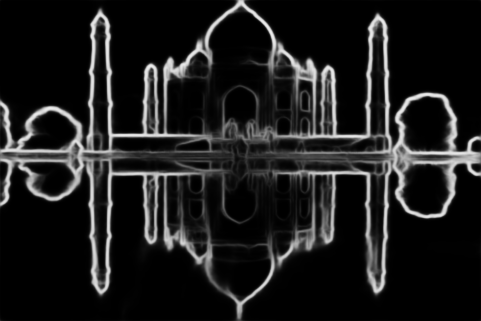} \\
        \bottomrule
    \end{tabular}
    \vspace{0.05in}
    \caption{
        Qualitative results of AFA-DLA-34 on the BSDS500 test set. Results are raw boundary maps obtained using multi-scale inference before Non-Maximum Suppression. Our model can predict both fine-grained scene details and object-level boundaries.
    }
	\label{fig:BSDS500}
\end{figure*}

\begin{figure*}[t]
    \centering
    \begin{tabular}{ccc}
        \toprule
        Input & Ground Truth & Prediction \\
        \midrule
        \includegraphics[width=0.3\linewidth]{images/NYUD/nyud1_img.pdf}
        &
        \includegraphics[width=0.3\linewidth]{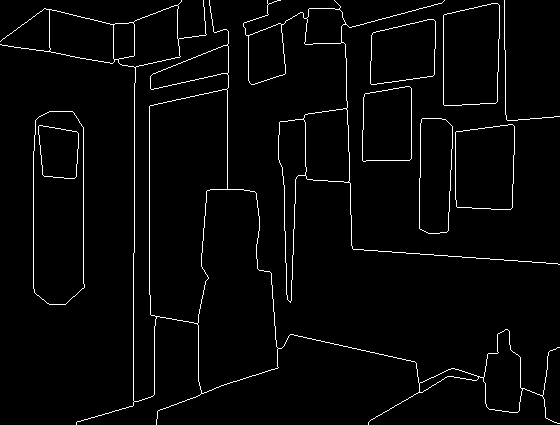}
        &
        \includegraphics[width=0.3\linewidth]{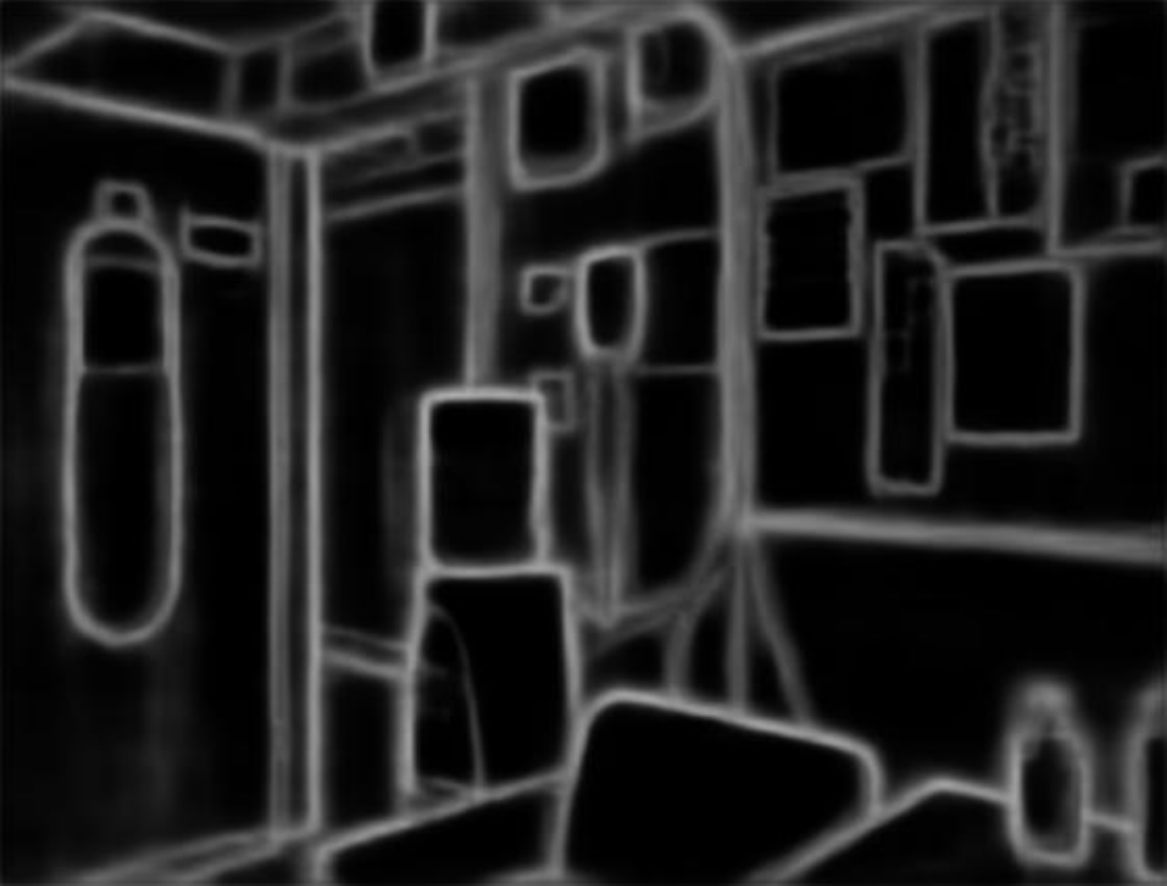} \\
        \includegraphics[width=0.3\linewidth]{images/NYUD/nyud2_img.pdf}
        &
        \includegraphics[width=0.3\linewidth]{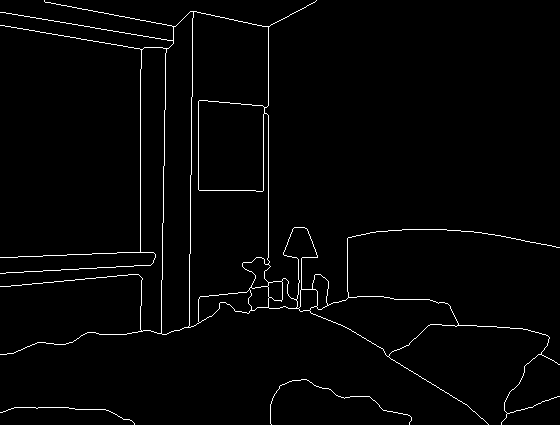}
        &
        \includegraphics[width=0.3\linewidth]{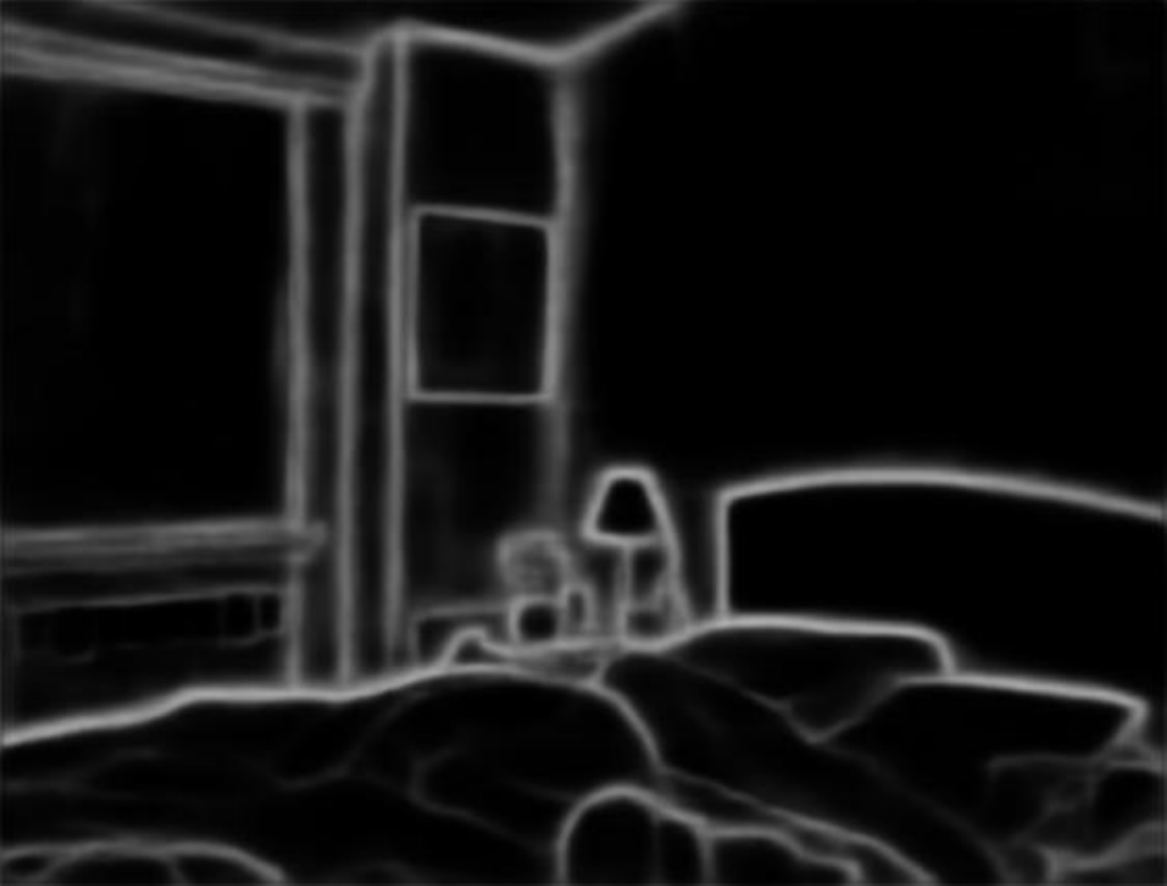} \\
        \includegraphics[width=0.3\linewidth]{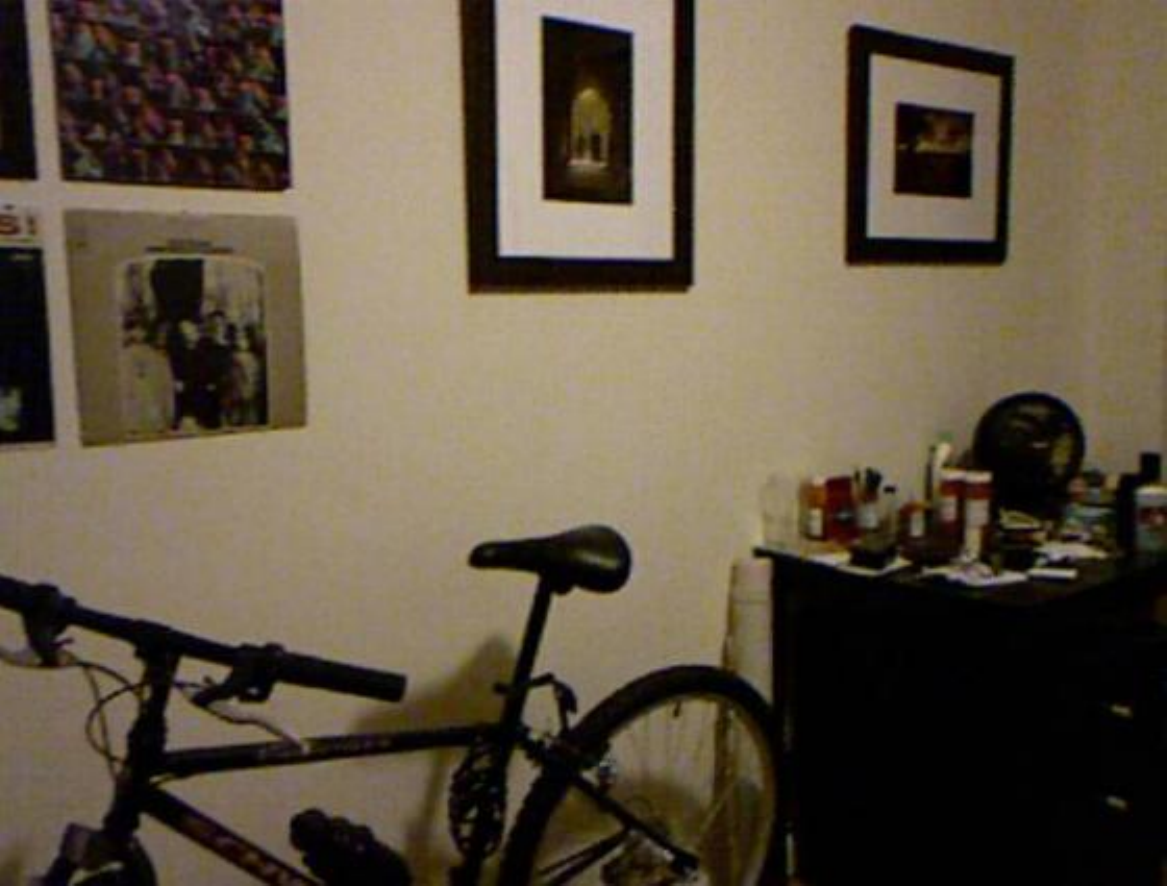}
        &
        \includegraphics[width=0.3\linewidth]{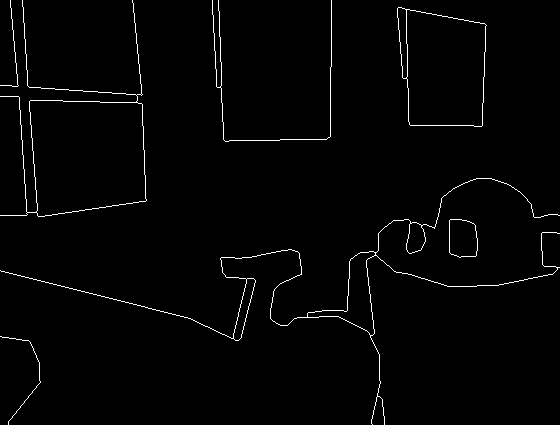}
        &
        \includegraphics[width=0.3\linewidth]{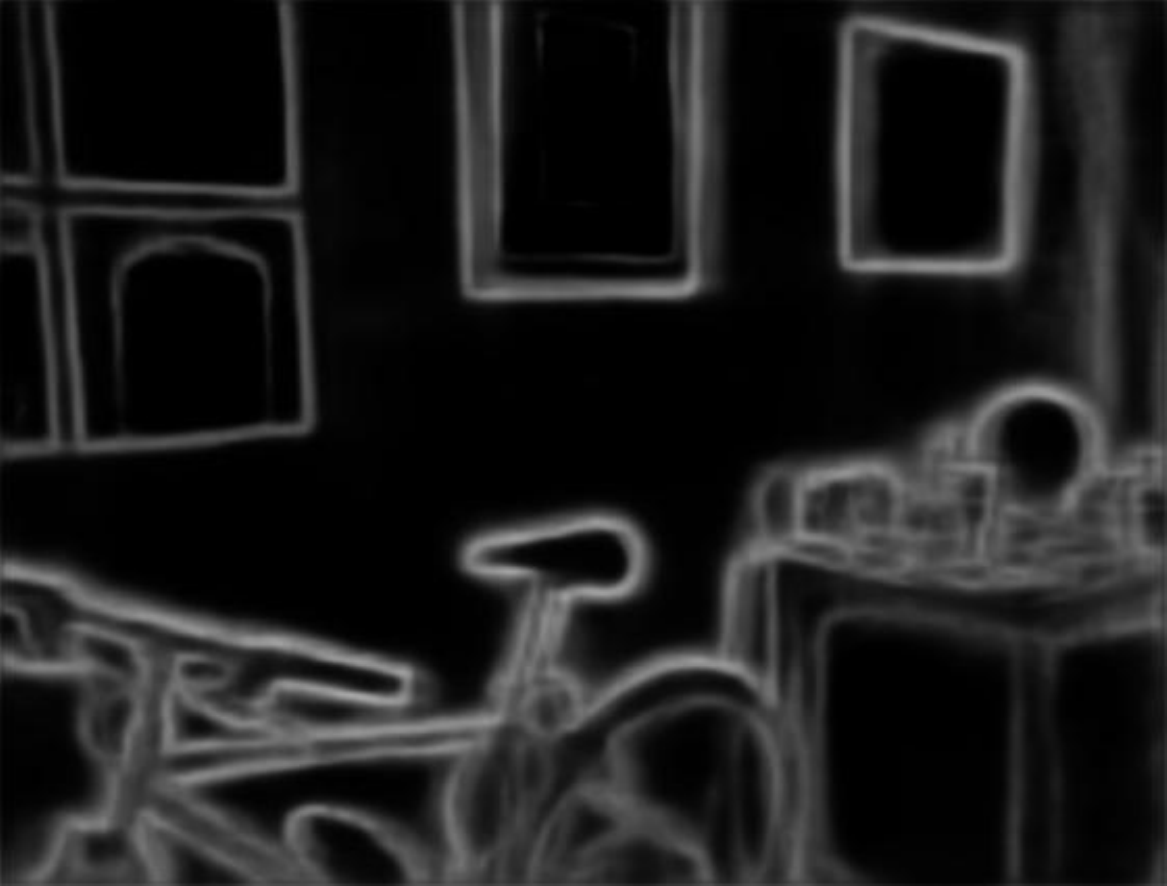} \\
        \includegraphics[width=0.3\linewidth]{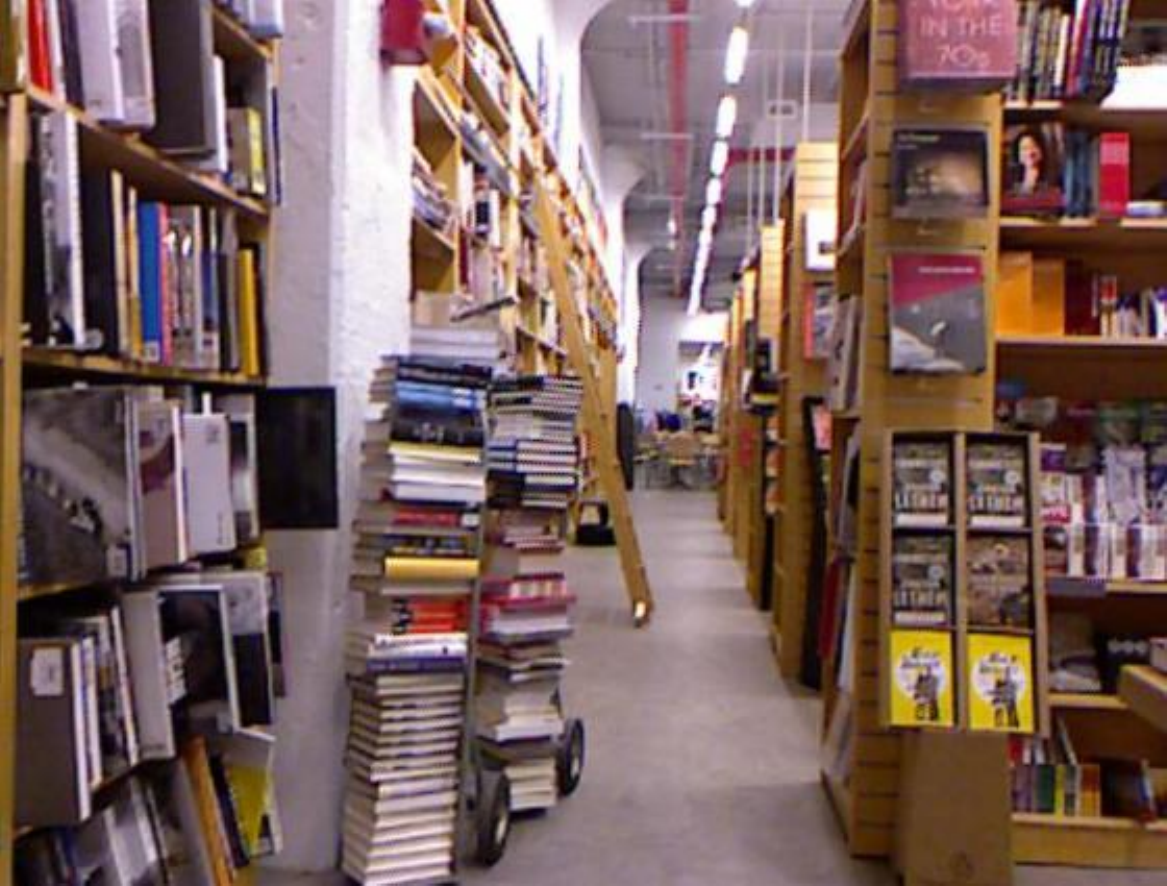}
        &
        \includegraphics[width=0.3\linewidth]{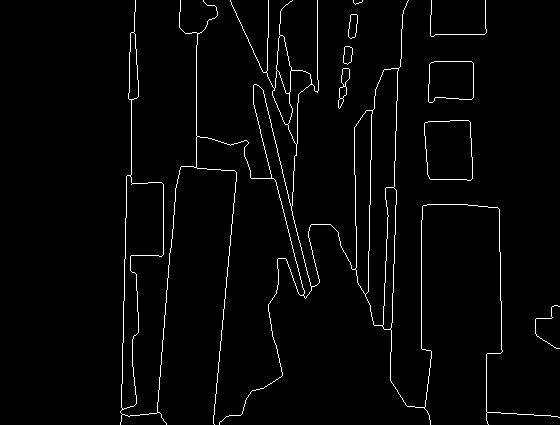}
        &
        \includegraphics[width=0.3\linewidth]{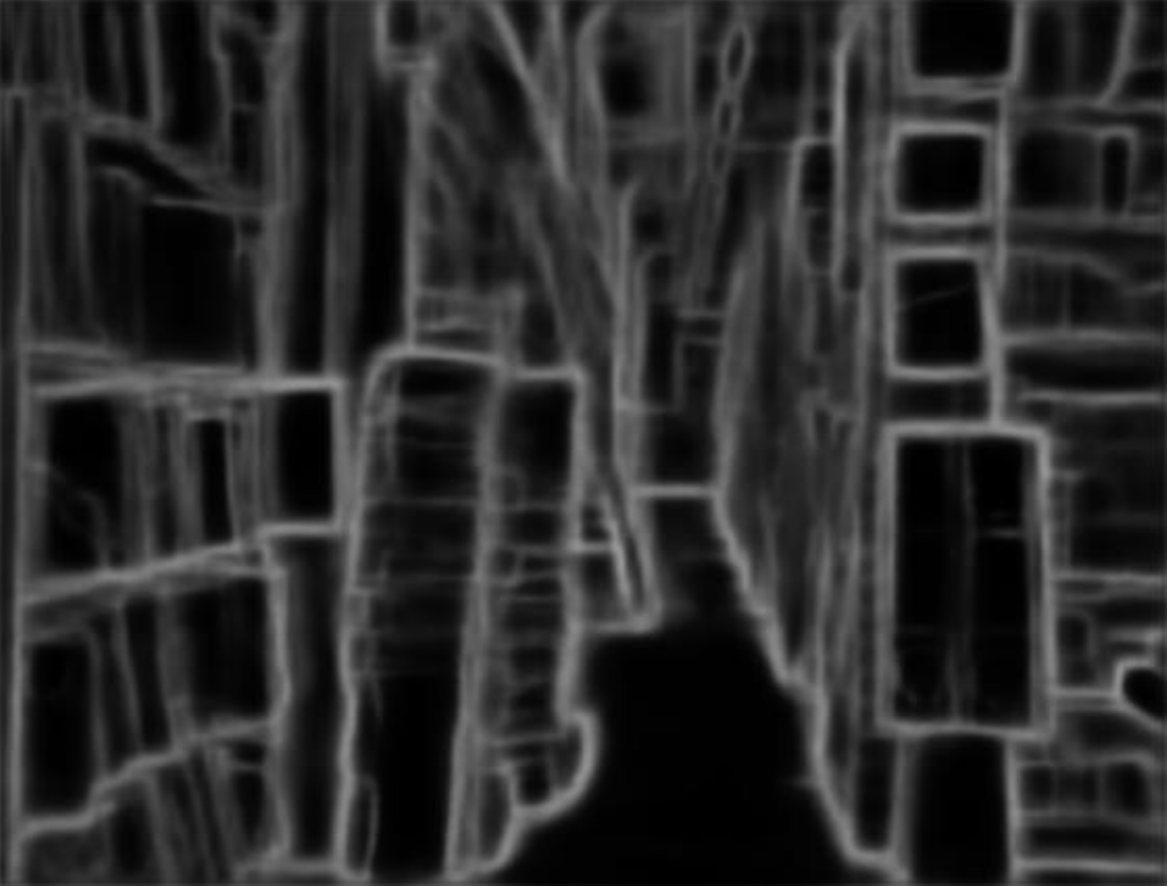} \\
        \bottomrule
    \end{tabular}
    \vspace{0.05in}
    \caption{
        Qualitative results of AFA-DLA-34 on the NYUDv2 test set. Results are raw boundary maps obtained by averaging predictions on both RGB and HHA images before Non-Maximum Suppression. Our model can extract more boundaries than the ground truth.
    }
	\label{fig:NYUD}
\end{figure*}

We provide more qualitative results in this section to visualize AFA-DLA's predictions.
We show full predictions of AFA-DLA on Cityscapes in Figure~\ref{fig:Cityscapes}, BDD100K in Figure~\ref{fig:BDD}, BSDS500 in Figure~\ref{fig:BSDS500}, and NYUDv2 in Figure~\ref{fig:NYUD}.
The results on Cityscapes show that our model can handle both fine and coarse details well and is robust towards different input scenes.
On BDD100K, the results show the ability of our model to handle more diverse urban scenes, with varying weather conditions and times of the day.
On both BSDS500 and NYUDv2, our model can predict both fine-grained scene details as well as object-level boundaries.
In particular, on NYUDv2, our model can recover more boundaries than the ground truth.
With results across different types of datasets and both semantic segmentation and boundary detection, AFA-DLA demonstrates its strong performance and applicability for dense prediction tasks.

\end{document}